\documentclass{article}
\usepackage[accepted]{icml2019}
\usepackage{flushend}
\usepackage{xr}
\makeatletter
\newcommand*{\addFileDependency}[1]{
  \typeout{(#1)}
  \@addtofilelist{#1}
  \IfFileExists{#1}{}{\typeout{No file #1.}}
}
\makeatother

\usepackage{subfiles}
\usepackage{blindtext}
\usepackage{appendix}
\usepackage[utf8]{inputenc} % allow utf-8 input
\usepackage[T1]{fontenc}    % use 8-bit T1 fonts
\usepackage{url}            % simple URL typesetting
\usepackage{booktabs}       % professional-quality tables
\usepackage{amsfonts,amsthm,amsmath}       % blackboard math symbols
\usepackage{nicefrac}       % compact symbols for 1/2, etc.
\usepackage{microtype}      % microtypography
\usepackage{latexsym}
\usepackage{bbm}
\usepackage{enumerate}% http://ctan.org/pkg/enumerate
\usepackage{paralist} %inline lists
\usepackage{bm}
\usepackage{wrapfig}
\usepackage{mathtools}
\usepackage[usenames,dvipsnames]{xcolor}
\usepackage{stmaryrd}
\usepackage{amsmath,amssymb,accents}
\usepackage{dsfont}
\usepackage{comment}
\usepackage{accents}
\DeclareMathOperator{\supp}{supp}
\usepackage{graphicx}
\usepackage{lipsum}
\usepackage{psfrag}
\usepackage{units}
\usepackage{setspace}
\allowdisplaybreaks
\newtheorem{theorem}{Theorem}

\newtheorem{definition}{Definition}

\newtheorem{lemma}{Lemma}
\newtheorem{remark}{Remark}

\newtheorem{proposition}{Proposition}

\newcommand\numberthis{\addtocounter{equation}{1}\tag{\theequation}}
{} %<- modify value to 
\newcommand{\vasti}{\bBigg@{3.5}}
\newcommand{\vast}{\bBigg@{4}}
\newcommand{\Vast}{\bBigg@{5}}
\newcommand{\Vastt}{\bBigg@{7}}
\makeatother
\newcommand{\be}{\begin{equation}}
\newcommand{\ee}{\end{equation}}
\newcommand{\ba}{\begin{align}}
\newcommand{\ea}{\end{align}}
\newcommand{\baa}{\begin{align*}}
\newcommand{\eaa}{\end{align*}}
\newcommand{\bea}{\begin{eqnarray}}
\newcommand{\eea}{\end{eqnarray}}
\newcommand{\beaa}{\begin{eqnarray*}}
\newcommand{\eeaa}{\end{eqnarray*}}
\newcommand{\p}[1]{\left(#1\right)}

\newcommand{\logdet}{\mathop{\mathrm{logdet}}}

\newcommand{\var}{\mathsf{var}}
\newcommand{\cp}[1]{\left\{#1\right\}}
\newcommand{\nmc}{n_{\mathsf{MC}}}

\newcommand{\brian}[1] {{\color{Green}\textsc{[#1]}}}

\newcommand{\igor}[1] {{\color{Orange}\textsc{[#1]}}}
\newcommand{\kristjan}[1] {{\color{Red}\textsc{[#1]}}}
\newcommand{\nam}[1] {{\color{Purple}\textsc{[#1]}}}
\newcommand{\ziv}[1] {{\color{Magenta}\textsc{[#1]}}}

\icmltitlerunning{Estimating Information Flow in Deep Neural Networks}

\begin{document}

\twocolumn[
\icmltitle{Estimating Information Flow in Deep Neural Networks}


\begin{icmlauthorlist}
\icmlauthor{Ziv Goldfeld}{mit,mitibm}
\icmlauthor{Ewout van den Berg}{mitibm,ibm}
\icmlauthor{Kristjan Greenewald}{mitibm,ibm}
\icmlauthor{Igor Melnyk}{mitibm,ibm}
\icmlauthor{Nam Nguyen}{mitibm,ibm}
\icmlauthor{Brian Kingsbury}{mitibm,ibm}
\icmlauthor{Yury Polyanskiy}{mit,mitibm}
\end{icmlauthorlist}

\icmlaffiliation{ibm}{IBM Research}
\icmlaffiliation{mit}{Massachusetts Institute of Technology}
\icmlaffiliation{mitibm}{MIT-IBM Watson AI Lab}

\icmlcorrespondingauthor{Ziv Goldfeld}{zivg@mit.edu}

\icmlkeywords{information theory, differential entropy, estimation, clustering, deep learning, information bottleneck}

\vskip 0.3in
]



\printAffiliationsAndNotice{} % otherwise use the standard text.

% BEDK - ICML guidelines recommend a 4-6 sentence abstract.
\begin{abstract}
We study the estimation of the mutual information $I(X;T_\ell)$ between the input $X$ to a deep neural network (DNN) and the output vector $T_{\ell}$ of its $\ell^{\mbox{th}}$ hidden layer (an ``internal representation''). Focusing on feedforward networks with fixed weights and noisy internal representations, we develop a rigorous framework for accurate estimation of $I(X;T_\ell)$. By relating $I(X;T_\ell)$ to information transmission over additive white Gaussian noise channels, we reveal that compression, i.e. reduction in $I(X;T_\ell)$ over the course of training, is driven by progressive geometric clustering of the representations of samples from the same class. Experimental results verify this connection. Finally, we shift focus to purely deterministic DNNs, where $I(X;T_\ell)$ is provably vacuous,
%either constant (discrete $X$) or infinite (continuous $X$),
and show that nevertheless,
%despite the degeneracy of $I(X;T_\ell)$,
these models also cluster inputs belonging to the same class. The binning-based approximation of $I(X;T_\ell)$ employed in past works to measure compression is identified as a measure of clustering, thus clarifying that these experiments were in fact tracking the same clustering phenomenon. Leveraging the clustering perspective, we provide new evidence that compression and generalization may \emph{not} be causally related and discuss potential future research ideas.
\end{abstract}

\section{Introduction}
\label{SEC:Intro}

\begin{figure*}[t!]
    \centering
    %,height=2.7cm
    \includegraphics[width=\textwidth]{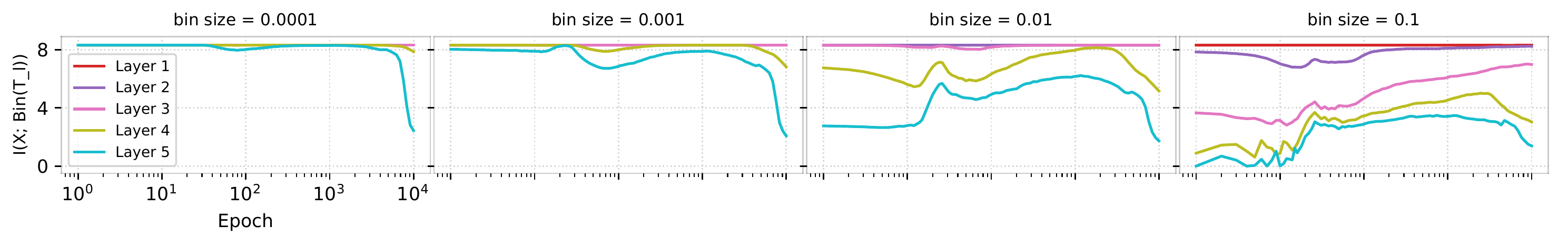}
    \vspace{-8.5mm}
    \caption{$I\big(X;\mathsf{Bin}(T_\ell)\big)$ vs. epochs for different bin sizes and the model in \cite{DNNs_Tishby2017}, where $X$ is uniformly distributed over a $2^{12}$-sized empirical dataset. The curves converge to $H(X)=\ln(2^{12})\approx 8.3$ for small bins.}
    \label{Fig:Binning_tanh}
    \vspace{-2mm}
\end{figure*}

Measuring the mutual information $I(X;T_{\ell})$ between the input feature $X$ to a deep neural network (DNN) and the output $T_{\ell}$ of its $\ell^{th}$ layer has long been a topic of research, with applications to unsupervised feature learning~\cite{Linsker1988,vandenOord2018,Hjelm2019} and deep learning analysis~\cite{DNNs_Tishby2017,DNNs_ICLR2018,Achille2018}. Mutual information is an appealing measure due to its invariance to smooth, invertible transformations and the fact that it has meaningful units (bits or nats). But, these benefits come at a price: mutual information is often impossible to compute analytically, and its estimation from samples is difficult~\cite{Paninski2003}. A variety of entropy, and thereby mutual information, estimators have been developed over the years, including $k$-nearest neighbors (kNN)~\cite{kozachenko1987sample,kraskov2004estimating}, kernel density~\cite{kandasamy2015nonparametric,han2017optimal}, and trainable neural estimators~\cite{Belghazi2018}. However, most previous information-theoretic studies of deep learning \cite{DNNs_Tishby2017,DNNs_ICLR2018} approximate the mutual information by discretizing neurons' outputs, an operation called `binning'.

%estimator has been a simple binning-based approach that discretizes the input features and outputs of the layer being examined, and then applies the standard estimator for the entropy of discrete distributions.

% Need to mention Amjad & Geiger somewhere, and also perhaps https://openreview.net/forum?id=rke4HiAcY7

The binning-based approach is attractive because of its computational efficiency when the number of bins is not too large; however, even mildly coarse discretizations can~yield inaccurate estimates.\footnote{The binning-based proxy approaches the true value when the bin sizes are shrunk to zero by definition \cite{cover2006elements}.} This fact is illustrated by the empirical mutual information plots from \cite{DNNs_Tishby2017,DNNs_ICLR2018} where the {\em compression}\/ phenomenon of the Information Bottleneck theory, i.e., a long-term decrease of $I(X;T_\ell)$ during DNN training, was studied. Both works plotted the binned mutual information $I\big(X;\mathsf{Bin}(T_\ell)\big)$ for deterministic DNNs (networks that deterministically map inputs to hidden representations) with $\mathsf{Bin}(T_\ell)$ being a per-neuron discretization of $T_\ell$. %, demonstrating the trajectories of this quantity over deterministic DNNs
%(namely, DNNs that define a deterministic mapping from input to hidden representations when their parameters are fixed. 
But, in deterministic DNNs with strictly monotone nonlinearities (e.g., tanh or sigmoid) the true mutual information $I(X;T_{\ell})$ is provably either infinite (continuous $X$) or a constant (discrete $X$). Therefore, the fluctuations of $I\big(X;\mathsf{Bin}(T_\ell)\big)$ observed during DNN training by \cite{DNNs_Tishby2017,DNNs_ICLR2018} must be due to estimation errors rather than changes in mutual information. Indeed, Fig.~\ref{Fig:Binning_tanh} illustrates how larger bin sizes can easily cause errors in $I\big(X;\mathsf{Bin}(T_\ell)\big)$ trajectories.

The degeneracy of $I(X;T_\ell)$ in deterministic DNNs is a consequence of $T_\ell$ being a deterministic function of $X$. 
%Recall that $I(X;T_{\ell})$ is a functional of the joint distribution of $(X,T_\ell) \sim P_{X,T_\ell} = P_XP_{T_\ell\mid X}$. 
If the DNN has continuous nonlinearities and $P_X$ is continuous, then so is $T_\ell$, and thus $I(X;T_\ell)=\infty$ (cf.\ Theorem 2.4 of~\citet{PolyWu-LecNotes}). When $P_X$ is discrete (e.g., when the features are discrete or if $X$ adheres to an empirical distribution over the dataset), the mutual information equals the entropy $H(X)$, a constant that is independent of the DNN parameters. This follows because the mapping from a discrete $X$ to the support of $T_\ell$ is injective for strictly monotone nonlinearities for any but a measure-zero set of weights. Both continuous and discrete degeneracies were previously observed 
\cite{geiger2018Bottleneck,kolchinsky2019caveats}. These are a consequence of the fact that deterministic DNNs can encode information about $X$ in arbitrarily fine variations of $T_\ell$ essentially without loss, even if deeper layers have fewer neurons.

That said, the estimate  $I\big(X;\mathsf{Bin}(T_\ell)\big)$ is system dependent and its compression observed in past works seems meaningful. What mechanism drives this compression? To answer this question, we develop a rigorous framework for tracking the flow of information in DNNs. To ensure $I(X;T_\ell)$ is useful for studying the learned representations, the map $X\mapsto T_\ell$ must be a stochastic parameterized channel whose parameters are the DNN's weights and biases. To obtain pertinent insights into practical systems, we impose the following requirements on the framework.
\begin{inparaenum}[(R1)]
\item The stochasticity should be intrinsic to the operation of the DNN, so that the characteristics of mutual information measures are related to the learned internal representations.
\item The stochasticity should relate the mutual information to the deterministic binned version $I\big(X;\mathsf{Bin}(T_\ell)\big)$, since this is the object whose compression was observed; this requires the injected noise to be isotropic over the domain of $T_\ell$ analogously to the per-neuron binning operation.
\item Most importantly, the network trained under this stochastic model should be closely related to those trained in practice.
\end{inparaenum}

In Section~\ref{SEC:Noisy_DNNs} we propose a stochastic DNN framework~in which independently and identically distributed (i.i.d.) Gaussian noise is added to the output of each of the DNN's hidden layer neurons. This makes the map $X\mapsto T_{\ell}$ stochastic, ensures the data processing inequality is satisfied, and makes $I(X;T_\ell)$ reflect the DNN's true operating conditions, per (R1). Since the noise is centered and isotropic, (R2) holds. As for (R3), experiments show that the DNN's learned representations and performance are not meaningfully affected by the addition of noise, for variances $\beta^2$ not too large. 
%it does not require any modification of the model training procedure and it is computationally inexpensive. 
%However, when relatively coarse discretizations are used, this procedure yields unreliable estimates of mutual information, as shown in Figure~\ref{Fig:Binning_tanh}, where only the smallest bin size produces correct estimates. 

%this paragraph starts talking about the estimation problem. the contents is not fitting the new presentation + we should try to shorten it. I wanted your take of whether you think we should significantly tone down its emphasis, essentially saying we tie our problem to another one from the anonymized paper and borrow their results to get what we need. ? (then say that on top of that we develop tools for implementing the estimator and theoretically vouch for them as well. sorry for the long message. ) sure felt like it too.%YEah that sounds like a good plan. As-is it goes into maybe too much detail for the intro?

Under the stochastic model, $I(X;T_\ell)$ has no exact analytic expression and is intractable to evaluate numerically. In Section~\ref{SEC:MI_estimation} we construct a provably accurate estimator for $I(X;T_\ell)$, employing a sampling technique that decomposes the estimation problem into several instances of the differential entropy estimation setup studied in \cite{anonymized_ISIT_estimation2019}. Leveraging the risk bounds derived therein, we prove that for any dimension of the hidden layer, the risk of our mutual information estimator converges at the parametric rate of estimation (see Section \ref{SEC:MI_estimation}). We then introduce a method for efficient implementation of our mutual information estimator and derive theoretical guarantees on its accuracy. 
Having a tool for accurately tracking $I(X;T_\ell)$ over the course of stochastic DNN training, we focus on the geometric phenomenon that drives its fluctuations. We relate $I(X;T_\ell)$ to the well-understood notion of data transmission over additive white Gaussian noise (AWGN) channels. Namely, $I(X;T_\ell)$ is the aggregate information transmitted over the channel $P_{T_\ell|X}$ with input $X$ drawn from a constellation defined by the data samples and the noisy DNN parameters. As training progresses, the representations of inputs from the same class tend to~cluster together and become increasingly indistinguishable at the channel's output, thereby decreasing $I(X;T_\ell)$. %WAS IN THE ORIGINAL: In all our experiments, tracking the trajectories of $I(X;T_\ell)$ alongside scatter plots of the input samples propagated up to layer $T_\ell$ reveals a remarkable correspondence between compression and how clustered the propagated samples are.
Furthermore, these clusters tighten as one moves into deeper layers, providing evidence that the DNN's layered structure progressively improves the representation of $X$ to increase its relevance for~$Y$. %WAS IN THE ORIGINAL: The identification of the clustering phenomena clarifies what drives compression and elucidates some of the machinery DNNs employ for learning. 

In Section~\ref{SUBSEC:SZT} we experimentally demonstrate that, in some cases, $I(X;T_{\ell})$ exhibits compression during training of noisy DNNs. It is further shown that regardless of whether $I(X;T_{\ell})$ compresses or not, its fluctuations always track the degree of clustering in the internal representation space. Finally, in Section~\ref{SUBSEC:MNIST}, we examine clustering in a deterministic DNN. Several methods for measuring clustering (valid for both noisy and deterministic systems) are identified and used to show that clusters of inputs in learned representations form during deterministic DNN training as well. We complete the circle back to $I\big(X;\mathsf{Bin}(T_\ell)\big)$ by demonstrating that this quantity measures clustering. This explains what previous works were actually observing in those deterministic systems: not (the theoretically impossible) compression of mutual information, but increased clustering of hidden representations. Leveraging the clustering perspective we then provide new evidence that compression of $I(X;T_\ell)$ and generalization may \emph{not} be causally related. The geometric clustering of internal representations is thus the fundamental phenomenon of interest, and we aim to test its connection to generalization performance, theoretically and experimentally, in future work. Code to replicate the experiments in this paper is in preparation, and~this preprint will be updated when it is available.

\section{Preliminary Definitions}
\label{SEC:Noisy_DNNs}

\begin{figure}[t!]	\centering
\includegraphics[width=0.355\textwidth]{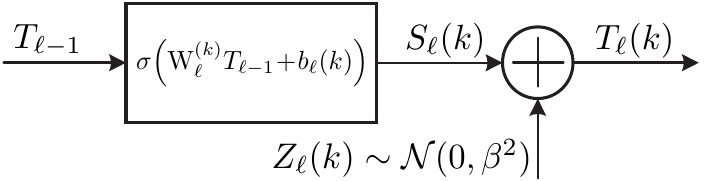}
%\vspace{-4mm}
\caption{$k$th noisy neuron in layer $\ell$: $\mathrm{W}_\ell^{(k)}$ and $b_\ell(k)$ are the $k$th row/entry of the weight matrix and the bias, respectively.}\label{FIG:noisy_neuron}
\vspace{-3mm}
\end{figure}

\textbf{Noisy DNNs:} For integers $k\leq\ell$, let $[k\mspace{-3mu}:\mspace{-3mu}\ell]\triangleq\big\{i\in\mathbb{Z}\big| k\leq i \leq \ell\big\}$ and use $[\ell]$ when $k=1$. Consider a noisy DNN with $L+1$ layers $\{T_\ell\}_{\ell\in[0:L]}$, with input $T_0=X$ and output $T_L$. The $\ell$th hidden layer, $\ell\in[L-1]$, is described by $T_\ell=f_\ell(T_{\ell-1})+Z_\ell$, where $f_\ell:\mathbb{R}^{d_{\ell-1}}\to\mathbb{R}^{d_\ell}$ is a deterministic function of the~previous layer and $Z_\ell\sim\mathcal{N}\big(0,\beta^2 \mathrm{I}_{d_\ell}\big)$; no noise is injected to the output, i.e., $T_L=f_L(T_{L-1})$. We set $S_\ell\triangleq f_\ell(T_{\ell-1})$ and use $\varphi_\beta$ for the probability density function (PDF) of $Z_\ell$. The functions $\{f_\ell\}_{\ell\in[L]}$ can represent any type of layer (fully connected, convolutional, max-pooling, etc.). Fig.~\ref{FIG:noisy_neuron} shows a neuron in a noisy DNN.

\begin{figure*}[!t]%
%\vspace{-3mm}
  \centering
  \includegraphics[width=0.8\textwidth]{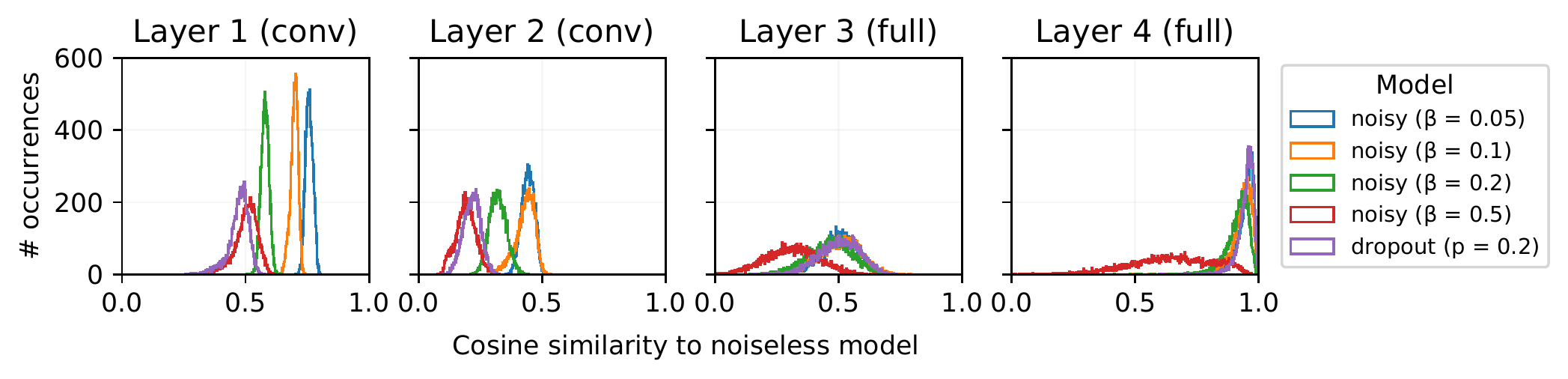}%
  \vspace{-5mm}
  \caption{Cosine similarity histograms between internal representations of deterministic, noisy, and dropout MNIST CNNs.}%
  \label{fig:MNIST-cos-similarity}%  
  \vspace{-2mm}
\end{figure*}

To explore the relation between noisy and deterministic DNNs under conditions representative of current machine learning practices, we trained four-layer convolutional neural networks (CNNs) on MNIST~\citep{MNIST}. The CNNs used different internal noise levels (including $\beta=0$) and one used dropout instead of additive noise. We measured their performance on the validation set and characterized the cosine similarities between their internal representations (see the appendix for full details). The results in Table~\ref{tab:MNIST-performance} show small amounts of internal noise ($\beta \leq 0.1$) have a minimal impact on classification performance, while dropout strongly improves it. The histograms in Fig.~\ref{fig:MNIST-cos-similarity} show that the noisy (for small $\beta$) and dropout models learn internal representations similar to those learned by the deterministic model. In this high-dimensional space, unrelated representations would create cosine similarity histograms with zero mean and standard deviation between 0.02--0.3, so the observed values are quite large. As~expected, dissimilarity increases as $\beta$ increases, and similarity is lower for the internal layers (2 and 3). %, and that dropout induces more dissimilarity in the early layers (1 and 2) than small amounts of additive noise does.

\begin{table} \centering
     \caption{MNIST validation errors for different models: mean $\pm$ standard deviation over eight initial random seeds.}
    \label{tab:MNIST-performance}
    \vskip 2.0mm
    \scalebox{0.9}{
    \begin{tabular}{ll}\toprule
      {\bf Model} & {\bf \# Errors}\\ %& {\bf Loss} \\ 
      \midrule
      Deterministic          & 50 $\pm$ 4.6\\ %   & 0.015 $\pm$ 0.0011 \\
      Noisy ($\beta = 0.05$) & 50 $\pm$ 5.0 \\ %  & 0.014 $\pm$ 0.00071 \\
      Noisy ($\beta = 0.1$)  & 51 $\pm$ 6.9 \\ %  & 0.016 $\pm$ 0.0019 \\
      Noisy ($\beta = 0.2$)  & 86 $\pm$ 9.8 \\ %  & 0.027 $\pm$ 0.0026 \\
      Noisy ($\beta = 0.5$)  & 2200 $\pm$ 520 \\ %& 0.71 $\pm$ 0.21 \\
      Dropout ($p = 0.2$)    & 39 $\pm$ 3.9 \\ %  & 0.011 $\pm$ 0.00073 \\ 
      \bottomrule
    \end{tabular}
    }
    \vspace{-3mm}
\end{table}

\textbf{Mutual Information:} Noisy DNNs induce a stochastic map from $X$ to the rest of the network, described by the conditional distribution $P_{T_1,\ldots,T_L|X}$. The corresponding PDF\footnote{$P_{T_1,\ldots,T_L|X=x}$ is absolutely continuous with respect to (w.r.t.) the Lebesgue measure for all $x\in\mathcal{X}$.}~is $p_{T_1,\ldots,T_L|X=x}$. %Its marginals are denoted by keeping only the relevant variables in the subscript.
Assuming $X\sim P_X$, the system is described by the joint distribution $P_{X,T_1,\ldots,T_L}\triangleq P_XP_{T_1,\ldots,T_L|X}$, under which $X-T_1-\ldots-T_{L-1}-T_L$ forms a Markov chain. For each $\ell\in[L-1]$, we study the mutual information $I(X;T_\ell)\triangleq h(T_\ell)-\int\mathsf{d}P_X(x)h(T_\ell|X=x)$. The composition of Gaussian noises and nonlinearities renders the stochastic map $X\mapsto T_\ell$ too complicated to analytically compute $I(X;T_\ell)$. Therefore, we focus on estimating $I(X;T_\ell)$ from samples.

%Although  are readily sampled from using the DNN's forward pass, these distributions are too complicated ( 

\vspace{-2mm}

\section{Mutual Information Estimation}
\label{SEC:MI_estimation}

%\vspace{-2mm}
We design a provably accurate estimator of $I(X;T_\ell)$ inspired by our recent work on differential entropy estimation \cite{anonymized_ISIT_estimation2019}. Given a feature dataset $\mathcal{X}=\{x_i\}_{i\in[n]}$ i.i.d.~according to $P_X$, our  $I(X;T_\ell)$ estimator is constructed from estimators of $h(T_\ell)$ and $h(p_{T_\ell|X=x}),\ \forall x\in\mathcal{X}$. We~propose a sampling method that reduces the estimation of each entropy to the framework from \cite{anonymized_ISIT_estimation2019}. Using entropy estimation risk bounds derived therein we control the error of our \emph{sample propagation} (SP) estimator $\hat{I}_{\mathsf{SP}}$ of $I(X;T_\ell)$. 

Each differential entropy is estimated and computed via a two-step process. First, we approximate each true entropy by the differential entropy of a \emph{known} Gaussian mixture (defined only through the available resources: the obtained samples and the noise parameter). This estimate converges to the true value at the \emph{parametric} estimation rate, uniformly in the dimension. However, since the Gaussian mixture entropy has no closed-form expression, in the second (computational) step we develop a Monte Carlo integration (MCI) method to numerically evaluate it. Mean squared error (MSE) bounds on the MCI computed value are established.

% %Using i.i.d. samples of $S_{\ell}$ and $S_{\ell}|X\mspace{-3mu}=\mspace{-3mu}x$, the SP estimator approximates  $h(p_{T_\ell})$ and $h(p_{T_\ell|X=x})$, respectively, 
%This estimator 

%the differential entropies are approximated via the SP estimator based on i.i.d. samples of $S_{\ell}$ and $S_{\ell}|X\mspace{-3mu}=\mspace{-3mu}x$. The obtained entropy proxies are
%The SP estimator approximates the true differential entropy as the differential entropy of a \emph{known} Gaussian mixture 

%\vspace{-3mm}

\subsection{Sampling Procedure and the Estimator}\label{SUBSEC:SP_estimator_summary}
%\vspace{-2mm}

% and set $g_\mathcal{A}(t)\triangleq\big(\hat{p}_\mathcal{A}\ast\varphi_\beta\big)(t)=\sum_{i\in[n]}\hat{p}_\mathcal{A}(a_i)\varphi_\beta(t-a_i)$ as a corresponding Gaussian mixture. \\

%\vspace{-5mm}
\textbf{Unconditional Entropy:} Since $T_\ell=S_\ell+Z_\ell$, where $S_\ell$ and $Z_\ell$ are independent, we have $p_{T_\ell}=p_{S_\ell}\ast \varphi_\beta$. To estimate $h(p_{T_\ell})$ feed each $x\in\mathcal{X}$ into the DNN and collect the output it produces at the $(\ell-1)$-th layer. The function $f_\ell$ is then applied on each collected output to obtain $\mathcal{S}_\ell\triangleq\{s_{\ell,i}\}_{i\in[n]}$, which is a set of $n$ i.i.d.~samples from $p_{S_\ell}$. Denoting by $\hat{p}_{\mathcal{A}}$ the empirical probability mass function of a set $\mathcal{A}=\{a_i\}_{i\in[n]}$, we estimate $h(p_{T_\ell})$ via $h(\hat{p}_{\mathcal{S}_\ell}\ast\varphi_\beta)$, which is the differential entropy of a Gaussian mixture with centers $\{s_{\ell,i}\}_{i\in[n]}$. %\footnote{No ``fitting" of a Gaussian mixture to the data or any data modeling assumptions are involved here; in fact, a separate mixture component is associated with each of the $n$ data points generated from the noisy network (with $n$ typically $> 10^6$).}. 
%The~term $h(\hat{p}_{\mathcal{S}_\ell}\ast\varphi_\beta)$ is referred to as the SP estimator of $h(p_{T_\ell})=h(p_{S_\ell}\ast\varphi_\beta)$.\\

%\vspace{-5mm}
\textbf{Conditional Entropies:} Fix $x\in\mathcal{X}$ and consider estimating $h(p_{T_\ell|X=x})$, where $p_{T_\ell|X=x}=p_{S_\ell|X=x}\ast\varphi_\beta$. We sample from $p_{S_\ell|X=x}$ by feeding $x$ into the DNN $n_x$ times, collecting $T_{\ell-1}$ outputs that correspond to different noise realizations, and applying $f_\ell$ on each. The obtained samples $\mathcal{S}_\ell^{(x)}\mspace{-6mu}\triangleq\mspace{-6mu}\big\{s_{\ell,i}^{(x)}\big\}_{i\in[n_x]}$ are i.i.d.~according~to $p_{S_\ell|X=x}$. Each $h(p_{T_\ell|X=x})$ is estimated by $h\big(\hat{p}_{\mathcal{S}_\ell^{(x)}}\ast\varphi_\beta\big)$.\footnote{For $\ell = 1$, we have $h(T_1|X)=h(Z_1)=\frac{d_1}{2}\log(2\pi e\beta^2)$.}\\
\textbf{Mutual Information Estimator:} We estimate $I(X;T_\ell)$ by
\vspace{-1mm}
\begin{equation}
    \hat I_{\mathsf{SP}}\triangleq h(\hat{p}_{\mathcal{S}_\ell}\ast\varphi_\beta)-\frac{1}{n}\sum\nolimits_{x\in\mathcal{X}}h\big(\hat{p}_{\mathcal{S}_\ell^{(x)}}\ast\varphi_\beta\big).\label{EQ:MI_estimator_final}
\end{equation}

\vspace{-2mm}
\subsection{Theoretical Guarantees and Computation}

%\vspace{-2mm}
This sampling procedure unifies the estimation of $h(p_{T_\ell})$ and $\big\{h(p_{T_\ell|X=x})\big\}_{x\in\mathcal{X}}$ into the problem of differential entropy estimation under Gaussian convolutions \cite{anonymized_ISIT_estimation2019}: estimate $h(p_S\ast\varphi_\beta)$ based on i.i.d.~samples $\mathcal{S}_n\triangleq\{S_i\}_{i\in[n]}$ from $p_S$ and knowledge of $\varphi_\beta$. Our $\hat I_{\mathsf{SP}}$ is inspired by the differential entropy estimator proposed in \cite{anonymized_ISIT_estimation2019}, which approximates $h(p_S\ast\varphi_\beta)$ by $h(\hat{p}_{\mathcal{S}_n}\ast\varphi_\beta)$. Before analyzing $\hat{I}_\mathsf{SP}$ performance, we note that its estimation is statistically difficult since any good estimator of $h(p_S\ast\varphi_\beta)$ using $\mathcal{S}_n$ and $\varphi_\beta$ requires exponentially many samples in $d$ (Theorem 1 of \citet{anonymized_ISIT_estimation2019}). Nonetheless, Theorem 2 therein shows that the absolute-error risk of $h(\hat{p}_{\mathcal{S}_n}\ast\varphi_\beta)$ converges at the best possible rate of $O\big(c^d\mspace{-3mu}/\mspace{-3mu}\sqrt{n}\big)$, for a constant $c$ and all $d$.

We now bound the estimation risk of $\hat{I}_\mathsf{SP}$ (the result is stated for bounded nonlinearities for simplicity; see Theorem~\ref{TM:sample_complex_bound} in the Appendix for corresponding ReLU results).

% Figure from Section 4 - Examples!
\begin{figure*}[t!]
    %\centering
    \setlength{\tabcolsep}{2.5pt}
    \begin{tabular}{cccc}
%\includegraphics[width=0.282\textwidth]{NIPS2018/Figures/Examples/FigTanhParam.pdf}%
%\begin{picture}(0,0)(0,0)\put(-100,67){\small({\bf{a}})}\end{picture}
%\includegraphics[width=0.3\textwidth]{NIPS2018/Figures/Examples/FigTanh_T250.pdf}%
%\begin{picture}(0,0)(0,0)\put(-102,66){\small({\bf{b}})}\end{picture} &
%\includegraphics[width=0.3\textwidth]{NIPS2018/Figures/Examples/FigTanh_T2500.pdf}%
%\begin{picture}(0,0)(0,0)\put(-102,66){\small({\bf{c}})}\end{picture} \\
%    \small{({\bf{a}})} & \small({\bf{b}}) & \small({\bf{c}})\\
\includegraphics[width=0.238\textwidth]{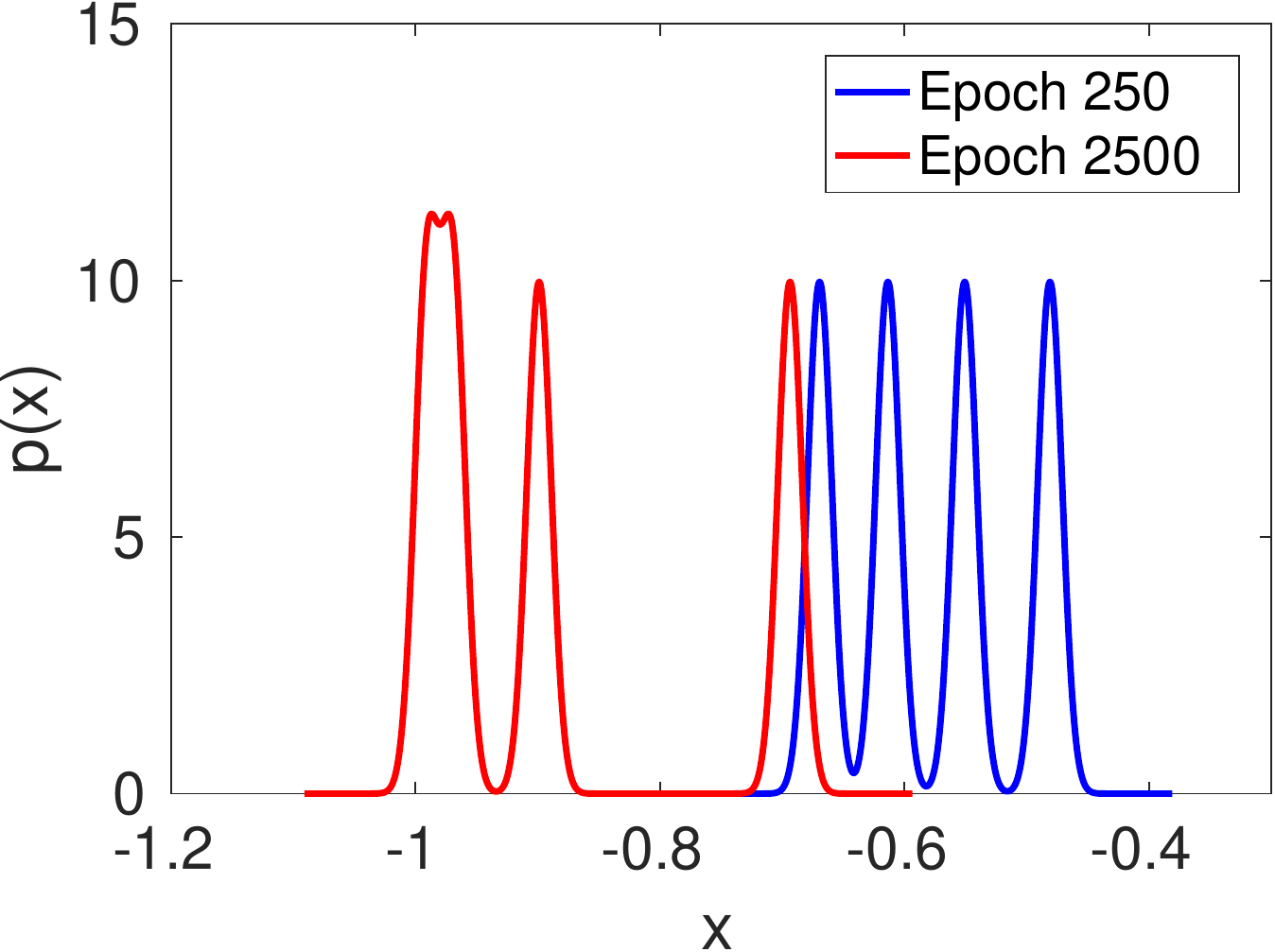}%
\begin{picture}(0,0)(0,0)\put(-97,75){\footnotesize({\bf{a}})}\end{picture}&
\includegraphics[width=0.240\textwidth]{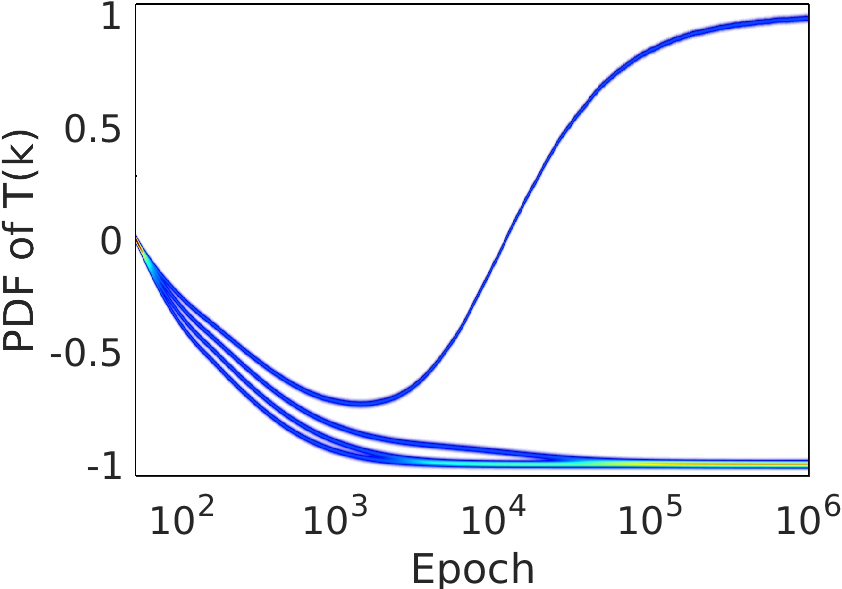}%
\begin{picture}(0,0)(0,0)\put(-96,73){\footnotesize({\bf{b}})}\end{picture}&
\includegraphics[width=0.238\textwidth]{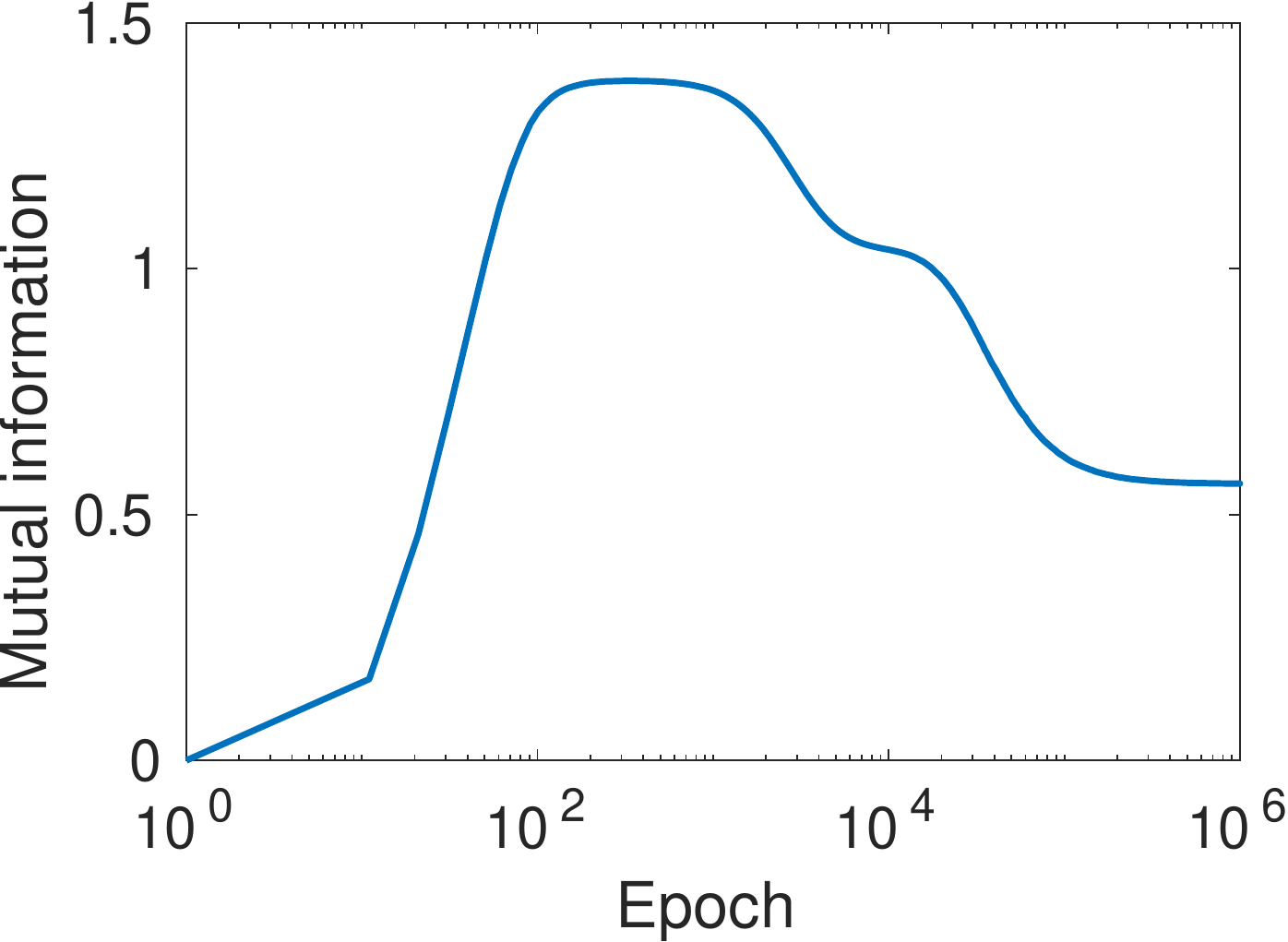}%
\begin{picture}(0,0)(0,0)\put(-96,73){\footnotesize({\bf{c}})}\end{picture} &
\includegraphics[width=0.236\textwidth]{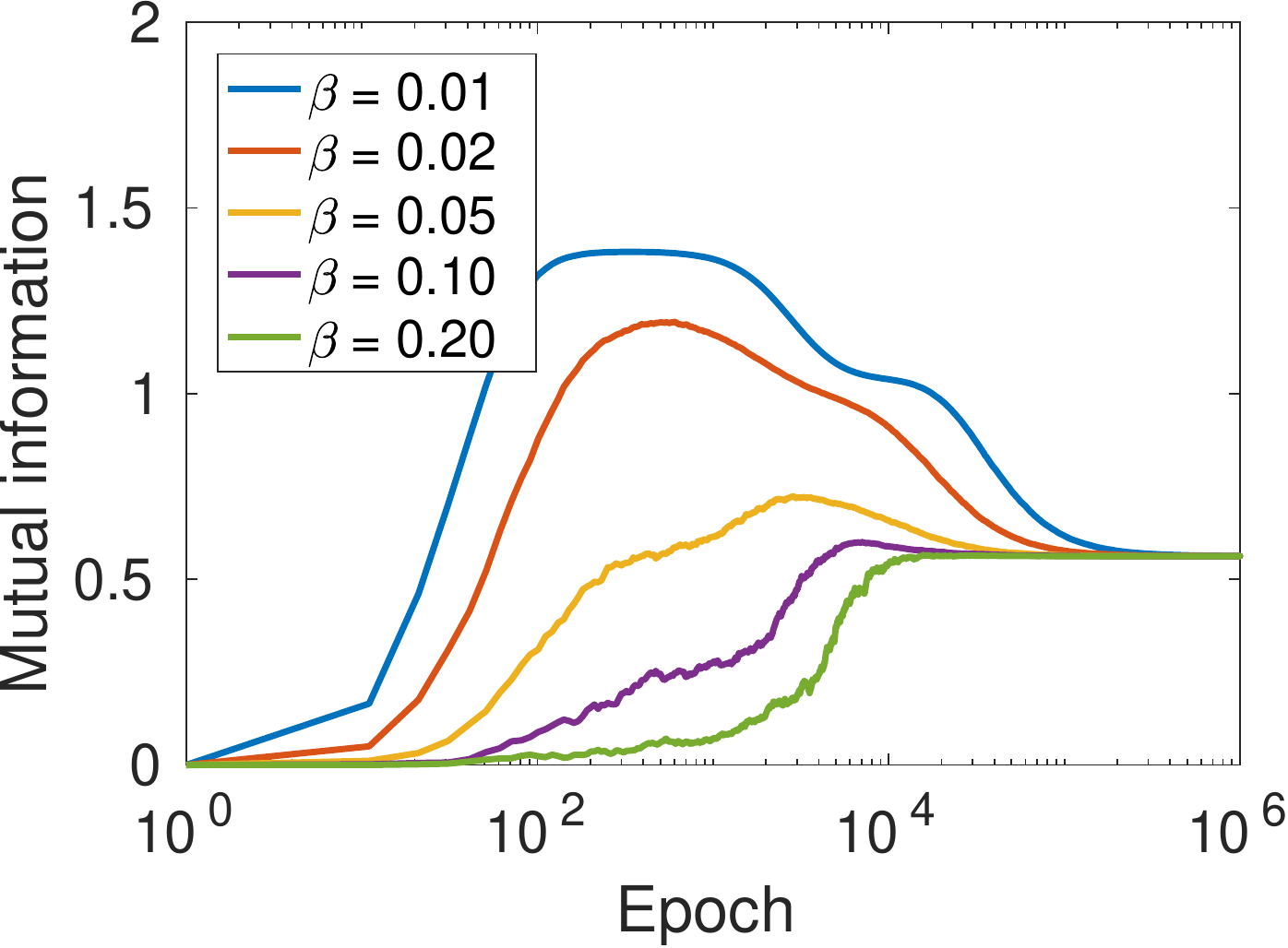}%
\begin{picture}(0,0)(0,0)\put(-19,73){\small({\bf{d}})}\end{picture}%
   \end{tabular}
   \vspace{-3mm}
    \caption{Single-layer tanh network: (a) the density $p_{T(k)}$ at epochs $k=250,2500$; (b) $p_{T(k)}$ and (c) $I\big(X; T(k)\big)$ as a function of $k$; and (d) mutual information as a function of $k$, for different $\beta$ values..}\label{Fig:OneLayerTanh}
    %\vspace{-2mm}
\end{figure*}
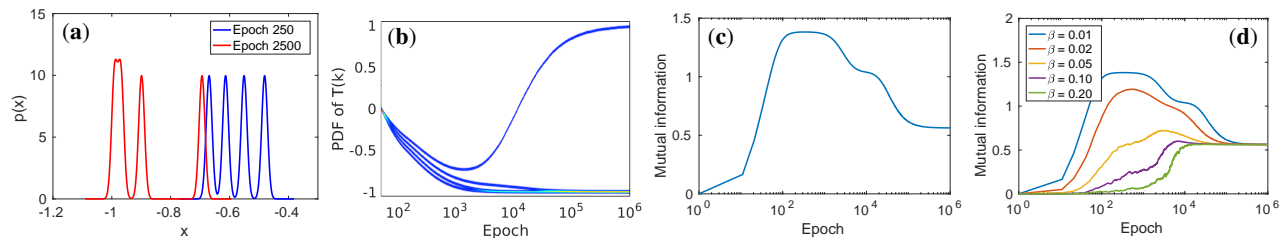

%\begin{theorem}[Theorem 2 of \cite{anonymized_ISIT_estimation2019}]\label{TM:SP_sample_complex_new}
%Fix $\beta>0$, $d\geq 1$, and let $\mathcal{F}_d$ be the class of $d$-dimensional PDFs supported inside $[-1,1]^d$. We have\ \ \ $\sup_{p_S\in\mathcal{F}_d}\mathbb{E}\big|h(p_S\ast\varphi_\beta)-h(\hat{p}_{S^n}\ast\varphi_\beta)\big|=
%O\big({c^d}/\sqrt{n}\big)$, for a numerical constant~$c$.
%\end{theorem}

%We obtain the following bound on the $\hat I_\mathsf{SP}$ estimation error.
\begin{theorem}\label{CORR:MI_True_Data_Dist}
Fix $\ell\in[L-1]$ and assume $\|f_\ell\|_\infty\leq 1$. For $\hat I_\mathsf{SP}$ from \eqref{EQ:MI_estimator_final} with $n=n_x$, for all $x\in\mathcal{X}$, we have
\begin{equation}
    \sup_{P_X}\mathbb{E} \left|I(X;T_\ell)-\hat{I}_{\mathsf{SP}}\right|\leq \frac{8c^{d_\ell}+d_\ell\log\left(1+\frac{1}{\sigma^2}\right)}{4\sqrt{n}},\label{EQ:MI_True_Data_Dist}
\end{equation}
where $c$ is a numerical constant explicitly given in Appendix \ref{SUPPSUBSEC:risk_bounds}.
\end{theorem}
Theorem \ref{CORR:MI_True_Data_Dist} is proven in Appendix \ref{SUPPSUBSEC:MI_True_Data_Dist_proof}. Interestingly, the quantity $\frac{1}{\sigma^2}$ is the signal-to-noise ratio (SNR) between $S$ and $Z$. The larger $\sigma$ is, the easier estimation becomes, since the noise smooths out the complicated $P_X$ distribution. %Also note that the dimension of the ambient space in which $X$ lies does not appear in the absolute-risk bound for estimating $I(X;T_\ell)$. The bound depends only on the dimension of $T$ (through $\Delta_{\sigma,d}$). This is because the additive noise resides in the $T$ domain, limiting the possibility of encoding the rich structure of $X$ into $T$ in full. On a technical level, the blurring effect caused by the noise enables uniformly lower bounding $\inf_x h(T|X=x)$ and thereby controlling the variance of the estimator for each conditional entropy. In turn, this reduces the impact of $X$ on the estimation of $I(X;T_\ell)$ to that of an empirical average converging to its expected value with rate $\frac{1}{\sqrt{n}}$.

Evaluating the SP estimator requires computing differential entropies of (known) Gaussian mixture distributions (see \eqref{EQ:MI_estimator_final}). Let $\hat{p}_{s^n}\ast\varphi_\beta$ be such a mixture and $G\sim\hat{p}_{s^n}\ast\varphi_\beta$. Noting that $h(\hat{p}_{s^n}\ast\varphi_\beta)=-\mathbb{E}\Big[\log\big((\hat{p}_{s^n}\ast\varphi_\beta)(G)\big)\Big]$, we numerically approximate the RHS via efficient MCI \citep{robert2004montecarlo}. Specifically, we generate $n_{\mathsf{MC}}$ i.i.d.~samples from $\hat{p}_{s^n}\mspace{-1mu}\ast\varphi_\beta$ and approximate the expectation~by an empirical average. This unbiased proxy achieves an MSE of $O\big((n\cdot n_{\mathsf{MC}})^{-1}\big)$ (Theorem~\ref{TM:MC_MSE} in Appendix C.4), and thus only adds a negligible error to the estimation process.\footnote{There are other ways to numerically evaluate $h(\hat{p}_{s^n}\ast\varphi_\beta)$, e.g., the Gaussian mixture bounds from \citet{kolchinsky2017estimating}; however, our proposed method is the fastest of which we are aware.}
 % We efficiently evaluate $\hat{p}_{s^n}\mspace{-3mu}\ast\mspace{-3mu}\varphi_\beta$ via kernel density estimation algorithms.

\textbf{Choosing $\bm{\beta}$ and $\bm{n}$:}
%Having established that the SP estimator has a satisfactory convergence rate, it is still necessary to
We describe practical guidelines for selecting $\beta$ and $n$ for estimating $I(X;T_{\ell})$ in actual classifiers. Ideally, $\beta$ is treated as a hyperparameter tuned to optimize the DNN's performance on held-out data, since internal noise serves as a regularizer similar to dropout. In practice, we sometimes back off from this optimal $\beta$ to a higher value to ensure accurate estimation of mutual information ($\hat{I}_{\mathsf{SP}}$ requires more samples for smaller $\beta$ values), depending on factors like the dimensionality of $T_\ell$ and the number of available data samples. 

While $n$ can be selected from the risk bound in \eqref{EQ:MI_True_Data_Dist}, this value can be quite large since Theorem~\ref{CORR:MI_True_Data_Dist} is a worst-case result. Furthermore, generating the estimated $I(X;T_\ell)$ curves shown in Section \ref{SEC:Empirical_results} requires repeatedly\footnote{Each $I(X;T_\ell)$, for a given set of DNN parameters, involves computing $n+1$ differential entropy estimates, and our experiments estimate the trajectory of $I(X;T_\ell)$ across training epochs.} running the differential entropy estimator, which makes the $n$ dictated by Theorem~\ref{CORR:MI_True_Data_Dist} infeasible. To overcome this computational burden while adhering to the theory, we tested the value of $n$ given by the theorem on a few points of each curve and reduced it until the overall computation cost became reasonable. To ensure estimation accuracy was not compromised we empirically tested that the estimate remained stable.

As a concrete example, for an error bound of 10\% of Fig. \ref{Fig:Tishby_exp} plot's vertical scale ($\approx 0.8$ absolute error) Theorem \ref{CORR:MI_True_Data_Dist} requires $n=4\cdot10^9$ samples, which is beyond our computational budget. Performing the above procedure to lower $n$, we find good accuracy is achieved for $n = 4\cdot10^6$. Adding more samples beyond this value does not change the results.

%Instead, we use Theorem~\ref{TM:SP_sample_complex_new} to set a maximum on $n$, and then test smaller values of $n$ to determine at what point the estimate of $I(X;T_{\ell})$ diverges from the correct one. Choosing a smaller $n$ that still produces accurate estimates of $I(X;T_{\ell})$ makes the experiments in Section~\ref{SEC:Empirical_results} computationally feasible.}

\section{Compression and Clustering}
\label{SEC:minimal_example} %Coalescing Gaussians

We use a minimal example to connect compression and clustering via an information-theoretic perspective. Consider one noisy neuron with a scalar input $X$. Let $T(k)=S(k)+Z=\sigma(w_kX+b_k)+Z$ be the neuron's output at epoch $k$, where $\sigma$ is strictly monotone  and $Z\sim\mathcal{N}(0,\beta^2)$. Invariance of $I\big(X;T(k)\big)$ to invertible operations implies $I\big(X;T(k)\big)=I\big(S(k);S(k)+Z\big)$. %\label{EQ:MI_minimal_example_expansion}
%which uses the invariance of mutual information to invertible operations and the definition of $S(k)$.
%Assume $X\sim\mathsf{Unif}(\mathcal{X})$, with the exact data set $\mathcal{X}\subset\mathbb{R}$ and corresponding classes to be specified later. 
%We track the mutual information between $X$ and $T$ as it evolves over training epochs. Epochs are indexed by $k\in[K]$ and we write $T(k)=S(k)+Z$ for the output neuron activity with $S(k) \triangleq \sigma(w_kX+b_k)$ and $Z\sim\mathcal{N}(0,\beta^2)$. For the mutual information we have
%\begin{equation}
%I\big(X;T(k)\big)\stackrel{(a)}= I\big(\sigma(w_kX+b_k);\sigma(w_kX+b_k)+Z\big)=I\big(S(k);S(k)+Z\big),\label{EQ:MI_minimal_example_expansion}
%\end{equation}
%where (a) uses the invariance of mutual information to invertible operations, and the definition of $S(k)$.
%you here?
Information-theoretically, $I\big(S(k);S(k)+Z\big)$ is the aggregate number of nats transmitted over an AWGN channel with input constellation $\mathcal{S}_k\triangleq\big\{\sigma(w_kx+b_k)\mid {x\in\mathcal{X}}\big\}$. In~other words, $I\big(S(k);S(k)+Z\big)$ measures how distinguishable the symbols of $\mathcal{S}_k$ are when composed with Gaussian noise (roughly equaling the $\log$ of the number of resolvable clusters under noise level $\beta$). Since the distribution of $T(k)$ is a Gaussian mixture with means $s\in\mathcal{S}_k$, the closer two constellation points $s$ and $s'$ are, the more the Gaussians centered on them overlap. %hence the closer two constellation points are, the more their Gaussians at overlap at the output.
%This makes the symbols more difficult to distinguish at the output, reducing $I\big(S(k);S(k)+Z\big)$. 
Thus reducing point spacing in $\mathcal{S}_k$ (by changing $w_k$ and $b_k$) directly reduces $I\big(X;T(k)\big)$.
% and vice versa.
%these constellation points if either is transmitted, which decreases mutual information. 
%Thus, to understand the fluctuations of $I\big(X;T(k)\big)$ it suffices to study the motion of Gaussians at the output as $k\in[K]$ increases.
Let $\sigma = \tanh$, $\beta = 0.01$ and $\mathcal{X} = \mathcal{X}_{-1} \cup \mathcal{X}_1$, with $\mathcal{X}_{-1} = \{-3,-1,1\}$ and $\mathcal{X}_1 = \{3\}$, labeled $-1$ and $1$, respectively. We train the neuron using mean squared loss and gradient descent with learning rate $0.01$ to illustrate $I\big(X;T(k)\big)$ trends. The Gaussian mixture $p_{T(k)}$ is plotted across epochs $k$ in Fig.~\ref{Fig:OneLayerTanh}(a)-(b). The learned bias is approximately $-2.3w$, ensuring that the tanh transition region correctly divides the two classes. %is centered at $x\approx2.3$ between the boundaries of the two classes. 
Initially $w=0$, so all four Gaussians in $p_{T(0)}$ are superimposed. As $k$ increases, the Gaussians initially diverge, with the three from $\mathcal{X}_{-1}$ eventually re-converging as they each meet the tanh boundary. This is reflected in the mutual information trend in Fig.~\ref{Fig:OneLayerTanh}(c), with the dips in $I\big(X;T(k)\big)$ around $k = 10^3$ and $k = 10^4$ corresponding to the second and third Gaussians respectively merging into the first. Thus, there is a direct connection between clustering and compression. %, with the tanh saturation playing a central role in the clustering process. 
%
%As the Gaussians spread out the overlap between them reduces and $I\big(X;T(k)\big)$ grows.
%The first dip in $I\big(X;T(k)\big)$ around $k=10^3$ occurs as the first two Gaussians start to coalesce, and a second dip at $k=10^4$ corresponds to the third Gaussian merging in. %, causing a decrease in mutual information until they are nearly completely merged. 
%After epoch $10^4$ the third Gaussian starts to merge with the bundle causing another decrease in $I\big(X;T(k)\big)$. 
%
Fig.~\ref{Fig:OneLayerTanh}(d) shows $I\big(X;T(k)\big)$ for different noise levels $\beta$ as a function of $k$. For small $\beta$ (as above) the $\mathcal{X}_{-1}$ Gaussians are distinct and merge in two stages as $w$ grows. For larger $\beta$, however, the $\mathcal{X}_{-1}$ Gaussians are indistinguishable for any $w$, making $I\big(X;T(k)\big)$ increase as the two classes gradually separate. Similar trends are observed for a two-neuron leaky-ReLU network in Appendix A.

%As $\beta$ increases the lower two peaks become less distinguishable and gradually merge into a single peak. 

%\caption{Single-layer tanh network: (a) the density $f_{T(k)}$ at epochs $k=250,2500$; (d) $f_{T(k)}$ and (e) $I\big(X; T(k)\big)$ as a function of $k$; and (f) mutual information as a function of weight $w$ with bias $-2w$.}\label{Fig:OneLayerTanh}

% Compression 0.6: 53, 52, 52, 52

\section{Empirical Results}
\label{SEC:Empirical_results}

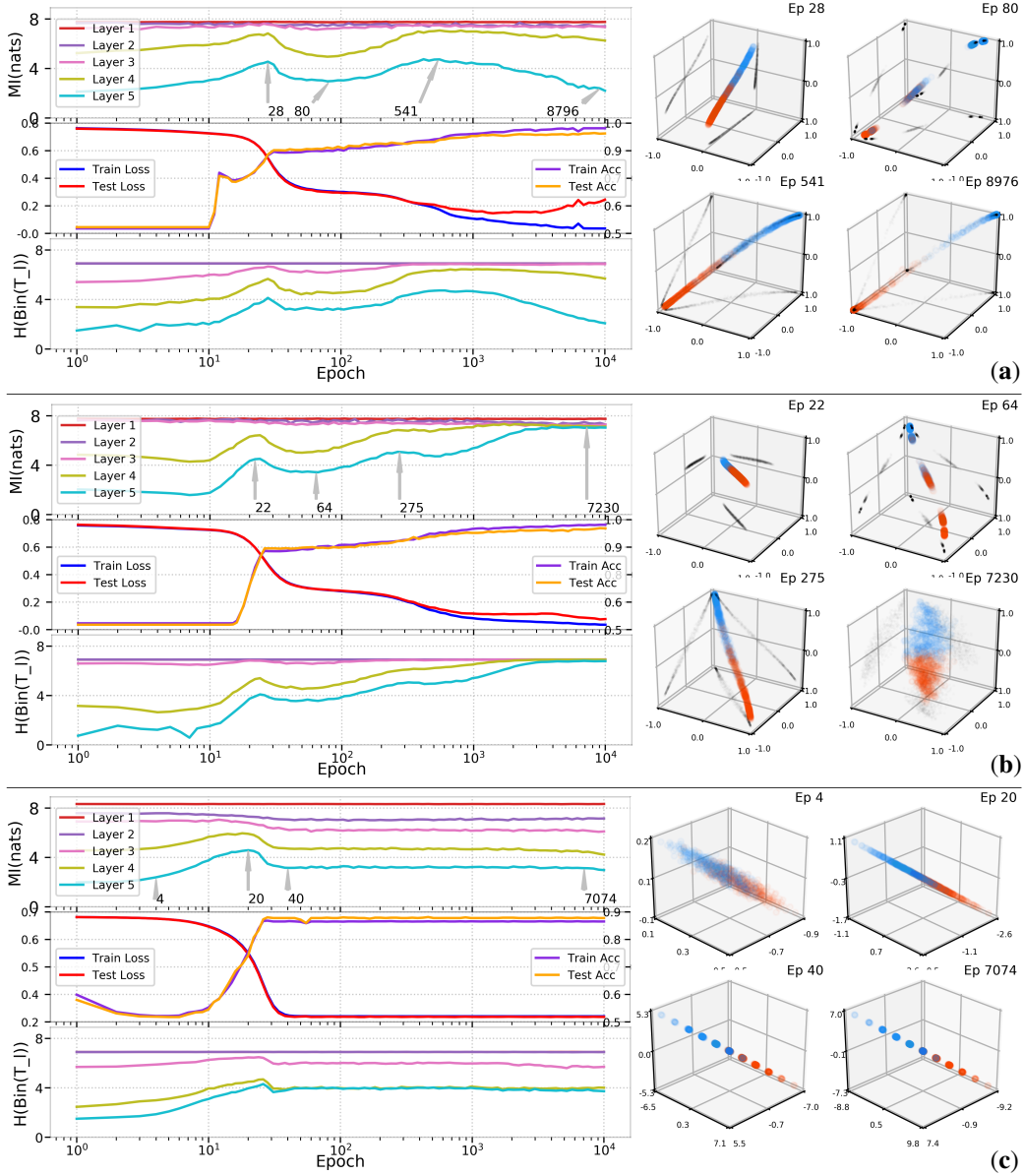
\begin{figure*}[t!]
    \centering
    \setlength{\tabcolsep}{0pt}
    %\begin{tabular}{c}
    %\includegraphics[width=\textwidth]{NIPS2018/Figures/Experiments/Tishby/large_exprs/Fig5a.png}%\\
    %\begin{picture}(0,0)(0,0)\put(-12,0){({\bf{a}})}\end{picture}\\
    %\hline
    %\includegraphics[width=\textwidth]{NIPS2018/Figures/Experiments/Tishby/large_exprs/Fig5b.png}%\\
    %\begin{picture}(0,0)(0,0)\put(-12,0){({\bf{b}})}\end{picture}\\
    %\hline
    %\includegraphics[width=\textwidth]{NIPS2018/Figures/Experiments/Tishby/large_exprs/Fig5c.png}%\\
    %\begin{picture}(0,0)(0,0)\put(-12,0){({\bf{c}})}\end{picture}
    %\end{tabular}
    %
    %
    \begin{tabular}{c}
    \includegraphics[width=0.8\textwidth]{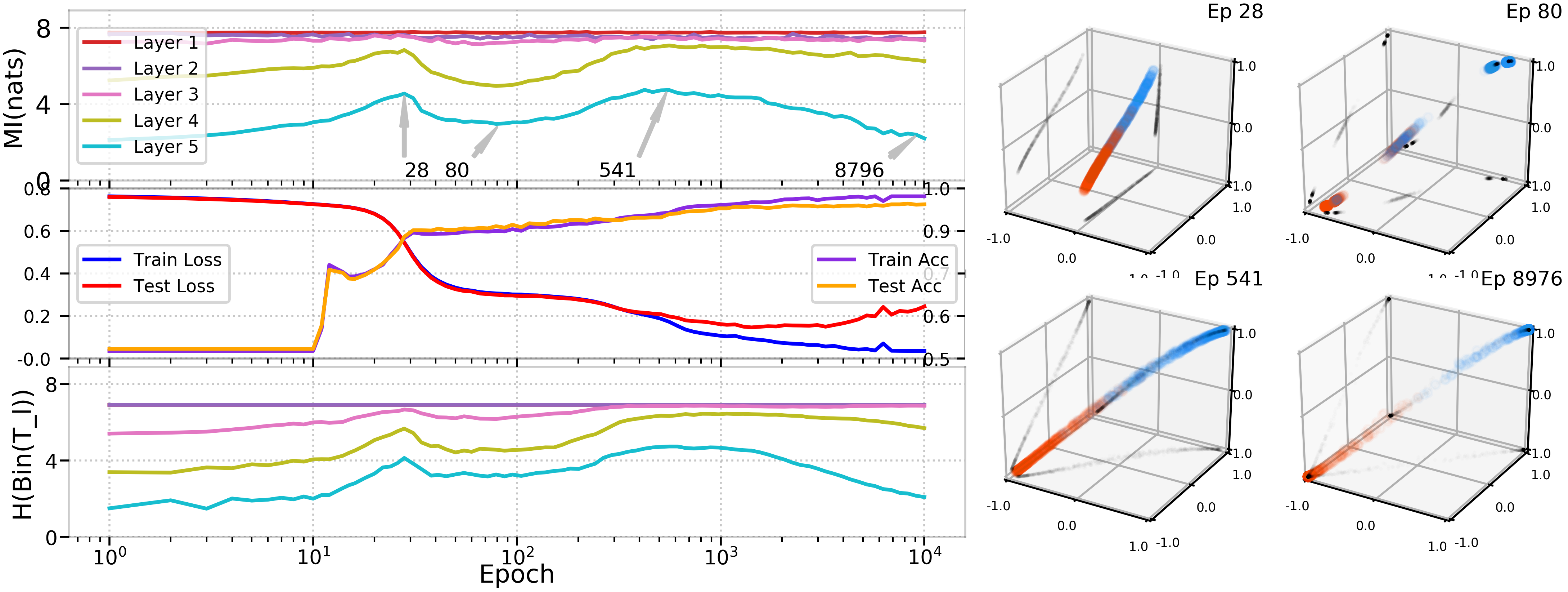}%
    \begin{picture}(0,0)(0,0)\put(-12,2){({\bf{a}})}\end{picture}\\
    \hline
    \includegraphics[width=0.8\textwidth]{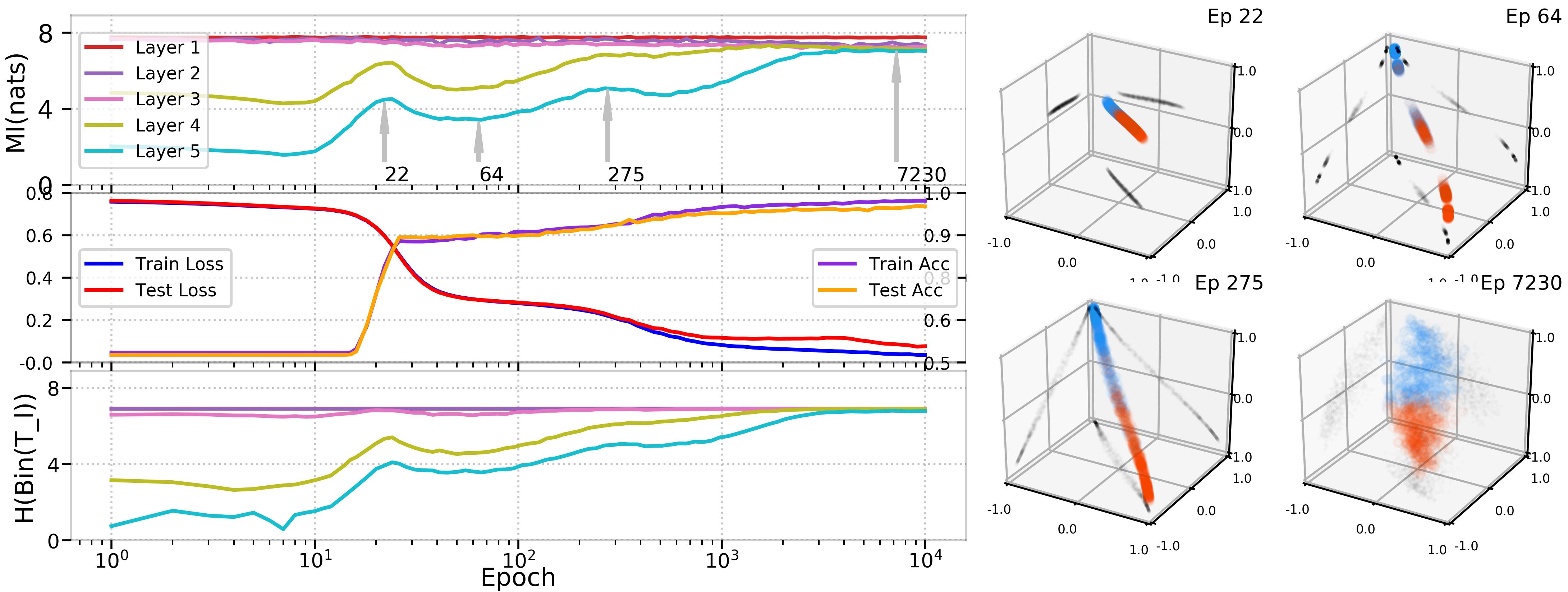}%
    \begin{picture}(0,0)(0,0)\put(-12,2){({\bf{b}})}\end{picture}\\
    \hline
    \includegraphics[width=0.8\textwidth]{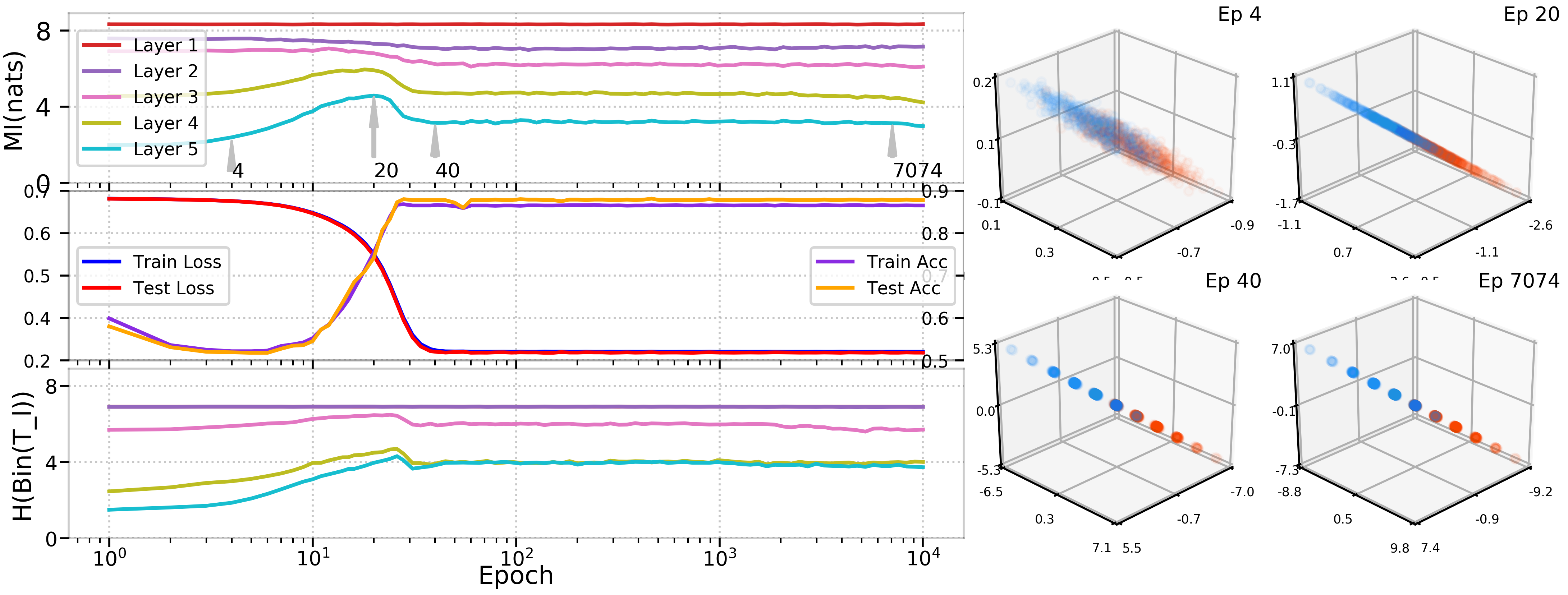}%
    \begin{picture}(0,0)(0,0)\put(-12,2){({\bf{c}})}\end{picture}\\
    \end{tabular}
    \vspace{-1mm}
\caption{(a) Evolution of $I(X;T_\ell)$ and training/test losses across training epochs for the SZT model~with $\beta= 0.005$ and tanh nonlinearities. The scatter plots show the values of Layer 5 ($d_5\mspace{-2mu}=\mspace{-2mu} 3$) at the arrow-marked epochs on the mutual information plot. The bottom plot shows $H\big(\mathsf{Bin}(T_\ell)\big)$ across epochs for bin size $B\mspace{-3mu}=\mspace{-3mu}10\beta$. (b) Same setup as in (a) but with regularization that encourages orthonormal weight matrices. (c) SZT model with $\beta= 0.01$ and \emph{linear}~activations. }
    \label{Fig:Tishby_exp}
    %\vspace{-2mm}
\end{figure*}

We show that the observations from Section \ref{SEC:minimal_example} hold for two larger networks. The experiments demonstrate that $I(X;T_\ell)$ compression in noisy DNNs is driven by clustering of internal representations, and that deterministic DNNs cluster samples as well. %(despite $I(X;T_\ell)$ being vacuous there). 
The considered DNNs are
\begin{inparaenum}[(1)]
\item the fully connected network (FCN) from~\citep{DNNs_Tishby2017,DNNs_ICLR2018}, dubbed the \emph{SZT model}, and
\item a convolutional network for MNIST classification, called \emph{MNIST CNN}.
\end{inparaenum} We present selected results; see the appendices for additional experiments.

\begin{figure*}
    \centering
    \setlength{\tabcolsep}{0pt}
    \begin{tabular}{ccc}
    \includegraphics[width=0.413\textwidth,height=4.7cm]{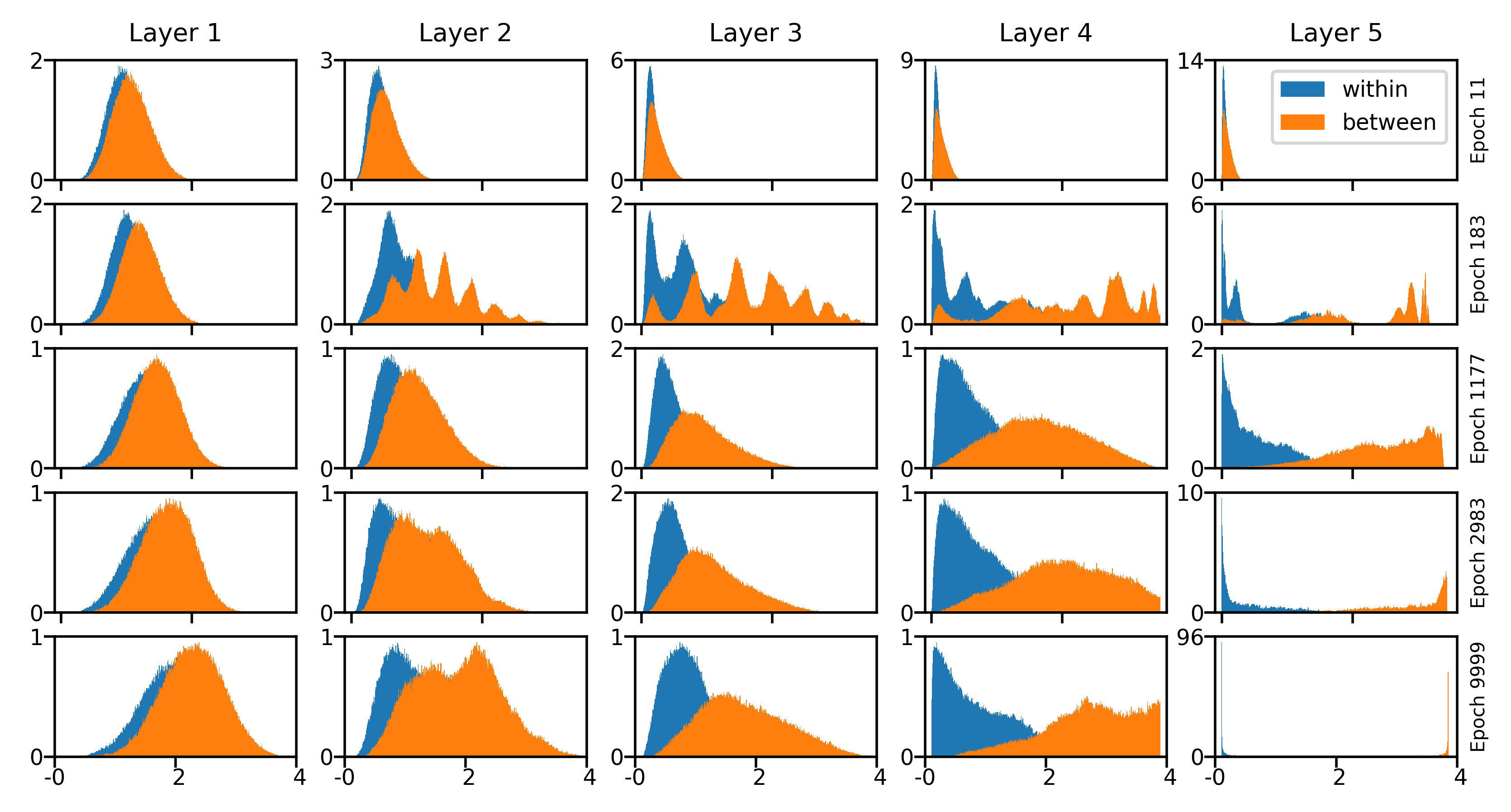} &
    \hspace*{4em} &
    \includegraphics[width=0.413\textwidth,height=4.7cm]{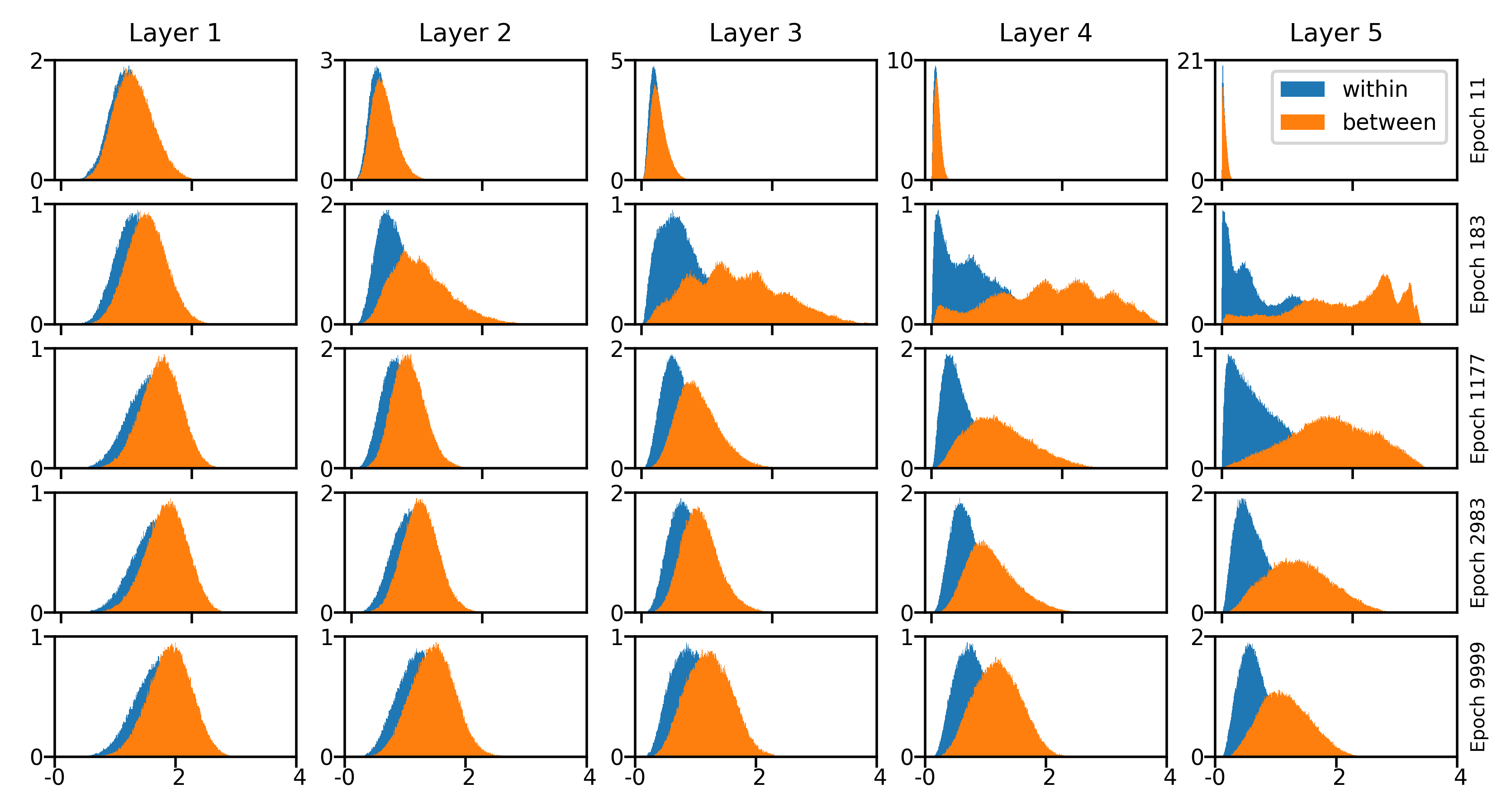}\\[-6pt]
    %\begin{picture}(0,0)(0,0)\put(-380,-1){({\bf{a}})}\end{picture}
    %\begin{picture}(0,0)(0,0)\put(-130,-1){({\bf{b}})}\end{picture}
    ({\bf{a}}) && ({\bf{b}})\\[-8pt]
    \end{tabular}
\caption{(a) Histogram of within- and between-class pairwise distances for SZT model with tanh non-linearities and additive noise $\beta= 0.005$. (b) Same as (a) but training with weight normalization.}
\vspace{-3mm}
\label{Fig:tishby_hist}
%\vspace{-6mm}
\end{figure*}
%}

\subsection{Noisy SZT Model}
\label{SUBSEC:SZT}
Consider the data and model of~\cite{DNNs_Tishby2017} for binary classification of 12-dimensional inputs using a fully connected 12--10--7--5--4--3--2 architecture. The FCN was tested with tanh and ReLU nonlinearities as well as a linear model. 
%The noise parameter $\beta$ was set to maximize accuracy on a held-out set. 
%Training used the Adam~\cite{Adam2014} optimizer with batch size 256 and learning rate $4.0 \times 10^{-4}$. 
Fig.~\ref{Fig:Tishby_exp}(a) presents results for the tanh SZT model with $\beta=0.005$ (test classification accuracy 97\%), showing the relation across training epochs between the estimated $I(X;T_\ell)$, train/test losses and the evolving internal representations. The rise and fall of $I(X;T_\ell)$ corresponds to how spread out or clustered the representations in each layer are. For example, $I(X;T_5)$ grows until epoch 28, when the Gaussians start to move away from each other along a curve (see scatter plots on the right). Around epoch 80 they start clustering and $I(X;T_5)$ drops. As training progresses, the saturating tanh units push the Gaussians to opposite corners of the cube, thereby reducing $I(X;T_5)$ even more.
% [EvdB-original:]push the Gaussians to two furthers corners of the cube, recuding $(I(X;T_5)$ even more

To confirm that clustering (via saturation) was central to the compression observed in Fig.~\ref{Fig:Tishby_exp}(a), we also trained the model using the regularization from~\citep{Parseval2017} (test classification accuracy 98\%), which encourages orthonormal weight matrices. The results are shown in Fig.~\ref{Fig:Tishby_exp}(b). Apart from minor initial fluctuations, compression is completely gone. The scatter plots show that the vast majority of neurons do not saturate and no clustering is observed at the later stages of training. Saturation is not the only mechanism that can cause clustering and consequently reduce $I(X;T_\ell)$. For example, in Fig.~\ref{Fig:Tishby_exp}(c) we illustrate the clustering behavior in a linear SZT model (test classification accuracy 89\%). As seen from the scatter plots, due to the formation of several clusters and projection to a lower dimensional space, $I(X;T_\ell)$ drops even without the nonlinearities.  

The results in Fig.~\ref{Fig:Tishby_exp}(a)~and~(b) also show that the relationship between compression and generalization performance is not a simple one. In Fig.~\ref{Fig:Tishby_exp}(a), the test loss begins to increase at roughly epoch 3200 and continues to grow until training ends. At the same time compression occurs in layers 4 and 5. In contrast, the test loss in Fig.~\ref{Fig:Tishby_exp}(b) does not increase, and compression does not occur. We believe this subject deserves further examination in future work.

%To illustrate the clustering phenomenon further,
To provide another perspective on clustering that is sensitive to class membership, we compute histograms of pairwise distances between representations of samples, distinguishing within-class distances from between-class distances. Fig.~\ref{Fig:tishby_hist} shows histograms for the SZT models from Figs. \ref{Fig:Tishby_exp}(a) and (b). %in  we also show the histogram of within-class and between-class pairwise distance for SZT model corresponding to the setups of Fig. \ref{Fig:Tishby_exp}(a) and (b).
As training progresses, the formation of clusters is clearly seen (layer 3 and beyond) for the unnormalized SZT model in Fig.~\ref{Fig:Tishby_exp}(a). In the normalized model (Fig.~\ref{Fig:Tishby_exp}(b)), however, no tight clustering is apparent, supporting the connection between clustering and compression.

Having identified clustering as the source of compression, we focus on it as the point of interest. To measure clustering we consider the discrete entropy $H\big(\mathsf{Bin}(T_\ell)\big)$, with the  number of equal-sized bins, $B$, as a tuning parameter. Note that $\mathsf{Bin}(T_\ell)$ partitions the dynamic range (e.g., $[-1,1]^{d_\ell}$ for a tanh layer) into $B^{d_\ell}$ cells or bins. When hidden representations are spread out, many bins are non-empty, each holding a positive probability mass. Conversely, for clustered representations, the distribution is concentrated on a few bins, each with relatively high probability. Recalling that the uniform distribution maximizes discrete entropy, we see why reduction in $H\big(\mathsf{Bin}(T_\ell)\big)$ measures clustering.% by attaining smaller values for increasingly clustered constellations.

To illustrate, the bottom plots in Figs.~\ref{Fig:Tishby_exp}(a), (b) and (c) show $H\big(\mathsf{Bin}(T_\ell)\big)$ for each SZT model using $B = 10\beta$. Its values differ from those of $I(X; T_{\ell})$, suggesting $H\big(\mathsf{Bin}(T_\ell)\big)$ is formally not an estimator of the mutual information. Nonetheless, a clear correspondence between the trends of $H\big(\mathsf{Bin}(T_\ell)\big)$ and $I(X; T_{\ell})$ is seen, indicating that $H\big(\mathsf{Bin}(T_\ell)\big)$ measures clustering in a manner similar to $I(X;T_{\ell})$. This is important when moving back to \textit{deterministic DNNs}, where $I(X;T_{\ell})$ is no longer informative (being independent of the system parameters). %either a constant or infinity, for discrete or continuous $X$, respectively. 
Fig.~\ref{Fig:Binning_tanh} shows $H\big(\mathsf{Bin}(T_\ell)\big)$ for the deterministic SZT  model ($\beta = 0$). The bin size is a free parameter, and depending on its value, $H\big(\mathsf{Bin}(T_\ell)\big)$ reveals different clustering granularities. Moreover, since in deterministic networks $T_\ell=f_\ell(X)$, for a deterministic map $f_\ell$ we have $I\big(X;\mathsf{Bin}(T_\ell)\big) = H\big(\mathsf{Bin}(T_\ell)\big)$. Thus, the plots from \citep{DNNs_Tishby2017}, \citep{DNNs_ICLR2018}, and our Fig.~\ref{Fig:Binning_tanh} all show that $H\big(\mathsf{Bin}(T_\ell)\big)$ evolution during training of deterministic DNNs \textit{tracks the degree of clustering in the internal representation spaces}, rather than the theoretically impossible compression of ${I(X;T_\ell)}$.

\subsection{Deterministic MNIST CNN}
\label{SUBSEC:MNIST}
\begin{figure}	\centering
\includegraphics[width=0.4\textwidth]{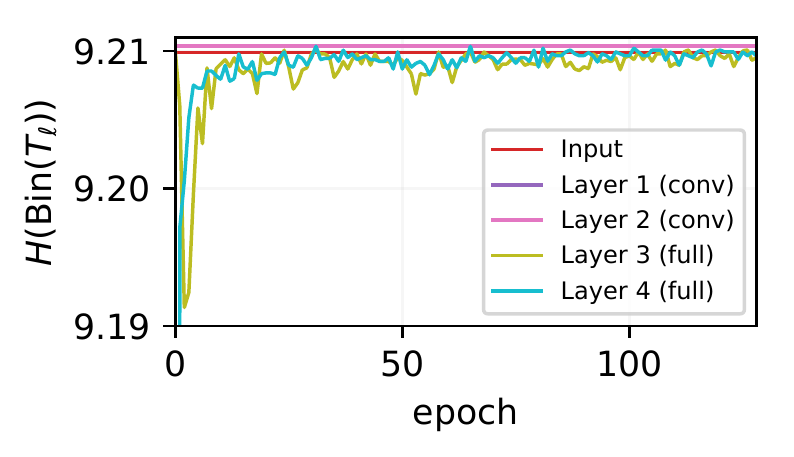}
%\vspace{-7mm}
\vspace{-3mm}
\caption{$H\big(\mathsf{Bin}(T_\ell)\big)$ for the MNIST CNN, computed using two bins: $[-1, 0]$ and $(0,1]$. The tiny range of the y axis shows the near injectivity of the model.}\label{fig:MNIST-CNN-binnedMI}
\vspace{-4mm}
\end{figure}

We also examine a deterministic model that is more representative of current machine learning practice: the MNIST CNN trained with dropout from Section~\ref{SEC:Noisy_DNNs}. Fig.~\ref{fig:MNIST-CNN-binnedMI} portrays the near-injective behavior of this model. Even when only two bins are used to compute $H\big(\mathsf{Bin}(T_\ell)\big)$, it takes values that are approximately $\ln(10000) = 9.210$, for all layers and training epochs, even though the two convolutional layers use max-pooling.  %This binning merges two samples in the validation set, so the input has $H\big(\mathsf{Bin}(T_\ell)\big) = 9.209$. 
While Fig.~\ref{fig:MNIST-CNN-binnedMI} does not show compression at the level of entire layers, computing $H\big(\mathsf{Bin}(T_\ell(k))\big)$ for individual units $k$ in layer 3 reveals a gradual decrease over epochs 1--128. To quantify this trend, we computed linear regressions predicting $H\big(\mathsf{Bin}(T_\ell(k))\big)$ from the epoch index, for all units $k$ in layer 3, and determined the mean and standard deviation of the slope of the linear predictions. If most slopes are negative, then compression occurs during training at the level of individual units. For a range of bin sizes from $10^{-4}$--$10^{-1}$ the least negative mean slope was $-0.002$ nats/epoch with a maximum standard deviation of $0.001$, showing that most units undergo compression.

In Fig.~\ref{fig:MNIST-CNN-cond-pairwise} we show histograms of pairwise distances between MNIST validation set samples in the input (pixel) space and in the four layers of the CNN. The histograms were computed for epochs 0, 1, 32, and 128, where epoch 0 is the initial random weights and epoch 128 is the final weights.  The histogram for the input shows that the mode of within-class pairwise distances is lower than the mode of between-class pairwise distances, but that there is substantial overlap. Layers 1 and 2, which are convolutional and therefore do not contain any units that receive the full input, do little to reduce this overlap, suggesting that the features learned in these layers are somewhat generic. In contrast, even after one epoch of training, layers 3 and 4, which are fully connected, separate the distribution of within-class distances from the distribution of between-class distances.

\subsection{Summary of Experiments}
\label{SUBSEC:SUMMARY}
\begin{figure}
  \centering
  \includegraphics[width=\columnwidth]{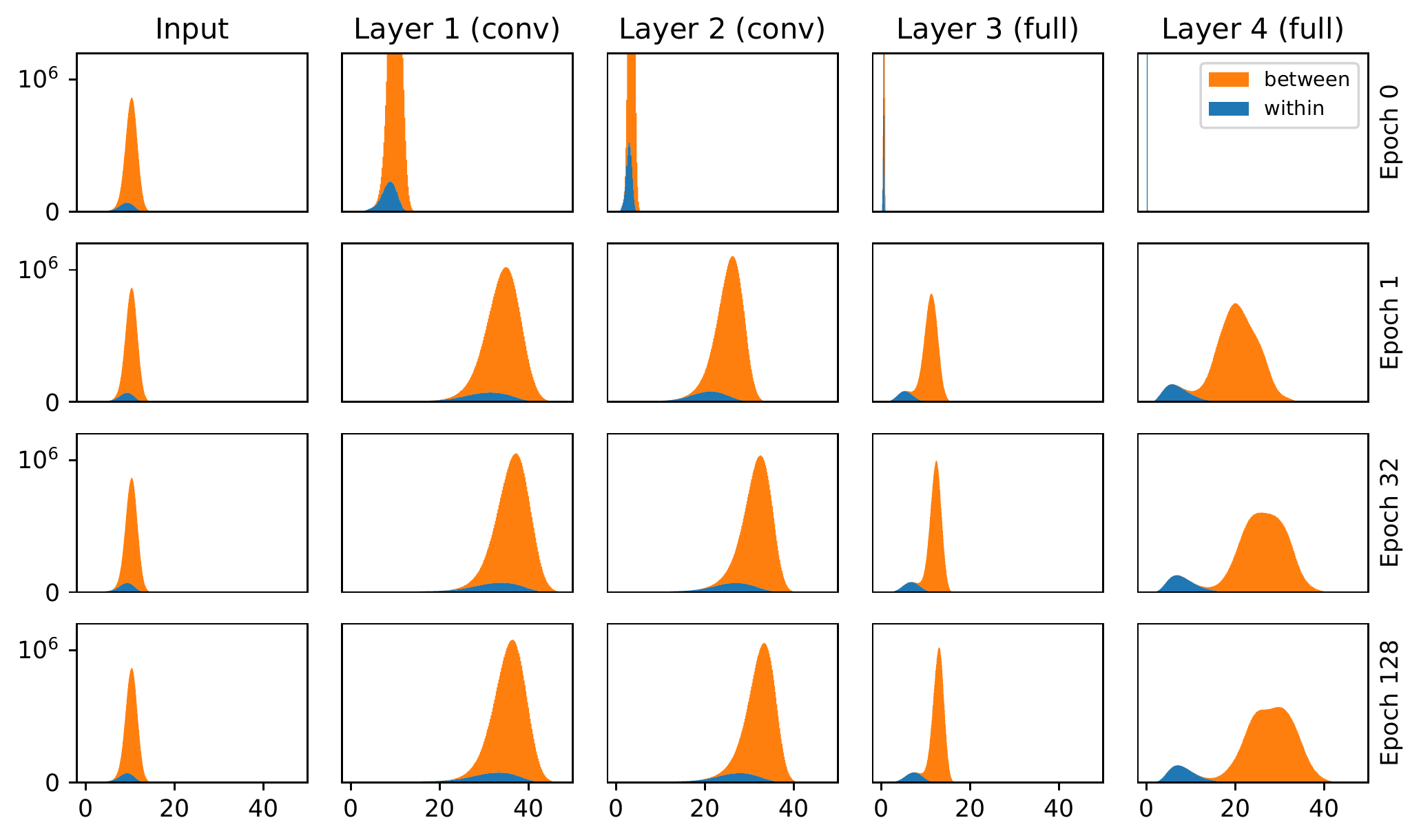}
  \vspace{-6mm}
  \caption{Histograms of within-class and between-class pairwise distances from the MNIST CNN.}% trained with dropout.}
  \label{fig:MNIST-CNN-cond-pairwise}
  \vspace{-3mm}
\end{figure}

We made the following observations in our experiments.
\begin{inparaenum}[(1)] \item Compression can occur in noisy networks, e.g., the noisy SZT model inspired by the deterministic network from \citep{DNNs_Tishby2017}, for which compression was first observed (upper left plot in Fig.~\ref{Fig:Tishby_exp}(a)). 
\item Compression is caused by clustering of internal representations, with clusters comprising mostly samples from the same class, as seen in the scatter plots on the right sides of Figs.~\ref{Fig:Tishby_exp}(a) and (c), and the distributions of pairwise distances in Figs.~\ref{Fig:tishby_hist} and \ref{fig:MNIST-CNN-cond-pairwise}.
\item Regularization that limits the network's ability to saturate hidden units can suppress the formation of tight clusters in the internal representation spaces and eliminate compression (Fig.~\ref{Fig:Tishby_exp}(b)). Observing that the regularized network from Fig.~\ref{Fig:Tishby_exp}(b) (where no compression occurs) generalizes better than the unregularized version in Fig.~\ref{Fig:Tishby_exp}(a), indicates that $I(X;T_\ell)$ \emph{compression and generalization may not be causally related}. This relation warrants further investigation. 
\item Fig.~\ref{Fig:Tishby_exp} demonstrated that $I(X;T_\ell)$ and $H\big(\mathsf{Bin}(T_\ell)\big)$ are highly correlated, establishing the latter as another measure for clustering (applicable both in noisy and deterministic DNNs). 
\item Clustering is also observed in a convolutional network trained on MNIST. While Fig.~\ref{fig:MNIST-CNN-binnedMI} shows that due to the high dimensionality of the CNN, $H\big(\mathsf{Bin}(T_\ell)\big)$ fails to track compression, strong evidence for clustering is found via estimates done at the individual units level and the analysis of pairwise distances between samples shown in Fig.~\ref{fig:MNIST-CNN-cond-pairwise}.

%[EvdB-Original:]\item Clustering of internal representations can also be observed in a somewhat larger, convolutional network trained on MNIST. While Fig.~\ref{fig:MNIST-CNN-binnedMI} shows that due to the high dimensionality, $H\big(\mathsf{Bin}(T_\ell)\big)$ fails to track compression in the larger CNN, strong evidence for clustering is found via estimates done at the individual units level and the analysis of pairwise distances between samples shown in Fig.~\ref{fig:MNIST-CNN-cond-pairwise}.
\end{inparaenum}

\section{Conclusions and Future Work}

This work studied the mutual information $I(X;T_\ell)$ in a DNN. We reexamined the `compression' aspect of the 
Information Bottleneck theory \citep{DNNs_Tishby2017},
noting that fluctuations of $I(X;T_\ell)$ in deterministic networks with strictly monotone nonlinearities are impossible. Aiming to test for $I(X;T_\ell)$ compression rigorously and discover its source, we:
\begin{inparaenum}[(1)]
\item created a rigorous framework for studying and accurately estimating information-theoretic quantities in DNNs whose weights are fixed;
%, and developed a toolkit for mutual information estimation in this framework;
\item identified clustering of the learned representations as the geometric phenomenon underlying compression; and
\item demonstrated that the compression-related experiments on deterministic networks in prior works were in fact measuring this clustering through the lens of binned mutual information.
\end{inparaenum}

We thus identify clustering as the common phenomenon of interest, which happens in both deterministic and noisy networks. In contrast, the mutual information $I(X;T_\ell)$ is meaningful only if a mechanism for shedding information (e.g., noise) exists in the network, and even then, it merely tracks the same clustering effect. Although binning-based measures do not accurately estimate mutual information, they are simple to compute (as opposed to the exponential in dimension burden of mutual information estimation) and are useful for tracking changes in clustering. We believe that further study of this geometric phenomenon is warranted to gain more insight into the learned representations and potentially establish connections with generalization. To this end we are currently developing efficient methods to track clustering in high-dimensional spaces. Such methods also open the door to algorithmic advances in deep learning. In fact, inspired by the clustering phenomenon, we are working on a new regularization procedure for DNN training that encourages intermediate layers of the network to increase a clustering-based analog of $I(Y;T_\ell)$. This makes the regularized layer learn well-separated representations of the data (with possibly nonlinear decision boundaries) and enhances the training process, according to our initial experiments. We aim to further develop this into a practical algorithm that accelerates DNN training in the near future.

%We believe that further study of geometric phenomena driven by DNN training is warranted to better understand the learned representations and to potentially establish connections with generalization.

%\ziv{add a paragraph about implications for the future: clustering easier to measure than MI and possible future directions...}
\appendix

\section{Two-Neuron Leaky-ReLU Network Example}\label{SUPP:ReLU_Example}

\begin{figure*}[!ht]
    \setlength{\tabcolsep}{2.5pt}
    \begin{tabular}{cccc}
\includegraphics[width=0.23\textwidth]{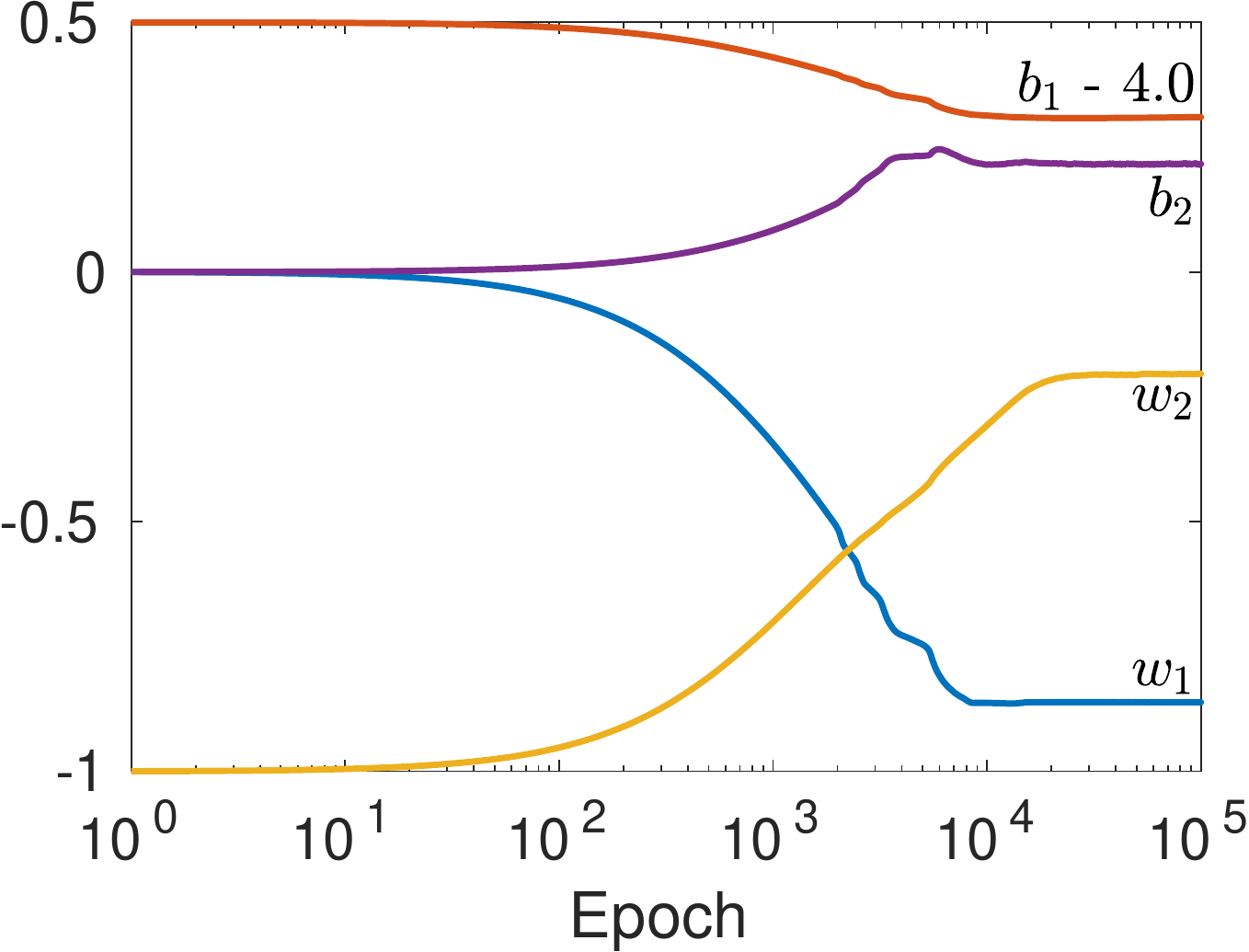}%
%\begin{picture}(0,0)(0,0)\put(-101,71){\footnotesize({\bf{a}})}\end{picture}
&
\includegraphics[width=0.24\textwidth]{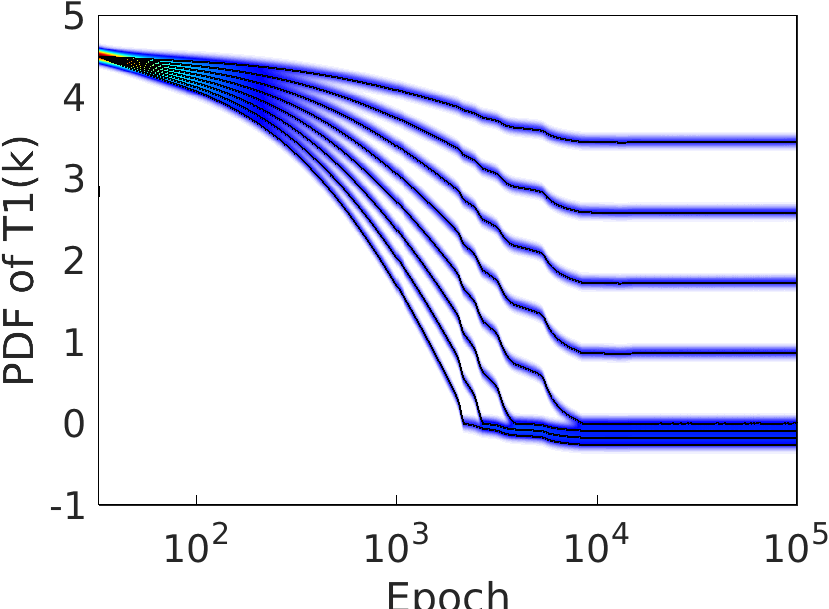}%
%\begin{picture}(0,0)(0,0)\put(-18,70){\footnotesize({\bf{b}})}\end{picture}
&
\includegraphics[width=0.24\textwidth]{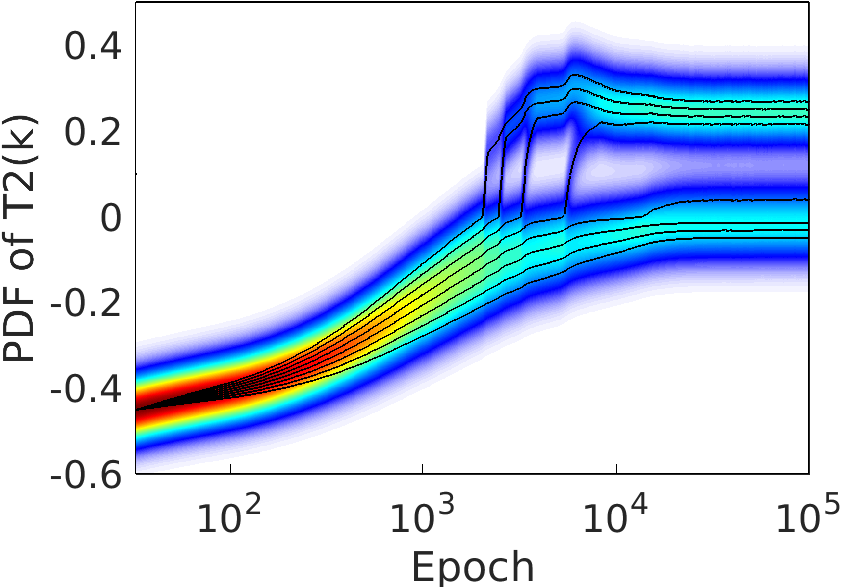}%
%\begin{picture}(0,0)(0,0)\put(-99,69){\footnotesize({\bf{c}})}\end{picture}
&
\includegraphics[width=0.238\textwidth]{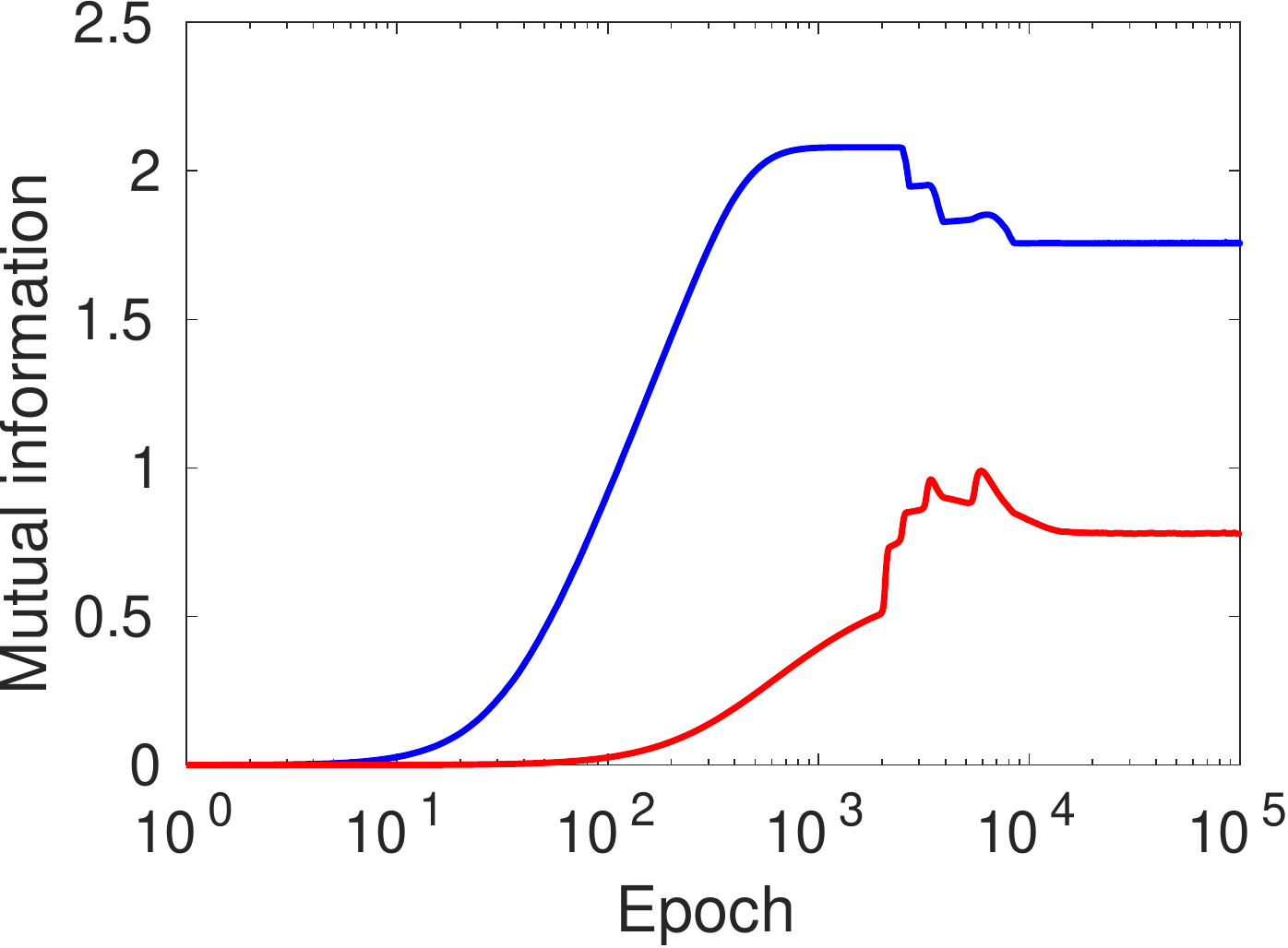}%
%\begin{picture}(0,0)(0,0)\put(-99.5,70){\footnotesize({\bf{d}})}\end{picture}
\\
%    \small({\bf{d}}) & \small({\bf{e}}) & \small({\bf{f}})
({\bf{a}}) & ({\bf{b}}) & ({\bf{c}}) & ({\bf{d}})\\[-5pt]
    \end{tabular}
\caption{Two-layer leaky ReLU network: (a) network parameters as a function of epoch, (b,c) the corresponding PDFs $p_{T_1(k)}$ and $p_{T_2(k)}$, and (d) the mutual information for both layers.}\label{Fig:TwoLayerRelu}
%  along with the lower and upper bounds obtained using the SP estimator.
\end{figure*}

To expand upon Section \ref{SEC:minimal_example}, we provide here a second example to illustrate the relation between clustering and compression of mutual information. In particular, this example also shows that as opposed to the claim from \citep{DNNs_ICLR2018}, non-saturating nonlinearities can achieve compression. Consider the non-saturating Leaky-ReLU nonlinearity $R(x)\triangleq \max(x,x/10)$. Let $\mathcal{X} = \mathcal{X}_0 \cup \mathcal{X}_{\nicefrac{1}{4}}$, with $\mathcal{X}_0 = \{1,2,3,4\}$ and $\mathcal{X}_{\nicefrac{1}{4}} = \{5,6,7,8\}$, and labels $0$ and $1/4$, respectively. We train the network via GD with learning rate 0.001 and mean squared loss. Initialization (shown in Fig. \ref{Fig:TwoLayerRelu}(a)) was chosen to best illustrate the connection between the Gaussians' motion and mutual information. 
The network converges to a solution where $w_1<0$ and $b_1$ is such that the elements in $\mathcal{X}_{\nicefrac{1}{4}}$ cluster. The output of the first layer is then negated using $w_2<0$ and the bias ensures that the elements in $\mathcal{X}_0$ are clustered without spreading out the elements in $\mathcal{X}_{\nicefrac{1}{4}}$. Figs.~\ref{Fig:TwoLayerRelu}(b) show the Gaussian motion at the output of the first layer and the resulting clustering. For the second layer (Fig.~\ref{Fig:TwoLayerRelu}(c)), the clustered bundle $\mathcal{X}_{\nicefrac{1}{4}}$ is gradually raised by growing $b_2$, such that its elements successively split as they cross the origin; further tightening of the bundle is due to shrinking $\vert w_2\vert$. Fig.~\ref{Fig:TwoLayerRelu}(d) shows the mutual information of the first (blue) and second (red) layers. The merging of the elements in $\mathcal{X}_{\nicefrac{1}{4}}$ after their initial divergence is clearly reflected in the mutual information. Likewise, the spreading of the bundle, and successive splitting and coalescing of the elements in $\mathcal{X}_{\nicefrac{1}{4}}$ are visible in the spikes in the red mutual information curve. The figure also shows how the bounds on $I\big(X;T(k)\big)$ precisely track its evolution.

%%%%%%%%%%%%%%%%%%%%%%%%%%%%%%%%%%%%%%%%%%%%%%%%%%%%%%%%%%%%%%%%%%%%%%%%%%%%%%%%%%%%%%%%%%%%%%%%%%%%%%%
%%%%%%%%%%%%%%%%%%%%%%%%%%%%%%%%%%%%%%%%%%%%%%%%%%%%%%%%%%%%%%%%%%%%%%%%%%%%%%%%%%%%%%%%%%%%%%%%%%%%%%%
%%%%%%%%%%%%%%%%%%%%%%                                                          %%%%%%%%%%%%%%%%%%%%%%%
%%%%%%%%%%%%%%%%%%%%%%               Details of the Experiments                 %%%%%%%%%%%%%%%%%%%%%%%
%%%%%%%%%%%%%%%%%%%%%%                                                          %%%%%%%%%%%%%%%%%%%%%%%
%%%%%%%%%%%%%%%%%%%%%%%%%%%%%%%%%%%%%%%%%%%%%%%%%%%%%%%%%%%%%%%%%%%%%%%%%%%%%%%%%%%%%%%%%%%%%%%%%%%%%%%
%%%%%%%%%%%%%%%%%%%%%%%%%%%%%%%%%%%%%%%%%%%%%%%%%%%%%%%%%%%%%%%%%%%%%%%%%%%%%%%%%%%%%%%%%%%%%%%%%%%%%%%
\section{Experimental Details}\label{SUPSECTION:experimentaldetails}

\subsection{SZT Model}
In this section we provide additional experimental details and results for the SZT model discussed in Section \ref{SEC:Empirical_results} of the main paper. 
%For both tanh and ReLU nonlinearities we have selected the standard deviation $\beta$ of the additive Gaussian noise so as to optimize the model's performance on the held out set. In particular, we tested standard deviations values $\beta\in\{1.0 \times 10^{-4}, 5.0\times 10^{-4}, 1.0 \times 10^{-3}, 5.0\times 10^{-3}, 1.0 \times 10^{-2}, 5.0 \times 10^{-2}, 1.0 \times 10^{-1}\}$ and found that for tanh the best noise parameter is $\beta = 0.001$ and for ReLU it is $\beta =0.01$. 

To regularize the network weights, we followed \citep{Parseval2017} and adopted their approach for enforcing an orthonormality constraint. Specifically, we first update the weights $\{\mathrm{W}_\ell\}_{\ell\in[L]}$ using the standard gradient descent step, and then perform a secondary update to set 
\begin{align*}
\mathrm{W}_\ell \leftarrow \mathrm{W}_\ell - \alpha \left(\mathrm{W}_{\ell}\mathrm{W}_{\ell}^T - \mathrm{I}_{d_\ell}\right)\mathrm{W}_{\ell},
\end{align*}
\noindent where the regularization parameter $\alpha$ controls the strength of the orthonormality constraint. The value of $\alpha$ was was selected from the set $\{1.0 \times 10^{-5}, 2.0\times 10^{-5}, 3.0\times 10^{-5}, 4.0\times 10^{-5}, 5.0\times 10^{-5}, 6.0\times 10^{-5}, 7.0\times 10^{-5}\}$ and the optimal value was found to be equal to $5.0\times 10^{-5}$ for both the tanh and ReLU.

%\begin{figure}[!ht]
%    \centering
%    \setlength{\tabcolsep}{0pt}
%    \begin{tabular}{c}
%    \includegraphics[width=\textwidth]{NIPS2018/Figures/Experiments/Tishby/tanh_beta02_unnorm.png}%
%    \begin{picture}(0,0)(0,0)\put(-394,2){({\bf{a}})}\end{picture}\\
%    \hline
%    \includegraphics[width=\textwidth]{NIPS2018/Figures/Experiments/Tishby/relu_beta01_unnorm.png}%
%    \begin{picture}(0,0)(0,0)\put(-394,2){({\bf{b}})}\end{picture}\\
%    \hline
%    \includegraphics[width=\textwidth]{NIPS2018/Figures/Experiments/Tishby/relu_beta01_norm.png}%
%    \begin{picture}(0,0)(0,0)\put(-394,2){({\bf{c}})}\end{picture}
%    \end{tabular}
%    \caption{Training results with (a) tanh nonlinearity and additive noise $\beta= 0.02$ without weight normalization, (b) ReLU nonlinearity and $\beta= 0.01$ without weight normalization, (c) ReLU nonlinearity and $\beta= 0.01$ with weight normalization. Test classification accuracy is 97\%, 96\%, and 97\%, respectively.}
%    \label{Fig:Tishby_extra}
%\end{figure}

\begin{figure*}[!ht]
    \centering
    \setlength{\tabcolsep}{0pt}
    \begin{tabular}{c}
    \includegraphics[width=\textwidth]{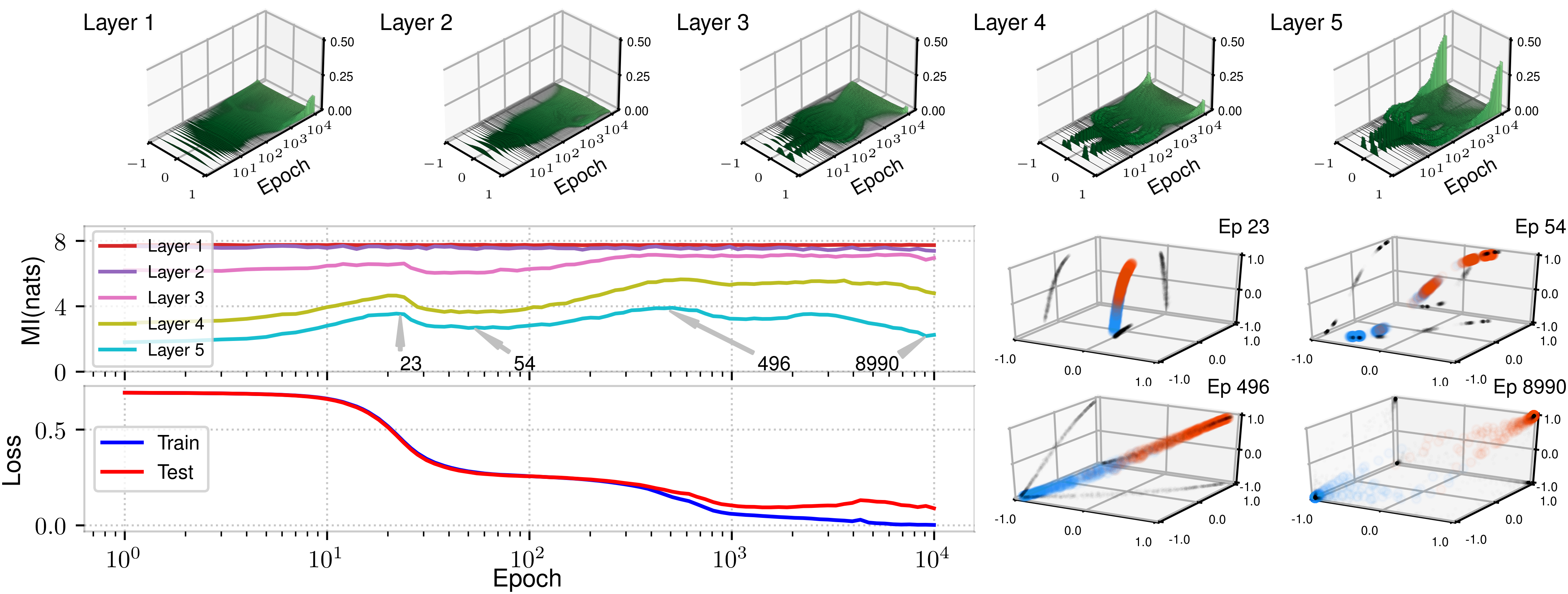}%
    \begin{picture}(0,0)(0,0)\put(-482,2){({\bf{a}})}\end{picture}\\
    \hline
    \includegraphics[width=\textwidth]{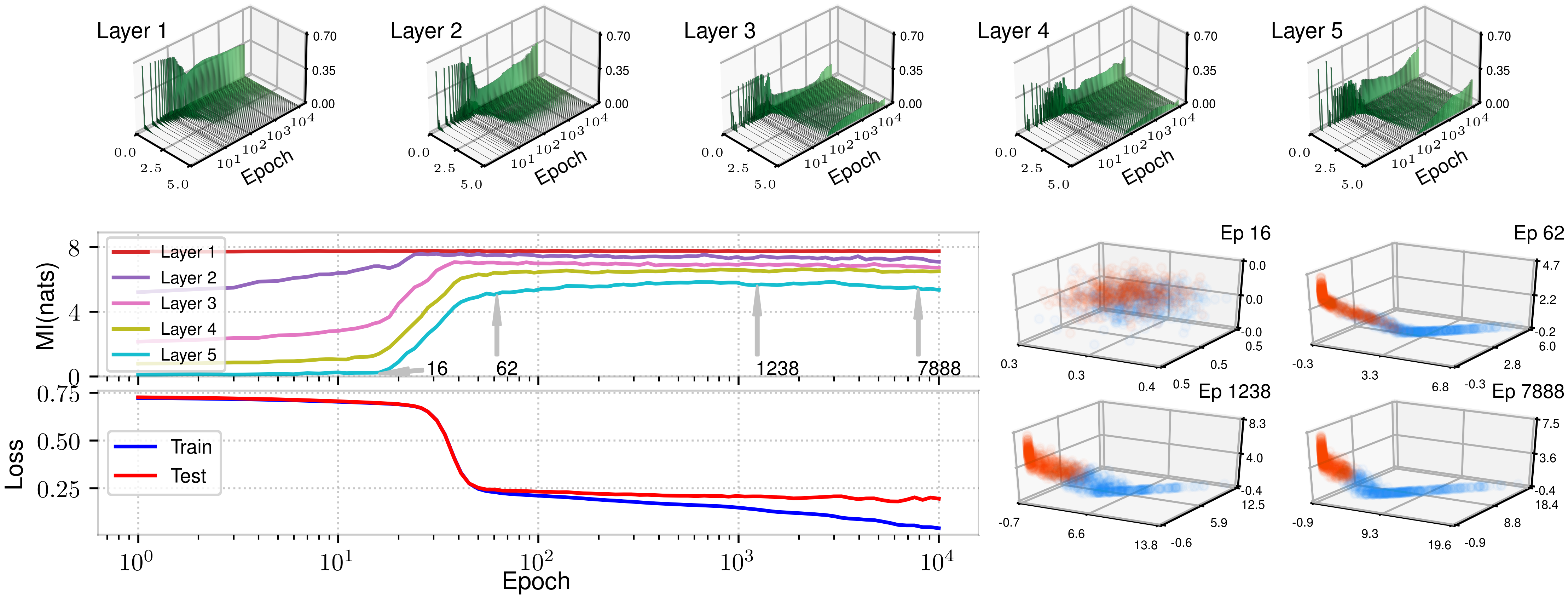}%
    \begin{picture}(0,0)(0,0)\put(-482,2){({\bf{b}})}\end{picture}\\
    \hline
    \includegraphics[width=\textwidth]{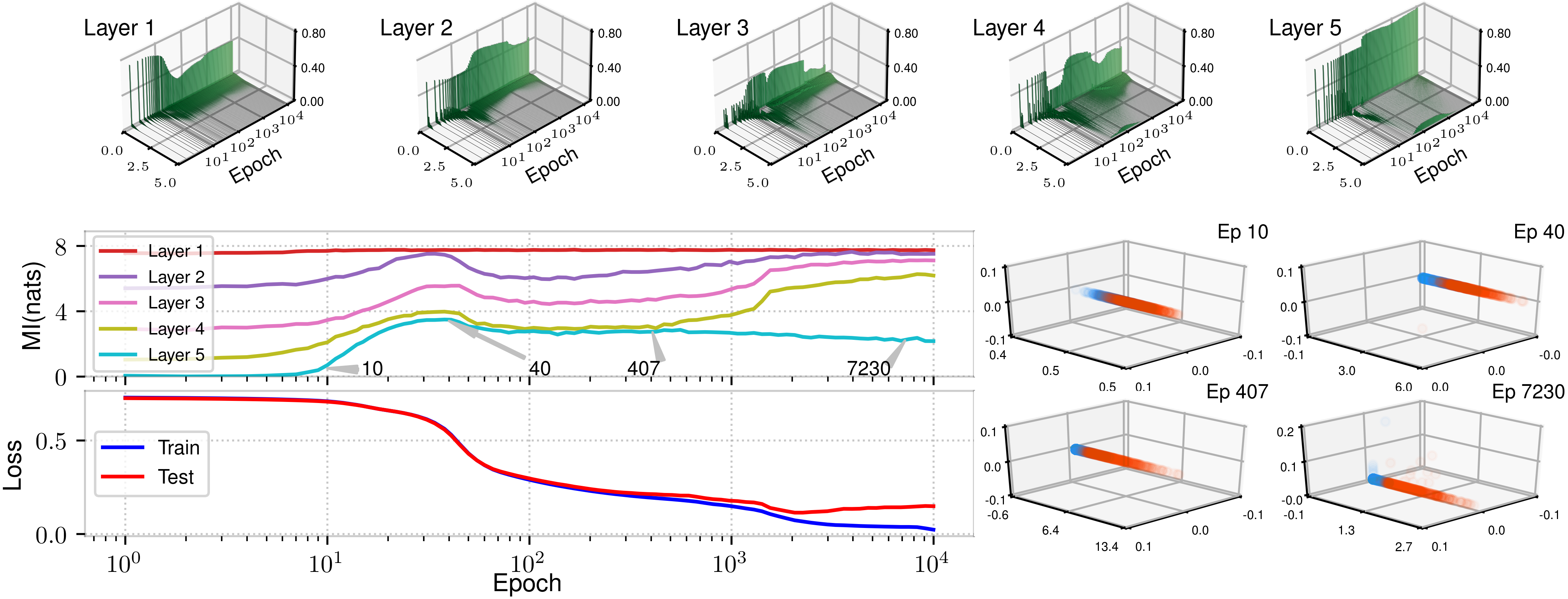}%
    \begin{picture}(0,0)(0,0)\put(-482,2){({\bf{c}})}\end{picture}
    \end{tabular}
    \caption{SZT model with (a) tanh nonlinearity and additive noise $\beta= 0.01$ without weight normalization, (b) ReLU nonlinearity and $\beta= 0.01$ without weight normalization, (c) ReLU nonlinearity and $\beta= 0.01$ with weight normalization. Test classification accuracy is 97\%, 96\%, and 97\%, respectively.}
    \label{Fig:Tishby_extra}\vspace{3mm}
\end{figure*}

In Fig.~\ref{Fig:Tishby_extra} we present additional experimental results that provide further insight into the clustering and compression phenomena for both tanh and ReLU nonlinearities. Fig.~\ref{Fig:Tishby_extra}(a) shows what happens when the additive noise has a high variance. In this case, although saturation still occurs (see the histograms on top of Fig.~\ref{Fig:Tishby_extra}(a)) and the Gaussians still cluster together (see the scatter plots on the right for the epoch 54 and epoch 8990), compression overall is very mild. The effect of increasing the noise parameter was explained in Section \ref{SEC:minimal_example} of the main text (see, in particular, Fig.~\ref{Fig:OneLayerTanh}(d) therein). Comparing Fig.~\ref{Fig:Tishby_extra}(a) to Fig.~\ref{Fig:Tishby_exp}(a) of the main text, for which $\beta=0.005$ was used and compression was observed, further highlights the effect of large $\beta$. Recall that smaller $\beta$ values correspond to narrow Gaussians, while larger $\beta$ values correspond to wider Gaussians. When $\beta$ is small, even Gaussians that belong to the same cluster are distinguishable so long as they are not too close. When clusters tighten, the in-class movement brings these Gaussians closer together, effectively merging them, and causing a reduction in mutual information (compression). One the other hand, for large $\beta$, the in-class movement is blurred at the outset (before clusters tighten). Thus, the only effect on mutual information is the separation between the clusters: as these blobs move away from each other, mutual information rises.

%Fig.~\ref{Fig:OneLayerTanh}(d) shows the mutual information for different noise levels $\beta$ as a function of weight $w$, with bias $-2w$. For small $\beta$ (as above) the $\mathcal{X}_{-1}$ Gaussians are distinct and merge in two stages as $w$ grows. For larger $\beta$, however, the $\mathcal{X}_{-1}$ Gaussians are indistinguishable for any $w$, making $I\big(X;T\big)$ only increase as the two classes gradually separate.

%As was discussed in the main text, mutual information is high when the Gaussians are far apart and it decreases when the Gaussians are closer to each other. Therefore, from the point of view of MI, the set of clustered Gaussians with high $\beta$ is equivalent to the set of spread-out Gaussians with small $\beta$. And this is precisely what is observed on the MI plot in Fig.~\ref{Fig:Tishby_tanh_unnorm02}: during the earlier epochs the Gaussians are clustered in one blob and the MI is low. Later, when they start to spread out and cluster in the corners of the tanh hypercube, the value $I(X;T_\ell)$ still grows because to the MI estimator these high-variance clustered Gaussians look more like spread out Gaussian centers.

Based on the above observation, we can conclude that while the two notions of ``clustering Gaussians'' and ``compression/decrease in mutual information'' are strongly related in the low-beta regime, once the noise becomes large, these phenomena decouple, i.e., the network may cluster inputs and neurons may saturate, but this will not be reflected in a decrease of mutual information. 

% \begin{comment}
% \begin{figure}[!ht]
%     \centering
%     \includegraphics[width=\textwidth]{NIPS2018/Figures/Experiments/Tishby/relu_beta01_unnorm.png}
%     \caption{ReLU nonlinearity and additive noise $\beta= 0.01$ without weight normalization. Test classification accuracy 96\%.}
%     \label{Fig:Tishby_relu_unnorm01}
% \end{figure}

% \begin{figure}[!ht]
%     \centering
%     \includegraphics[width=\textwidth]{NIPS2018/Figures/Experiments/Tishby/relu_beta01_norm.png} 
%     \caption{ReLU nonlinearity and additive noise $\beta= 0.01$ with weight normalization. Test classification accuracy 97\%.}
%     \label{Fig:Tishby_relu_norm01}
% \end{figure}
% \end{comment}

Finally, we present results for ReLU activation without weight normalization (Fig.~\ref{Fig:Tishby_extra}(b)) and with orthonormal weight regularization (Fig.~\ref{Fig:Tishby_extra}(c)). We see that both these networks exhibit almost no compression. For Fig.~\ref{Fig:Tishby_extra}(c), the lack of compression is attributed to regularization of the weight matrices, as explained in Section \ref{SEC:Empirical_results} of the main text. For Fig.~\ref{Fig:Tishby_extra}(b), the reduction in compression can be explained by the fact that although ReLU forces saturation of the neurons at the origin (which promotes clustering), since the positive axes remain unconstrained, the Gaussians can move off towards infinity without bound. This is visible from the histograms in the top row of Fig.~\ref{Fig:Tishby_extra}(b), where, for example, in layer 5 the neurons can take arbitrarily large positive values (note that the bin corresponding to the value $5$ accumulates all the values from $5$ to infinity). Therefore, the clustering at the origin and the potential drop in mutual information is counterbalanced by the spread of Gaussians along the positive axes and the potential increase of mutual information it causes. Eventually, this leads to the approximately constant profile of the mutual information plot in Fig.~\ref{Fig:Tishby_extra}(b).

The behavior of the weight-normalized ReLU in Fig.~\ref{Fig:Tishby_extra}(c) is similar to Fig.~\ref{Fig:Tishby_extra}(b), although now the growth of the network weights is bounded and the saturation around origin is reduced. For example, for layers 4 and 5 we can see an upward trend in the mutual information, which is then flattened at the end of training. This occurs since more Gaussians are moving away from the origin, although their motion remains bounded (see the histograms on the top and the scatter plots on the right), thus decreasing the clustering density, leading to the rise in the mutual information profile. Once the Gaussians are prevented from moving any further along the positive axes, a slight compression occurs and the mutual information flattens.

\subsection{Spiral Model}
\begin{figure}[!ht]
    \centering
    \setlength{\tabcolsep}{0pt}
    \begin{tabular}{c}
    \includegraphics[width=0.5\textwidth]{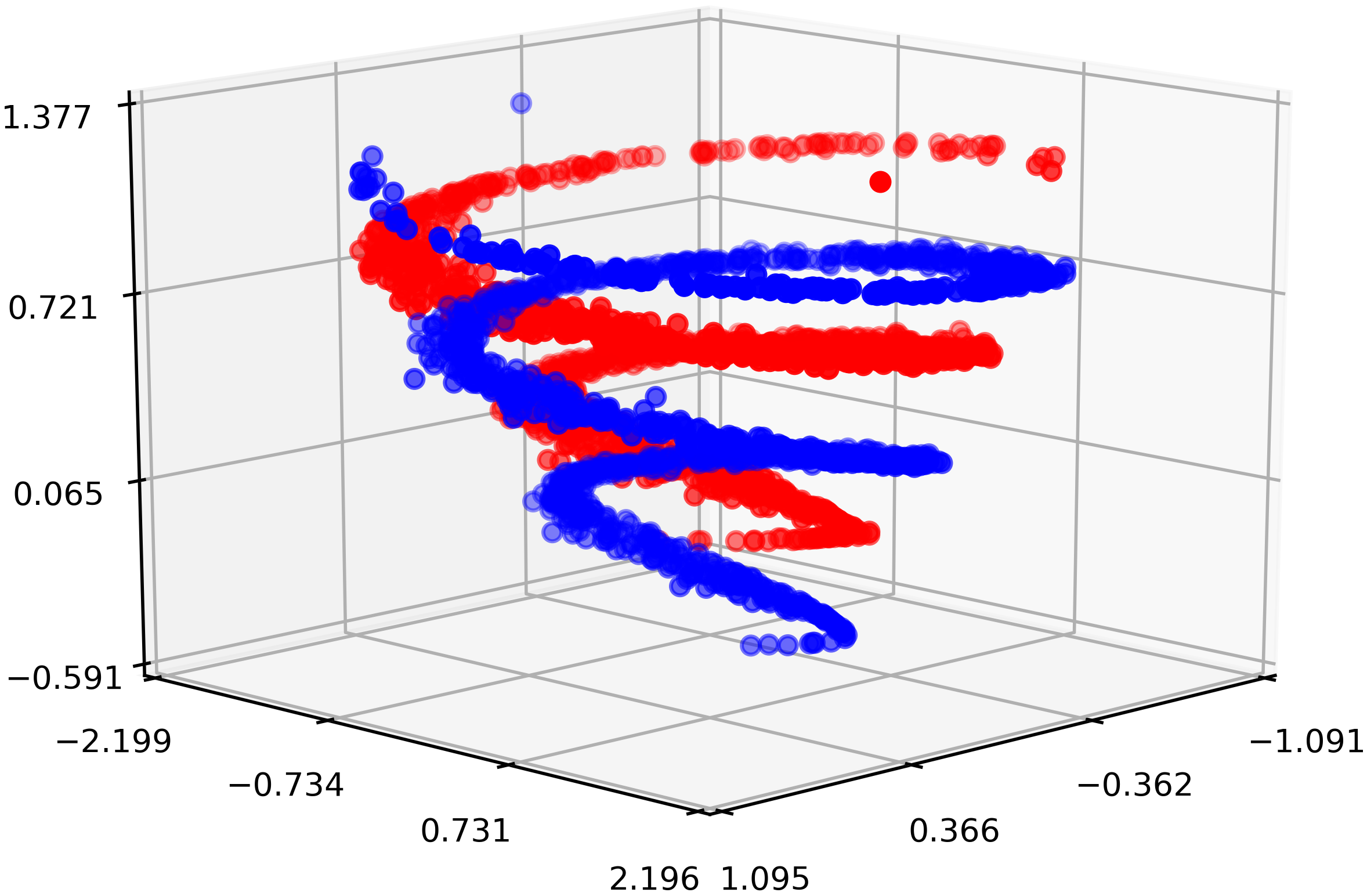}%
    \end{tabular}
    \caption{Generated spiral data for binary classification problem.}
    \label{Fig:Spiral_data}
\end{figure}

In this section we present results for another synthetic example. We generated data in the form of spiral as in Fig. \ref{Fig:Spiral_data}. The network architecture was similar to SZT model, except that the size of each layer was set to 3.

\begin{figure*}[!ht]
    \centering
    \setlength{\tabcolsep}{0pt}
    \begin{tabular}{c}
    \includegraphics[width=\textwidth]{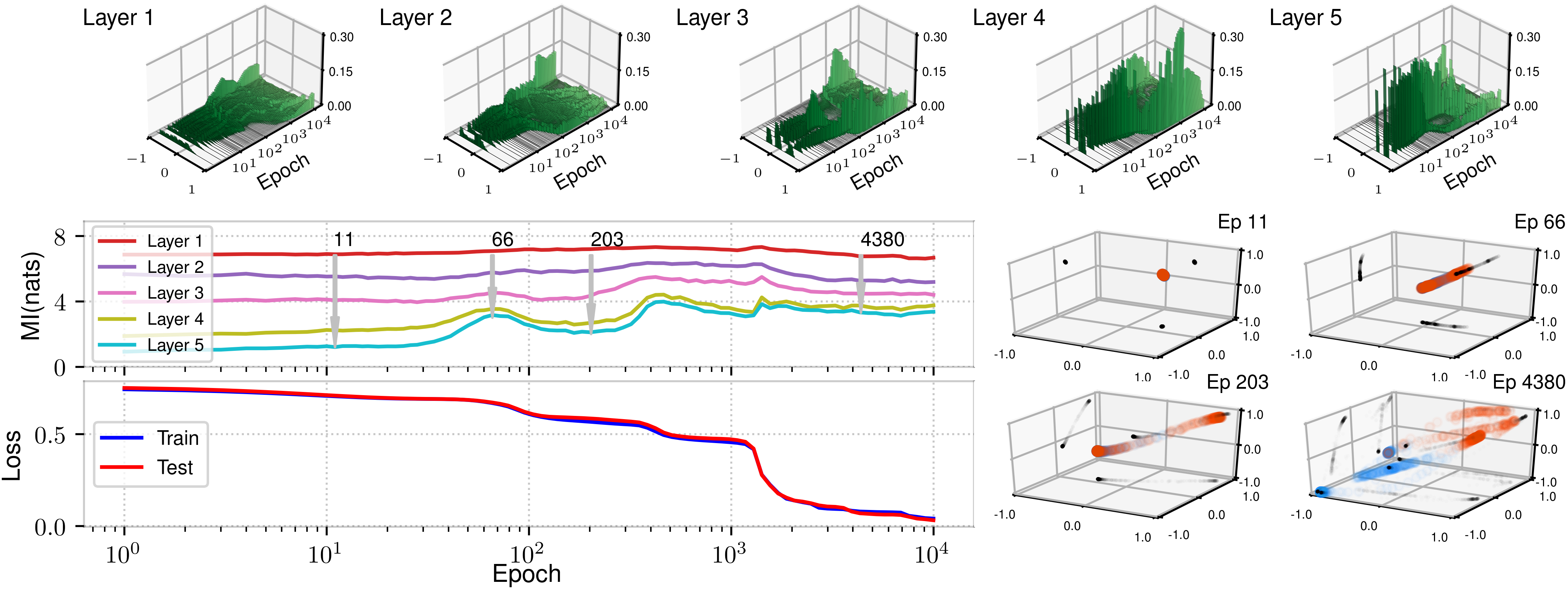}\\
    \includegraphics[width=0.6\textwidth]{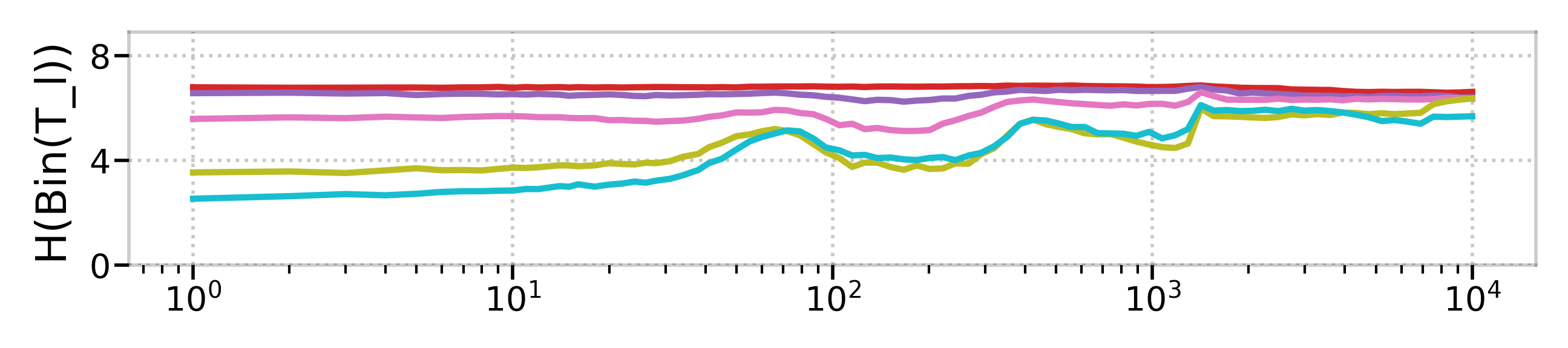}%
    \begin{picture}(0,0)(0,0)\put(-387,2){({\bf{a}})}\end{picture}\\
    \hline
    \includegraphics[width=\textwidth]{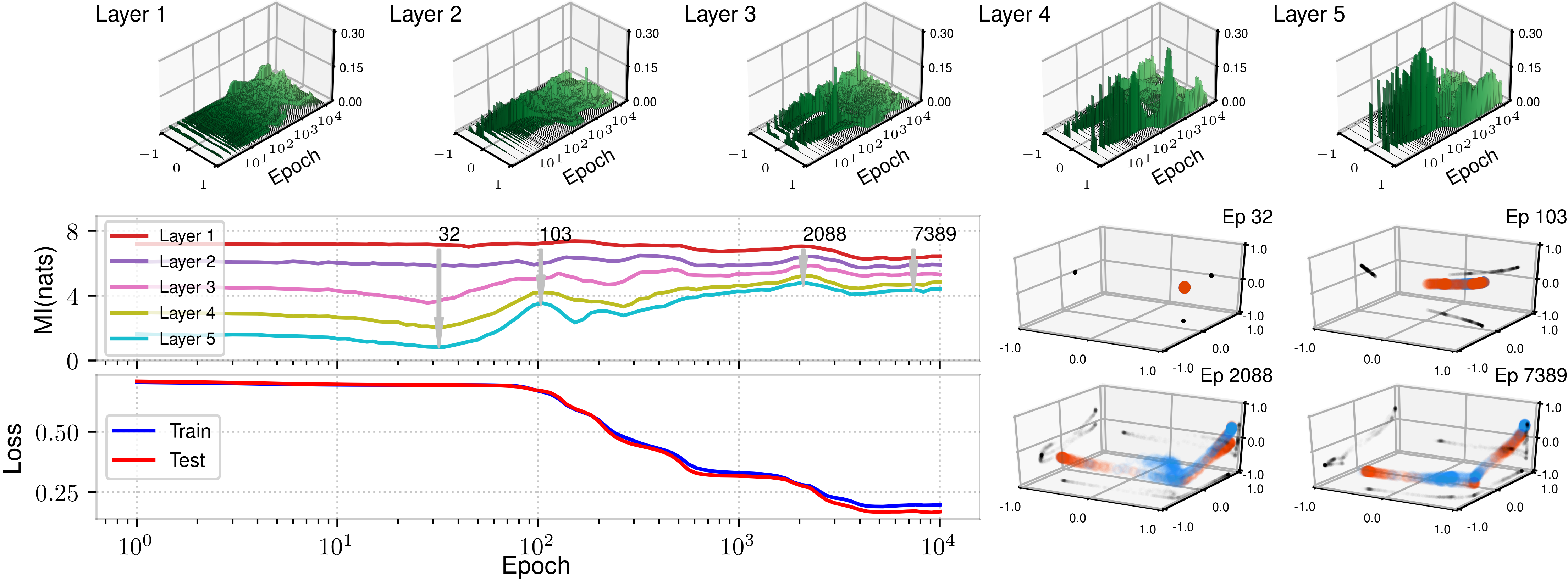}\\
    \includegraphics[width=0.6\textwidth]{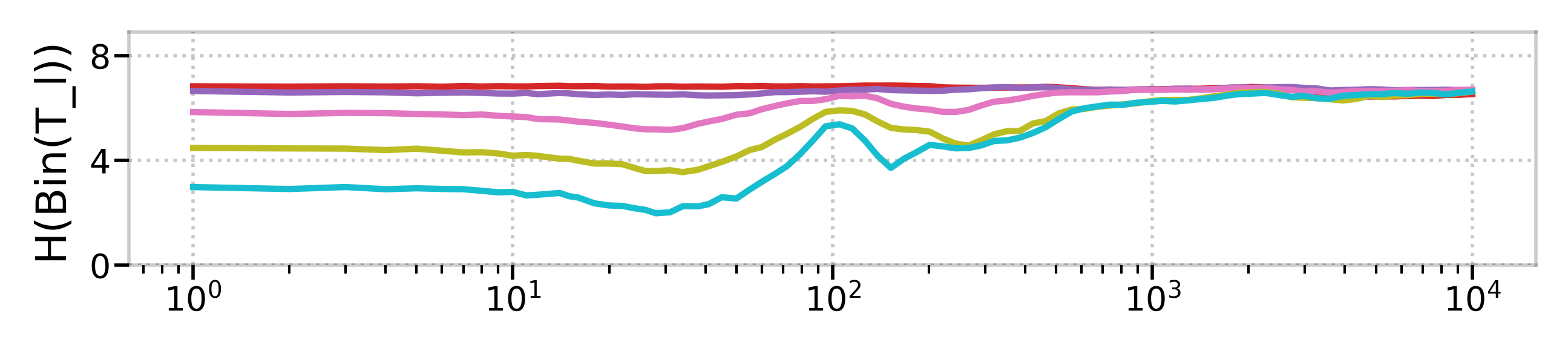}%
    \begin{picture}(0,0)(0,0)\put(-387,2){({\bf{b}})}\end{picture}
    \end{tabular}
\caption{(a) Evolution of $I(X;T_\ell)$ and training/test losses across training epochs for Spiral dataset with $\beta= 0.005$ and tanh nonlinearities. The scatter plots on the right are the values of Layer 5 ($d_5 = 3$) at the arrow-marked epochs on the mutual information plot. The bottom plot shows the entropy estimate $H\big(\mathsf{Bin}(T_\ell)\big)$ across epochs for bin size $B = 10\beta$. (b) Same setup as in (a) but with a regularization that encourages orthonormal weight matrices.}
    \label{Fig:spiral_noisy}
\end{figure*}

Fig. \ref{Fig:spiral_noisy} shows MI estimates $I(X;T_\ell)$ computed using SP estimator and the discrete entropy estimates $H\big(\mathsf{Bin}(T_\ell)\big)$ for weight un-normalized Fig. \ref{Fig:spiral_noisy} (a) and normalized models Fig. \ref{Fig:spiral_noisy} (b) and using additive noise $\beta = 0.005$. Similar as in the main paper, the results in the figure illustrate a connection between clustering and compression.  
 
\begin{figure*}[!ht]
    \centering
    \setlength{\tabcolsep}{0pt}
    \begin{tabular}{c}
    \includegraphics[width=0.6\textwidth]{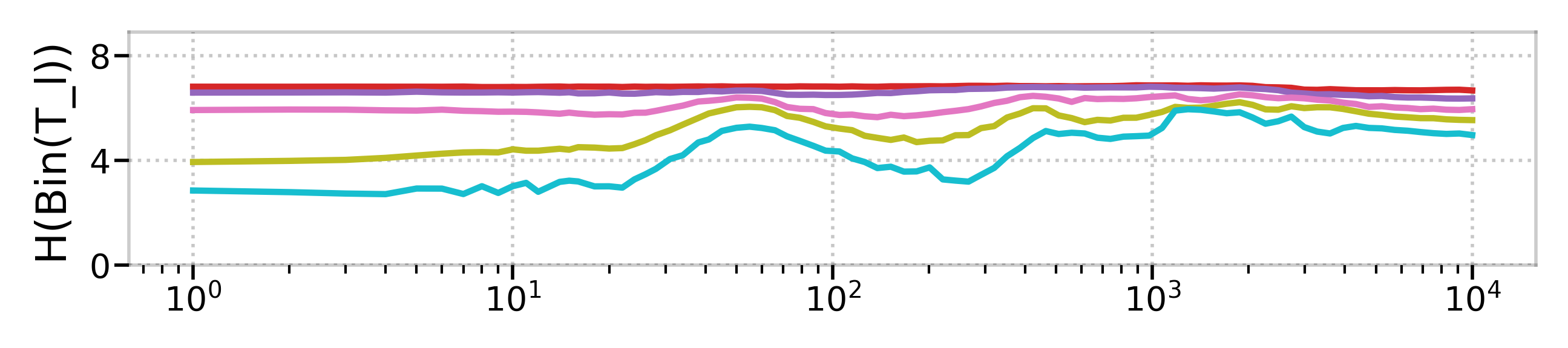}%
    \end{tabular}
\caption{$H\big(\mathsf{Bin}(T_\ell)\big)$ estimate for deterministic net using spiral data. Bin size was set to $B=0.001$.}
    \label{Fig:spiral_determ}
\end{figure*}

Finally, in Fig. \ref{Fig:spiral_determ} we also show an estimate of $H\big(\mathsf{Bin}(T_\ell)\big)$ for the case of deterministic DNN trained on spiral data. For the particular choice of the bin size, the result of the estimated entropy reveal a certain level of clustering granularity.

\subsection{MNIST CNN}
\label{supp:MNIST}
In this section, we describe in detail the architecture of the MNIST CNN models used in Sections~\ref{SEC:Noisy_DNNs}~and~\ref{SEC:Empirical_results} in the main paper.

The MNIST CNNs were trained using PyTorch~\citep{PyTorch2017} version {\tt 0.3.0.post4}. The CNNs use the following fairly standard architecture with two convolutional layers, two fully connected layers, and batch normalization.
\begin{enumerate}
\item 2-d convolutional layer with 1 input channel, 16 output channels, 5x5 kernels, and input padding of 2 pixels
\item Batch normalization
\item Tanh() activation function
\item Zero-mean additive Gaussian noise with variance $\beta^{2}$ or dropout with a dropout probability of 0.2
\item 2x2 max-pooling
\item 2-d convolutional layer with 16 input channels, 32 output channels, 5x5 kernels, and input padding of 2 pixels
\item Batch normalization
\item Tanh() activation function
\item Zero-mean additive Gaussian noise with variance $\beta^{2}$ or dropout with a dropout probability of 0.2
\item 2x2 max-pooling
\item Fully connected layer with 1586 (32x7x7) inputs and 128 outputs
\item Batch normalization
\item Tanh() activation function
\item Zero-mean additive Gaussian noise with variance $\beta^{2}$ or dropout with a dropout probability of 0.2
\item Fully connected layer with 128 inputs and 10 outputs
\end{enumerate}
All convolutional and fully connected layers have weights and biases, and the weights are initialized using the default initialization, which draws weights from $\mathsf{Unif}[-1/\sqrt{m},1/\sqrt{m}]$, with $m$ the fan-in to a neuron in the layer. Training uses cross-entropy loss, and is performed using stochastic gradient descent with no momentum, 128 training epochs, and 32-sample minibatches. The initial learning rate is $5 \times 10^{-3}$, and it is reduced following a geometric schedule such that the learning rate in the final epoch is $5 \times 10^{-4}$. To improve the test set performance of our models, we applied data augmentation to the training set by translating, rotating, and shear-transforming each training example each time it was selected. Translations in the $x$- and $y$-directions were drawn uniformly from $\{-2,-1,0,1,2\}$, rotations were drawn from $\mathsf{Unif}(-10^\circ,10^\circ)$, and shear transforms were drawn from $\mathsf{Unif}(-10^\circ,10^\circ)$.

To obtain more reliable performance results, we train eight different models and report the mean number of errors and standard deviation of the number of errors on the MNIST validation set. To ensure that the internal representations of different models are comparable, which is necessary for the use of the cosine similarity measure between internal representations, for each noise condition (deterministic, noisy with $\beta = 0.05$, noisy with $\beta = 0.1$, noisy with $\beta = 0.2$, noisy with $\beta = 0.5$, and dropout with $p = 0.2$), we use a common random seed (different for the eight replications, of course) so the models have the same initial weights and access the training data in the same order (use the same minibatches).

At test time, all models are fully deterministic: the additive noise blocks and dropout layers are replaced by identities. Thus, in the figures and text in the main paper, ``Layer 1'' is the output of step 5 (2x2 max-pooling), ``Layer 2'' is the output of step 10 (2x2 max-pooling), ``Layer 3'' is the output of step 13 (Tanh() activation function), and ``Layer 4'' is the output of step 15 (fully connected layer with 10 outputs).

\section{Sample Propagation Estimator - Theoretic Guarantees}\label{SUPPSEC:Sample_prop_theory}

%Both conditional and unconditional entropy estimators reduce to the problem of estimating $h(p_S\ast\varphi_\beta)$ using i.i.d. samples $S^n\triangleq(S_i)_{i\in[n]}$ from $S\sim p_S$ while knowing $\varphi_\beta$. 

In this section we state performance guarantees for the SP estimator. We cite several foundational theorems from our work \citep{anonymized_ISIT_estimation2019}, where this estimation problem is thoroughly studied. An anonymized copy of that paper is found at the end of the supplement and cited when needed. Proofs of all other results are relegated to Supplement \ref{SUPPSEC:proofs}.

%The interested reader is referred to \citep{SP_estimation2018} for proofs of the subsequently stated results. 
%For simplicity of presentation, throughout this subsection (and the corresponding proofs) we omit the layer index $\ell$ from our notation. Thus, we consider the estimation of $h(T)=h(S+Z)$, where $S\sim f_S$ and $Z\sim\mathcal{N}(0,\beta^2I_{d})$ are independent, provided $n$ i.i.d. samples $S^n\triangleq(S_j)_{j\in[n]}$ from $f_S$ and knowing $\beta^2$.

\subsection{Preliminary Definitions}

Consider the estimation of the differential entropy $h(S+Z)=h(P\ast\varphi_\beta)$ based on $n$ i.i.d. samples of $S\sim P$, where $P$ is unknown and belongs to some nonparametric class, and $\varphi_\beta$ (a PDF of an isotropic Gaussian with parameter $\beta$) is known. The minimax absolute-error risk over a given nonparametric class of distributions $\mathcal{F}$ is
\begin{equation}
    \mathcal{R}^\star(n,\beta,\mathcal{F})\triangleq\inf_{\hat{h}}\sup_{P\in\mathcal{F}}\mathbb{E}\left|h(P\ast\varphi_\beta)-\hat{h}(S^n,\beta)\right|,\label{EQ:nonparametric_class1}
\end{equation}
where $\hat{h}$ is the estimator and $S^n\triangleq (S_i)_{i\in[n]}$ are the samples from $P$. In \eqref{EQ:nonparametric_class1}, by $P\ast\varphi_\beta$ we mean either: (i) $(P\ast\varphi_\beta)(x)=\int p(u)\varphi_\beta(x-u)du=(p\ast\varphi_\beta)(x)$, when $P$ is continuous with density $p$; or (ii) $(P\ast\varphi_\beta)(x)=\sum_{u:\ p(u)>0} p(u)\varphi_\beta(x-u)$, if $P$ is discrete with PMF $p$. This convolved distribution can be defined generally in a way that the two instances above as special cases using measure-theoretic concepts (see \citep{anonymized_ISIT_estimation2019}). Regardless of the nature of $P$, however, we stress that $P\ast\varphi_\beta$ is always a continuous distribution since it corresponds to the random variable $S+Z$, where $Z$ is an isotropic Gaussian vector. The sample complexity $n^\star(\eta,\beta,\cal F)$ is defined as the smallest number of samples $n$ required to achieve a risk value less than or equal to a specified constant $\eta$ in \eqref{EQ:nonparametric_class1}.

Let $\mathcal{F}_d$ be the set of distributions $P$ with $\supp(P)\subseteq[-1,1]^d$.\footnote{Any support included in a compact subset of $\mathbb{R}^d$ would do. We focus on the case of $\supp(P)\subseteq[-1,1]^d$ due to its correspondence to a noisy DNN with tanh nonlinearities.} Furthermore, let $\mathcal{F}^{(\mathsf{SG})}_{d,\mu,K}$ be the class of $K$-subgaussian distributions, where we adopt the subgaussianity definition from \citep{hsu2012tail}. Namely, $P\in\mathcal{F}^{(\mathsf{SG})}_{d,\mu,K}$, for $\mu\geq 0$ and $K>0$, if $X\sim P$ satisfies $\|\mathbb{E} X\|\leq \mu$ and 
\begin{equation}
\mathbb{E}\Big[\exp\big(\alpha^T(X - \mathbb{E}X)\big)\Big] \leq \exp\big(0.5 K^2 \|\alpha\|^2\big), \forall\alpha \in \mathbb{R}^d,
\end{equation}
i.e., every one-dimensional projection of $X$ is subgaussian. Clearly, there exists a $K'>0$ such that $\mathcal{F}_d\subseteq\mathcal{F}^{(\mathsf{SG})}_{d,0,K'}$. We therefore state our lower bound results (Theorem \ref{TM:sample_complex_worstcase_asymp}) for  $\mathcal{F}_d$, while the upper bound (Theorem \ref{TM:sample_complex_bound}) is given for $\mathcal{F}^{(\mathsf{SG})}_{d,\mu,K}$. The class $\mathcal{F}_d$ corresponds to hidden layers with bounded nonlinearities (such as tanh or sigmoid), while $\mathcal{F}^{(\mathsf{SG})}_{d,\mu,K}$ accounts for ReLU nonlinearities (when, for example, the input $X$ is itself subgaussian).

\subsection{Sample Complexity is Exponential in Dimension}

We start with Theorem 1 from \cite{anonymized_ISIT_estimation2019}, which states that the sample complexity of any good estimator of $h(P\ast\varphi_\beta)$ (to within an additive gap $\eta$) is exponential in $d$. %The first claim states that there exists a class of distributions $P$, for which the estimation of $h(P\ast\varphi_\beta)$ cannot be done with fewer than exponentially many samples in $d$, when $d$ is sufficiently large. 

\begin{theorem}[Theorem 1 from \cite{anonymized_ISIT_estimation2019}]\label{TM:sample_complex_worstcase_asymp}  
The following~holds:
\begin{enumerate}%[leftmargin=*]
    \item Fix $\beta > 0$. There exist $d_0(\beta)\in\mathbb{N}$, $\eta_0(\beta)>0$ and $\gamma(\beta)>0$ (monotonically decreasing in $\beta$), such that for all $d\geq d_0(\beta)$ and $\eta<\eta_0(\beta)$ we have sample complexity $n^\star(\eta,\beta,\mathcal{F}_d) \ge \Omega\left(\frac{2^{\gamma(\beta) d}}{d\eta}\right)$.
    
    \item Fix $d\in\mathbb{N}$. There exist $\beta_0(d),\eta_0(d)>0$, such that for~all $\beta<\beta_0(d)$ and $\eta<\eta_0(d)$ we have sample complexity $n^\star(\eta,\beta,\mathcal{F}_d) \mspace{-2mu}\ge\mspace{-2mu}\Omega\left(\frac{2^d}{\eta d}\right)$.
\end{enumerate}
\end{theorem}
The exponent $\gamma(\beta)$ being monotonically decreasing in $\beta$ suggests that larger values of $\beta$ are favorable for estimation. Part 1 of the theorem states that an exponential sample complexity is inevitable when $d$ is large. As a complementary result, the second part gives a sample complexity lower bound valid in any dimension for a small noise parameter. Nonetheless, the result accounts for orders of $\beta$ considered in this work.%that were found in \citep{NIPS_Info_flow2018} as optimizing performance for, e.g., MNIST classification  (see Remark \ref{REM:beta_vals}).

%xperiments the noisy DNN estimation we use throughout the experimental part of this work
%The main idea of the proof is to relate the estimation of $h(S+Z)$ to the transmission rate over an additive white Gaussian noise (AWGN) channel. Then, we show that estimating the entropy of interest is equivalent to estimating the entropy of a discrete random variable with some distribution over a certain good constellation for transmission, whose size is exponential in $d$. Therefore, (discrete) entropy estimation over the constellation cannot be done in less than roughly $\frac{2^{\gamma d}}{d}$ samples, for some $\gamma>0$. For a fixed $d$, the exponent $\gamma$ is monotonically decreasing in $\beta$, implying that larger values of $\beta$ are favorable for estimation. 
%A technical assumption the result relies on is that $2d\cdot Q\left(\frac{1}{2\beta}\right)<1$, where $Q$ is the Q-function\footnote{The Q-function is defined as $Q(x)\triangleq\frac{1}{\sqrt{2\pi}}\int_x^\infty e^{-\frac{t^2}{2}}dt$.}. this ensures that the argument of a certain logarithm is greater than 1. For any relevant values of $d$ and $\beta$ this is satisfied since $2d\cdot Q\left(\frac{1}{2\beta}\right)$ is extremely small. For instance, the largest value of the noise standard deviation we consider is $\beta=0.1$. Even here $Q(5)\approx 2.86\cdot 10^{-7}$, which produces a very small positive number for any reasonable dimension. 

\begin{remark}[Critical $\bm{\beta}$ Values]\label{REM:beta_vals}
Theorem \ref{TM:sample_complex_worstcase_asymp} is stated in asymptotic form for simplicity. We note that, for any $d$, the critical $\beta_0(d)$ value from the second part can be extracted by following the constants through the proof (which relies on Proposition 3 from \cite{discrete_entropy_est_Wu2016}). These critical values are not unreasonably small. For example for $d = 1$, a careful analysis gives that Theorem \ref{TM:sample_complex_worstcase_asymp} holds for all $\beta < 0.08$, which is satisfied by most of the experiments in this paper. This threshold on $\beta$ changes very slowly when increasing $d$ due to the rapid decay of the PDF of the normal distribution. %Accordingly, the result of Theorem \ref{TM:sample_complex_worstcase_smalld} is valid for most of the experiments therein.
\end{remark}

\subsection{Estimation Risk Bounds}\label{SUPPSUBSEC:risk_bounds}

We next focus on analyzing the performance of the SP mutual information estimator. We start by citing Theorem 2 of \cite{anonymized_ISIT_estimation2019}, where the risk of the entropy estimation problem is bounded. Recall that the estimator of $h(P\ast\varphi_\beta)$ is $h(\hat P_{S^n}\ast\varphi_\beta)$, where $S^n=(S_i)_{i=1}^n$ is an i.i.d. sample set from $P$ and $\hat P_{S^n}$ is their empirical distribution. The following theorem shows that the expected absolute error of this estimator decays at a rate of estimation  $O\left(\frac{c^d}{\sqrt{n}}\right)$, for a numerical constant $c$ and all dimensions $d$. A better rate of convergence with $n$ cannot be attained due to the parametric estimation lower bound (see, e.g., Proposition 1 of \cite{chen1997general}). The exponential dependence in $d$ is also necessary as established by Theorem \ref{TM:sample_complex_worstcase_asymp}.

\begin{theorem}[Theorem 2 from \cite{anonymized_ISIT_estimation2019}]\label{TM:sample_complex_bound}
Fix $\beta>0,d\geq1$. Then 
\begin{align*}
    &\sup_{P \in \mathcal{F}_{d,\mu,K}^{(\mathsf{SG})}}\mathbb{E}\left|h(P\ast\varphi_\beta)-h(\hat{P}_{S_n}\ast\varphi_\beta)\right|\\
    &\leq \left(\mspace{-2mu}\frac{1}{\sqrt{2}}\mspace{-2mu}+\mspace{-2mu}\frac{K}{\beta}\right)^{\mspace{-7mu}\frac{d}{2}}\mspace{-2mu}\\
    &\qquad\times\left(\frac{8\big(2 \mu^4 \mspace{-2mu}+\mspace{-2mu} 32 d^2K^4 \mspace{-2mu}+\mspace{-2mu} d(d+2)(K\mspace{-2mu} +\mspace{-2mu} \beta/ \sqrt 2)^4\big)}{\beta^4}\right)^{\frac{1}{2}}\\
&\quad\quad\quad\quad\quad\times\exp\left( \frac{3d}{16}+\frac{\mu^2}{4(K+\beta/\sqrt{2})^2}\right)\frac{1}{\sqrt{n}}.\numberthis\label{EQ:Plugin_risk_bound}
\end{align*}
\end{theorem}

\begin{remark}[Improved Constant for Bounded Support]
Theorem \ref{TM:sample_complex_bound} also applies to the narrower nonparametric class $\mathcal{F}_d$ in place of $\mathcal{F}_{d,\mu,K}^{(\mathsf{SG})}$. By directly analyzing this bounded support scenario\footnote{e.g., by employing Proposition 5 from \cite{PolyWu} to control the entropy difference via a Wasserstein 1 distance and them using Theorem 6.15 from \cite{villani2006optimal} to bound the latter by an expression that lands itself for an elementary analysis.} ($P \in \mathcal{F}_d$) one may improve the constant factor in Theorem \ref{TM:sample_complex_bound} to give a bound of $\max\{1,\beta^{-d}\}2^{d+2}\sqrt{\frac{d}{n}}$.
\end{remark}

\begin{remark}[Comparison to Generic Estimators]
   Note that one could always sample $\varphi_\beta$ and add the obtained noise samples to $S^n$ to obtain a sample set from $P\ast\varphi_\beta$. These samples can be used to get a proxy of $h(P\ast\varphi_\beta)$ via a kNN- or a KDE-based differential entropy estimator. However, $P\ast\varphi_\beta$ violates the boundedness away from zero assumption that most of the convergence rate results in the literature rely on \cite{levit1978asymptotically,hall1984limit,joe1989estimation,hall1993estimation,tsybakov1996root,el2009entropy,sricharan2012estimation,singh2016finite,kandasamy2015nonparametric}. Two recent works that weakened/dropped the boundedness from below assumption, providing general-purpose estimators whose risk bounds are valid in our setup, are \cite{han2017optimal} and \cite{berrett2019efficient}. However, the analysis of the KDE-based estimator proposed in \cite{han2017optimal} holds only for Lipschitz smoothness parameters up to $s\leq2$ and attains the slow rate (overlooking multiplicative polylogarithmic factors) of $O\left(n^{-\frac{s}{s+d}}\right)$. The second work \cite{berrett2019efficient} studies a weighted-kNN estimator in the high smoothness regime and proved its asymptotic efficiency. However, no explicit risk bounds were derived in that work and empirically the estimator is significantly outperformed by $h(\hat P_{S^n}\ast\varphi_\beta)$ (see Section V of \cite{anonymized_ISIT_estimation2019}).
\end{remark}

We now show how the theoretical guarantee on the accuracy of the differential entropy estimator (Theorem \ref{TM:sample_complex_bound}) translates to mutual information estimation via the SP estimator from \eqref{EQ:MI_estimator_final}. To formulate the claim, recall that $T_\ell=S_\ell+Z_\ell$, where $S_\ell\sim P_{S_\ell}=P_{f_\ell(T_{\ell-1})}$ and $Z_\ell\sim\mathcal{N}(0,\beta^2\mathrm{I}_{d_\ell})$ are independent. Thus, 
\begin{subequations}
\begin{align}
h(T_\ell)&=h(P_{S_\ell}\ast\varphi_\beta)\label{EQ:uncond_ent}\\
h(T_\ell|X=x)&=h(P_{S_\ell|X=x_i}\ast\varphi_\beta).\label{EQ:cond_ent}
\end{align}
\end{subequations}
Provided $n$ i.i.d. samples $\mathcal{X}=\{X_i\}_{i\in[n]}$ from $P_X$, the DNN's generative model enables sampling from $P_{S_\ell}$ and $P_{S_\ell|X}$ as follows: %and the relation to estimating $h(P\ast\varphi_\beta)$ (for an appropriate distribution $P$) is described next. 
\begin{enumerate}
    \item \textbf{Unconditional Sampling:} To generate the sample set from $P_{S_\ell}$, feed each $X_i$, for $i\in[n]$, into the DNN and collect the outputs it produces at the $(\ell-1)$-th layer. The function $f_\ell$ is then applied to each collected output to obtain $S_\ell^n\triangleq\{S_{\ell,1},S_{\ell,2},\ldots,S_{\ell,n}\}$, which is a set of $n$ i.i.d. samples from $P_{S_\ell}$. %The unconditional entropy therefore is $h(p_{T_\ell})=h(P_{S_\ell}\ast\varphi_\beta)$ - we will estimate this from $S_\ell^n$ and the knowledge of $\varphi_\beta$.% A similar argument applies for the conditional entropies since $(X,S_\ell)$ is independent of $Z_\ell$.

    \item \textbf{Conditional Sampling Given $\bm{X}$:} %Since features do not repeat themselves we assume that each $x_i$ is unique. 
    To generate i.i.d. samples from $P_{S_\ell|X=x_i}$, for $i\in[n]$, we feed $X_i$ into the DNN $n$ times, collect outputs from $T_{\ell-1}$ corresponding to different noise realizations, and apply $f_\ell$ on each. Denote the obtained samples by $S_\ell^n(X_i)$.% These samples are used to estimate $h(p_{T_\ell|X=x_i})=h(P_{S_\ell|X=x_i}\ast\varphi_\beta)$.
    \footnote{The described sampling procedure is valid for any layer $\ell\geq 2$. For $\ell=1$, $S_1$ coincides with $f_1(X)$ but the conditional samples are undefined. Nonetheless, noting that for the first layer $h(T_1|X)=h(Z)=\frac{d}{2}\log(2\pi e \beta^2)$, we see that no estimation of the conditional entropy is needed. The mutual information estimator given in  \eqref{EQ:MI_CONT} is modified by replacing the subtracted term with $h(Z)$.}

    %\footnote{For $\ell = 1$, we have $h(T_1|X)=h(Z_1)=\frac{d_1}{2}\log(2\pi e\beta^2)$ because its previous layer is $X$ (fixed). Therefore, no estimation of $h(T_1|X)$ in \eqref{EQ:DNN_mutual_information_input} is needed.} 

    %\item \textbf{Conditional Sampling and Estimation Given $\bm{Y}$:} Denote the (finite and) known set of labels by $\mathcal{Y}$. For a fixed label $y\in\mathcal{Y}$, to sample from $P_{S_\ell|Y=y}$ we first collect all the features labeled as $y$, i.e., we consider the set $\mathcal{X}_y\triangleq\{x_i\}_{i\in\mathcal{I}_y}$, where $\mathcal{I}_y\triangleq\big\{i\in[n]\big|y_i=y\big\}$. The features in $\mathcal{X}_y$ are i.i.d. samples from $P_{X|Y=y}$. Feeding each $x\in\mathcal{X}_y$ in to the DNN, collecting produced values at $T_{\ell-1}$ and applying $f_\ell$ on each, results is a set of i.i.d. samples from $P_{S_\ell|Y=y}$. These samples are used to estimate $h(T_\ell|Y=y)=h(P_{S_\ell|Y=y}\ast\varphi_\beta)$.
\end{enumerate}
The knowledge of $\varphi_\beta$ and the generated samples $S_\ell^n$ and $S_\ell^n(X_i)$ can be used to estimate the unconditional and the conditional entropies, from \eqref{EQ:uncond_ent} and \eqref{EQ:cond_ent}, respectively.

For notational simplicity, the layer index $\ell$ is dropped for the remainder of this subsection. With the above sampling procedure we construct an estimator $\hat{I}_\mathsf{SP}\big(X^n,\hat{h}\big)$ of $I(X;T)$ based on a given estimator $\hat{h}(A^n,\beta)$ of $h(P\ast\varphi_\beta)$ for $P\in\mathcal{F}_d$ that uses  i.i.d. samples $A^n=(A_1,\ldots,A_n)$ from $P$ and knowledge of $\varphi_\beta$. Assume that $\hat{h}$~attains
\begin{equation}
    \sup_{P\in\mathcal{F}_d} \mathbb{E}\left|h(P\ast\varphi_\beta)-\hat{h}(A^n,\beta)\right|\leq \Delta_{\beta,d}(n).\label{EQ:Data_Dist_Assumption}
\end{equation}
An example of such an $\hat{h}$ is the estimator $h(\hat P_{S^n}\ast\varphi_\beta)$ from Theorem \ref{TM:sample_complex_bound}; the corresponding $\Delta_{\beta,d}(n)$ term is the RHS of \eqref{EQ:Plugin_risk_bound}. Our SP mutual information estimator is (see \eqref{EQ:MI_estimator_final})
\begin{equation}
\hat{I}_{\mathsf{SP}}\left(X^n,\hat{h},\beta\right)\triangleq \hat{h}(S^n,\beta) - \frac{1}{n} \sum_{i=1}^n \hat{h}\big(S^n(X_i),\beta\big).\label{EQ:MI_CONT}
\end{equation}
The following theorem bounds the expected absolute error of $\hat{I}_{\mathsf{SP}}\p{X^n,\hat{h},\beta}$. The proof is given in Supplement \ref{SUPPSUBSEC:MI_True_Data_Dist_proof}.

\begin{theorem}\label{TM:MI_True_Data_Dist}
For the above described setup, we have
\begin{align*}
    \sup_{P_X}\mathbb{E} \Big|I(X;T)-&\hat{I}_{\mathsf{SP}}\p{X^n,\hat{h},\beta}\Big|\\&\leq 2\Delta_{\beta,d}(n)+\frac{d\log\left(1+\frac{1}{\beta^2}\right)}{4\sqrt{n}}.\numberthis
\end{align*}
\end{theorem}
Theorem \ref{CORR:MI_True_Data_Dist} of the main text is an immediate consequence of Theorems \ref{TM:sample_complex_bound} and \ref{TM:MI_True_Data_Dist}. Interestingly, the quantity $\frac{1}{\beta^2}$ is the signal-to-noise ratio (SNR) between $S$ and $Z$. The larger $\beta$ is the easier estimation becomes, since the noise smooths out the complicated $P_X$ distribution. Also note that the dimension of the ambient space in which $X$ lies does not appear in the absolute-risk bound for estimating $I(X;T)$. The bound depends only on the dimension of $T$ (through $\Delta_{\beta,d}$). This is because the additive noise resides in the $T$ domain, limiting the possibility of encoding the rich structure of $X$ into $T$ in full. On a technical level, the blurring effect caused by the noise enables uniformly lower bounding $\inf_x h(T|X=x)$ and thereby controlling the variance of the estimator for each conditional entropy. In turn, this reduces the impact of $X$ on the estimation of $I(X;T)$ to that of an empirical average converging to its expected value with rate $\frac{1}{\sqrt{n}}$.

\subsection{Sample Propagation Estimator Bias}

The results of the previous subsection are of a minimax flavor. That is, they state worst-case convergence rates of $h(P\ast\varphi_\beta)$ estimation over a nonparametric class of distributions. In practice, the true distribution may not be one that attains these worst-case rates, and convergence may be faster. However, while variance of $h(\hat P_{S^n}\ast\varphi_\beta)$ can be empirically evaluated using bootstrapping, there is no empirical test for the bias. Specifically, even if multiple estimations of $h(P\ast\varphi_\beta)$ via $h(\hat P_{S^n}\ast\varphi_\beta)$ consistently produce similar values, this does not necessarily suggest that these values are close to the true $h(P\ast\varphi_\beta)$. To have a guideline to the least number of samples needed to avoid biased estimation, we present the following lower bound on the estimation bias.

%While variance of the SP estimator can be empirically evaluated using bootstrapping, there is no empirical test for the bias. Even if multiple estimations of $h(T)$ via $h_{\mathsf{SP}}(S^n)$ consistently produce similar values, this does not necessarily suggest that these values are close to the true $h(T)$. For relatively low dimensions, Corollaries \ref{CORR:sampleSig} and \ref{CORR:sampleReLU} give satisfactory bounds on the estimation errors. However, the $n^{-\frac{1}{d}}$ rate of decay means that the error bounds from the corollaries drop very slowly with $n$ for large $d$.. To have a guideline to the least number of samples needed to avoid biased estimation, we present the following lower bound on $\sup_{P\in\mathcal{F}_d}\mathbb{E}\left|h(P\ast\varphi_\beta)-\hat{h}_{\mathsf{SP}}(S^n)\right|$. %Similarly to Theorem \ref{TM:sample_complex_worstcase}, the result assumes $2d\cdot Q\left(\frac{1}{2\beta}\right)<1$. Like before, we note that this trivially holds for any relevant values of $d$ and $\beta$ (and much beyond).

\begin{theorem}\label{TM:SP_Bias_LB}
Fix $\beta>0,\ d\geq 1$, and let $\epsilon\in\left(1-\left(1-2Q\left(\frac{1}{2\beta}\right)\right)^d,1\right]$, where $Q$ is the Q-function.\footnote{The Q-function is defined as $Q(x)\triangleq\frac{1}{\sqrt{2\pi}}\int_x^\infty e^{-\frac{t^2}{2}}dt$.} Set $k_\star\triangleq\bigg\lfloor\frac{1}{\beta Q^{-1}\left(\frac{1}{2}\left(1-(1-\epsilon)^{\frac{1}{d}}\right)\right)}\bigg\rfloor$, where $Q^{-1}$ is the inverse of the Q-function. By the choice of $\epsilon$, clearly $k_\star\geq 2$, and the bias of the SP estimator over the class $\mathcal{F}_d$ is bounded as
\begin{equation}
 \sup_{P\in\mathcal{F}_d}\mspace{-3mu}\left|h(P\mspace{-3mu}\ast\mspace{-3mu}\varphi_\beta)\mspace{-3mu}-\mspace{-3mu}\mathbb{E}h(\hat P_{S^n}\mspace{-3mu}\ast\mspace{-3mu}\varphi_\beta)\mspace{-1mu}\right|\mspace{-3mu}\geq\mspace{-3mu} \log\mspace{-3mu}\left(\mspace{-3mu}\frac{k_\star^{d(1-\epsilon)}}{n}\mspace{-3mu}\right)\mspace{-3mu}-\mspace{-3mu}H_b(\mspace{-1mu}\epsilon\mspace{-1mu}). \label{EQ:SP_Bias_LB}
\end{equation}
Consequently, the bias cannot be less than a given $\delta>0$ so long as $n\leq k_\star^{d(1-\epsilon)}\cdot e^{-(\delta+H_b(\epsilon))}$.
\end{theorem}
Theorem \ref{TM:SP_Bias_LB} is proved in Supplement \ref{SUPPSUBSEC:SP_Bias_LB_proof}. Since $H_b(\epsilon)$ shrinks with $\epsilon$, for sufficiently small $\epsilon$ values the lower bound from \eqref{EQ:SP_Bias_LB} shows that the SP estimator will not have negligible bias unless $n> k_\star^{d(1-\epsilon)}$ is satisfied. The condition $\epsilon>1-\left(1-2Q\left(\frac{1}{2\beta}\right)\right)^d$ is non-restrictive in any relevant regime of $\beta$ and $d$. For instance, for typical $\beta$ values we work with - around $0.1$ - this lower bound is at most 0.0057 for all dimensions up to at least $d=10^4$. Setting, e.g., $\epsilon=0.01$ (for which $H_b(0.01)\approx 0.056$), the corresponding $k_\star$ equals 3 for $d\leq 11$ and 2 for $12\leq d\leq 10^4$. Thus, with these parameters, in order to have negligible bias the number of estimation samples $n$ should be at least $2^{0.99d}$, for any conceivably relevant dimension $d$.

\subsection{Computing the Sample Propagation Estimator}\label{SUBSEC:MoG_Entropy}

%Evaluating $\hat{h}_{\mathsf{SP}}(S^n)$ requires computing the differential entropy of a Gaussian mixture. Although it cannot be computed in closed form, this section presents a method for approximate computation via MC integration \citep{robert2004montecarlo}. To simplify the presentation, we present the method for an arbitrary Gaussian mixture without referring to the notation of the estimation setup. 

Evaluating the SP mutual information estimator requires computing the differential entropy of a Gaussian mixture. Although it cannot be computed in closed form, this section presents a method for approximate computation via MCI \citep{robert2004montecarlo}. To simplify the presentation, we present the method for an arbitrary Gaussian mixture without referring to the notation of the estimation setup.

Let $g(t)\triangleq\frac{1}{n}\sum_{i\in[n]} \varphi_\beta(t-\mu_i)$ be a $d$-dimensional $n$-mode Gaussian mixture, with $\{\mu_i\}_{i\in[n]}\subset\mathbb{R}^d$ and $\varphi_\beta$ as the PDF of $\mathcal{N}(0,\beta^2\mathrm{I}_d)$. Let $C\sim\mathsf{Unif}\{\mu_i\}_{i\in[n]}$ be independent of $Z\sim\mathcal{N}(0,\beta^2\mathrm{I}_d)$ and note that $V\triangleq C+Z\sim g$.

We use MCI \citep{robert2004montecarlo} to compute $h(g)$. First note that
\begin{align*}
    h(g)&=-\mathbb{E}\log g(V)\\
    &=-\frac{1}{n}\sum_{i\in[n]}\mathbb{E}\Big[\log g(\mu_i+Z)\Big|C=\mu_i\Big]\\
    &=-\frac{1}{n}\sum_{i\in[n]}\mathbb{E}\log g(\mu_i+Z),\numberthis\label{EQ:MC_expansion}
\end{align*}
where the last step follows by the independence of $Z$ and $C$. Let $\big\{Z_j^{(i)}\big\}_{\substack{i\in[n]\\j\in[n_\mathsf{MC}]}}$ be $n\times n_\mathsf{MC}$ i.i.d. samples from $\varphi_\beta$. For each $i\in[n]$, we estimate the $i$-th summand on the RHS of \eqref{EQ:MC_expansion} by
\begin{subequations}
\begin{equation}
    \hat{L}_\mathsf{MC}^{(i)}\triangleq \frac{1}{n_\mathsf{MC}}\sum_{j\in[n_\mathsf{MC}]}\log g\left(\mu_i+Z_j^{(i)}\right),\label{EQ:MC_per_i_est}
\end{equation}
which produces 
\begin{equation}
    \hat{h}_\mathsf{MC}\triangleq \frac{1}{n}\sum_{i\in[n]}\hat{L}_\mathsf{MC}^{(i)}\label{EQ:MC_full_est}
\end{equation}
\end{subequations}
as our estimate of $h(g)$. Define the mean squared error (MSE) of $\hat{h}_\mathsf{MC}$ as
\begin{equation}
    \mathsf{MSE}\left(\hat{h}_\mathsf{MC}\right)\triangleq \mathbb{E}\bigg[\Big(\hat{h}_\mathsf{MC}-h(g)\Big)^2\bigg].
\end{equation}
We have the following bounds on the MSE for tanh and ReLU networks. 

\begin{theorem}[MSE Bounds for MC Estimator]\label{TM:MC_MSE}The following holds:\\
\begin{enumerate}
    \item Assume $C\in[-1,1]^d$ almost surely (i.e., tanh network), then 
    \begin{equation}
        \mathsf{MSE}\left(\hat{h}_\mathsf{MC}\right)\leq \frac{2d(2+\beta^2)}{\beta^2}\frac{1}{n\cdot n_\mathsf{MC}}.\label{EQ:MC_MSE_Tanh}
    \end{equation}
    \item Assume $M_C\triangleq\mathbb{E}\|C\|_2^2<\infty$ (e.g., ReLU network with bounded second moments), then
    \begin{align*}
        &\mathsf{MSE}\left(\hat{h}_\mathsf{MC}\right)\\&\leq \frac{9d\beta^2+8(2+\beta\sqrt{d})M_C+3(11\beta\sqrt{d}+1)\sqrt{M_C}}{\beta^2}\\
        &\quad\quad\quad\quad\quad\quad\quad\quad\quad\quad\quad\quad\quad\times\frac{1}{n\cdot n_\mathsf{MC}}.\numberthis\label{EQ:MC_MSE_ReLU}
    \end{align*}
\end{enumerate}
\end{theorem}
The proof of Theorem \ref{TM:MC_MSE} is found in Supplement \ref{SUPPSUBSEC:MC_MSE_proof}.
The MSE bounds scale only linearly with the dimension $d$, making $\beta^2$ in the denominator often the dominating factor experimentally.

\section{Proofs}\label{SUPPSEC:proofs}

%%%%%%%%%%%%%%%%%%%%%%%%%%%%%%%%%%%%%%%%%%%%%%%%%%%%%%%%%%%%%%%%%%%%%%%%%%%%%%%%%%%%%%%%%%%%%%%%%%%%%%%
%%%%%%%%%%%%%%%%%%%%%%%%%%%%%%%%%%%%%%%%%%%%%%%%%%%%%%%%%%%%%%%%%%%%%%%%%%%%%%%%%%%%%%%%%%%%%%%%%%%%%%%
%%%%%%%%%%%%%%%%%%%%%%%%%%%%%          Corollary 1 Proof           %%%%%%%%%%%%%%%%%%%%%%%%%%%%%%%%%%%%
%%%%%%%%%%%%%%%%%%%%%%%%%%%%%%%%%%%%%%%%%%%%%%%%%%%%%%%%%%%%%%%%%%%%%%%%%%%%%%%%%%%%%%%%%%%%%%%%%%%%%%%
%%%%%%%%%%%%%%%%%%%%%%%%%%%%%%%%%%%%%%%%%%%%%%%%%%%%%%%%%%%%%%%%%%%%%%%%%%%%%%%%%%%%%%%%%%%%%%%%%%%%%%%
\subsection{Proof of Theorem \ref{TM:MI_True_Data_Dist}}\label{SUPPSUBSEC:MI_True_Data_Dist_proof}

Fix $P_X$, define $g(x)\triangleq h(T|X=x)=h(P_{S|X=x}\ast\varphi_\beta)$ and write
\begin{equation}
    I(X;T)=h(T)-h(T|X)=h(P_S\ast\varphi_\beta)-\mathbb{E}g(X).
\end{equation}
Applying the triangle inequality to \eqref{EQ:MI_CONT} we obtain
\begin{align*}
\mathbb{E}\Big|\hat{I}_{\mathsf{SP}}&\p{X^n,\hat{h},\beta} - I(X;T)\Big|\\
&\leq \mathbb{E} \left|\hat{h}(S^n,\beta) - h(P_S\ast\varphi_\beta) \right|\\
&\quad\quad\quad\quad+ \mathbb{E}\left|\frac{1}{n} \sum_{i=1}^n \hat{h}\big(S^n(X_i),\beta\big) - \mathbb{E} g(X)\right|\\
&\leq \underbrace{\mathbb{E} \left|\hat{h}(S^n,\beta) - h(P_S\ast\varphi_\beta) \right|}_{(\mathrm{I})}\\
&\quad\quad\quad+ \underbrace{\frac{1}{n} \sum_{i=1}^n \mathbb{E} \left| \hat{h}\big(S^n(X_i),\beta\big) -  g(X_i)\right|}_{(\mathrm{II})}\\
&\quad\quad\quad\quad\quad\quad+ \underbrace{\mathbb{E}\left|\frac{1}{n} \sum_{i=1}^n g(X_i)- \mathbb{E} g(X)\right|}_{(\mathrm{III})}\numberthis\label{eq:BarError}
\end{align*}
By assumption \eqref{EQ:Data_Dist_Assumption} and because $\supp(P_S)\subseteq[-1,1]^d$, we have \begin{equation}
\mathbb{E} \left|\hat{h}(S^n,\beta) - h(P_S\ast\varphi_\beta)\right| \leq \Delta_{\beta,d}(n).\label{EQ:FIRSTTERM}
\end{equation}
Similarly, for any fixed $X^n=x^n$, $\supp(P_{S|X=x_i})\subseteq[-1,1]^d$ for all $x_i$, where $i\in[n]$, and hence
\begin{align*}
    \mathbb{E}\bigg[&\left| \hat{h}(S^n(X_i),\beta)- g(X_i)\right|\bigg|X^n=x^n\bigg]\\
    &\quad\quad\quad\stackrel{(a)}=\mathbb{E}\left| \hat{h}\big(S^n(x_i),\beta\big)- h(P_{S|X=x_i}\ast\varphi_\beta)\right|\\
    &\quad\quad\quad\leq \Delta_{\beta,d}(n),\numberthis\label{EQ:FIRSTTERM_temp}
\end{align*}
where (a) is because for a fixed $x_i$, sampling from $P_{S|X=x_i}$ corresponds to drawing multiple noise realization for the previous layers of the DNN. Since these noises are independent of $X$, we may remove the conditioning from the expectation. Taking an expectation on both sides of \eqref{EQ:FIRSTTERM_temp} and the law of total expectation we have
\begin{equation}
(\mathrm{II}) = \frac{1}{n} \sum_{i=1}^n \mathbb{E} \left|\hat{h}(S^n(X_i))- g(X_i)\right|\leq\Delta_{\beta,d}(n).\label{EQ:SECONDTERM}
\end{equation}

Turning to term $(\mathrm{III})$, observe that $\big\{g(X_i)\big\}_{i\in[n]}$ are i.i.d random variables. Hence 
\begin{equation}
\frac{1}{n} \sum_{i=1}^n g(X_i)- \mathbb{E}g(X)
\end{equation}
is the difference between an empirical average and the expectation. By monotonicity of moments we have
\begin{align*}
(\mathrm{III})^2 &= \left(\mathbb{E} \left|\frac{1}{n} \sum_{i=1}^n g(X_i)- \mathbb{E}g(X)\right|\right)^2\\
&\leq \mathbb{E}\left[\left(\frac{1}{n} \sum_{i=1}^n g(X_i)- \mathbb{E}g(X)\right)^2\right]\\
&= \frac{1}{n} \var \big(g(X)\big)\\
&\leq \frac{1}{4n}\left(\sup_x h(p_{T|X=x}) - \inf_x {h}(p_{T|X=x})\right)^2.\numberthis\label{EQ:THIRDTERM}
\end{align*}
The last inequality follows since $\var(A)\leq \frac{1}{4}(\sup A - \inf A)^2$ for any random variable $A$.

It remains to bound the supremum and infimum of $h(p_{T|X=x})$ uniformly in $x\in\mathbb{R}^{d_0}$. By definition $T = S + Z$, where $S$ and $Z$ are independent and $Z \sim \mathcal{N}(0,\beta^2 \mathrm{I}_d)$. Therefore, for all $x\in\mathbb{R}^{d_0}$
\begin{equation}
    h(p_{T|X =x})\geq h(S + Z | S,X=x)=\frac{d}{2}\log( 2\pi e \beta^2),\label{EQ:INFX}
\end{equation}
where we have used the independence of $Z$ and $(S,X)$ and the fact that conditioning cannot increase entropy. On the other hand, denoting the entries of $T$ by $T\triangleq\big(T(k)\big)_{k\in[d]}$, we can obtain an upper bound as
\begin{equation}%\label{EQ:SUPEX}
    h(p_{T|X=x})=h(T|X=x)\leq \sum_{k=1}^d h\big(T(k)\big|X=x\big),
\end{equation}
since independent random variables maximize differential entropy. Now for any $k\in[d]$, we have
\begin{equation}
    \var\big(T(k)\big|X=x\big)\leq\mathbb{E}\big[T^2(k)\big|X=x\big]\leq 1+\beta^2,
\end{equation}
since $S(k)\in[-1,1]$ almost surely. For a fixed variance the Gaussian distribution maximizes differential entropy, and therefore
\begin{equation}
h(p_{T|X =x}) \leq \frac{d}{2}\log\big(2\pi e (1+\beta^2)\big).\label{EQ:SUPEX}
\end{equation}
for all $x\in\mathbb{R}^{d_0}$. Substituting the lower bound \eqref{EQ:INFX} and upper bound \eqref{EQ:SUPEX} into \eqref{EQ:THIRDTERM} gives
\begin{equation}
(\mathrm{III})^2 \leq \left(\frac{d\log\left(1+\frac{1}{\beta^2}\right)}{4\sqrt{n}}\right)^2. \end{equation}
Inserting this along with \eqref{EQ:FIRSTTERM} and  \eqref{EQ:SECONDTERM} into the bound \eqref{eq:BarError} bounds the expected estimation error as
\begin{equation}
    \mathbb{E}\mspace{-3mu}\left|\hat{I}_{\mathsf{SP}}\mspace{-3mu}\p{\mspace{-3mu}X^n\mspace{-3mu},\mspace{-2mu}\hat{h},\beta\mspace{-3mu}}\mspace{-3mu}-\mspace{-3mu}I(X;T)\right|\mspace{-3mu}\leq\mspace{-3mu} 2\Delta_{\beta,d}(n)+\frac{d\log\left(1\mspace{-4mu}+\mspace{-4mu}\frac{1}{\beta^2}\right)}{4\sqrt{n}}.
\end{equation}
Taking the supremum over $P_X$ concludes the proof.

%%%%%%%%%%%%%%%%%%%%%%%%%%%%%%%%%%%%%%%%%%%%%%%%%%%%%%%%%%%%%%%%%%%%%%%%%%%%%%%%%%%%%%%%%%%%%%%%%%%%%%%
%%%%%%%%%%%%%%%%%%%%%%%%%%%%%%%%%%%%%%%%%%%%%%%%%%%%%%%%%%%%%%%%%%%%%%%%%%%%%%%%%%%%%%%%%%%%%%%%%%%%%%%
%%%%%%%%%%%%%%%%%%%%%%%%%%%%%           Theorem 4 Proof            %%%%%%%%%%%%%%%%%%%%%%%%%%%%%%%%%%%%
%%%%%%%%%%%%%%%%%%%%%%%%%%%%%%%%%%%%%%%%%%%%%%%%%%%%%%%%%%%%%%%%%%%%%%%%%%%%%%%%%%%%%%%%%%%%%%%%%%%%%%%
%%%%%%%%%%%%%%%%%%%%%%%%%%%%%%%%%%%%%%%%%%%%%%%%%%%%%%%%%%%%%%%%%%%%%%%%%%%%%%%%%%%%%%%%%%%%%%%%%%%%%%%
\subsection{Proof of Theorem \ref{TM:SP_Bias_LB}}\label{SUPPSUBSEC:SP_Bias_LB_proof}

First note that since $h(q)$ is concave in $q$ and because $\mathbb{E}\hat{P}_{S^n}=P$, by Jensen's inequality we have
\begin{equation}
    \mathbb{E} h(\hat{P}_{S^n}\ast \varphi_\beta) \leq h(P\ast\varphi_\beta).\label{EQ:SP_LB_truth}
\end{equation}
Now, let $W\sim\mathsf{Unif}([n])$ be independent of $(S^n,Z)$ and define $Y=S_W+Z$. We have the following lemma.

\begin{lemma}\label{LEMMA:SP_bias_MI}
The following equality holds:
\begin{equation}
    h(P\ast\varphi_\beta)-\mathbb{E}h(\hat{P}_{S^n}\ast \varphi_\beta)=I(S^n;Y).
\end{equation}
\end{lemma}

\begin{proof}
We expand $I(S^n;Y)=h(Y)-h(Y|S^n)$ and denote by $F_A$ the cumulative distribution function (CDF) of a random variable $A$. Let $T=S+Z\sim P\ast\varphi_\beta$ and first note that
\begin{equation}
    F_Y(y)=\mathbb{P}\big(S_W+Z\leq y\big)=\frac{1}{n}\sum_{i=1}^n\mathbb{P}(S_i+Z\leq y)=F_T(y).
\end{equation}
Thus, $h(Y)=h(P\ast\varphi_\beta)$. 

It remains to show that $h(Y|S^n)=\mathbb{E}h(\hat{P}_{S^n}\ast\varphi_\beta)$. Fix $S^n=s^n$ and consider
\begin{equation}
    F_{Y|S^n}(y|s^n)=\mathbb{P}\big(S_W+Z\leq y\big|S^n=s^n\big)=\frac{1}{n}\mathbb{P}\big(s_i+Z\leq y\big),
\end{equation}
which implies that the density $p_{Y|S^n=s^n}=\hat{P}_{s^n}\ast\varphi_\beta$. Consequently, $h(Y|S^n=s^n)=h(\hat{P}_{s^n}\ast\varphi_\beta)$, and by definition of conditional entropy $h(Y|S^n)=\mathbb{E}h(\hat{P}_{S^n}\ast\gamma)$.

\end{proof} 
Using the lemma, we have
\begin{equation}
    \left|\sup_{P\in\mathcal{F}_d}\mathbb{E}h(P\ast\varphi_\beta)-h(\hat{P}_{S^n}\ast\varphi_\beta)\right|=\sup_{P\in\mathcal{F}_d}I(S^n;Y),\label{EQ:Bias_MI_relation}
\end{equation}
where the right hand side is the mutual information between $n$ i.i.d. random samples $S_i$ from $P$ and the random vector $Y=S_W+Z$, formed by choosing one of the $S_i$'s at random and adding Gaussian noise. 

To obtain a lower bound on the supremum, we consider the following $P$. Partition the hypercube $[-1,1]^d$ into $k^d$ equal-sized smaller hypercubes, each of side length $k$. Denote these smaller hypercubes as $\mathsf{C}_1,\mathsf{C}_2,\ldots,\mathsf{C}_{k^d}$ (the exact order does not matter). For each $i\in[k^d]$ let $c_i\in\mathsf{C}_i$ be the centroid of the hypercube $\mathsf{C}_i$. Let $\mathcal{C}\triangleq\{c_i\}_{i=1}^{k^d}$ and choose $P$ as the uniform distribution over $\mathcal{C}$.

By the mutual information chain rule and the non-negativity of discrete entropy, we have
\begin{align*}
    I(S^n;Y)&=I(S^n;Y,S_W)-I(S^n;S_W|Y)\\
    &\stackrel{(a)}\geq I(S^n;S_W)-H(S_W|Y)\\
    &= H(S_W)-H(S_W|S^n)-H(S_W|Y),\numberthis\label{EQ:MI_LB_SP_bias}
\end{align*}
where step (a) uses the independence of $(S^n,W)$ and $Z$. Clearly $H(S_W)=\log|\mathcal{C}|$, while $H(S_W|S^n)\leq H(S_W,W|S^n)\leq H(W)=\log n$, via the independence of $W$ and $S^n$. For the last (subtracted) term in \eqref{EQ:MI_LB_SP_bias} we use Fano's inequality to obtain
\begin{align*}
    H(S_W|Y)&\leq H\big(S_W\big|\psi_\mathcal{C}(Y)\big)\\
    &\leq H_b\big(\mathsf{P}_\mathsf{e}(\mathcal{C})\big)+\mathsf{P}_\mathsf{e}(\mathcal{C})\cdot\log|\mathcal{C}|,\numberthis
\end{align*}
where $\psi_\mathcal{C}:\mathbb{R}^d\to\mathcal{C}$ is a function for decoding $S_W$ from $Y$ and $\mathsf{P}_\mathsf{e}(\mathcal{C})\triangleq \mathbb{P}\big(S_W\neq\psi_\mathcal{C}(Y)\big)$ is the probability that $\psi_\mathcal{C}$ commits an error. 

Fano's inequality holds for any decoding function $\psi_\mathcal{C}$. We choose $\psi_\mathcal{C}$ as the maximum likelihood decoder, i.e., upon observing a $y\in\mathbb{R}^d$ it returns the closest point to $y$ in $\mathcal{C}$. Denote by $\mathcal{D}_i\triangleq\psi_\mathcal{C}^{-1}(c_i)$ the decoding region on $c_i$, i.e., the region $\cp{y \in \mathbb{R}^d \middle| \psi_\mathcal{C}(y) = c_i}$ that $\psi_\mathcal{C}$ maps to $c_i$. Note that $\mathcal{D}_i=\mathsf{C}_i$ for all $i\in[k^d]$ for which $\mathsf{C}_i$ doesn't intersect with the boundary of $[-1,1]^d$. When $Y = S_W + Z$, $S_W \sim \mathsf{Unif}(\mathcal{C})$ and the probability of error for the decoder $\psi_\mathcal{C}$ is bounded as:
\begin{align*}
    \mathsf{P}_{\mathsf{e}}(\mathcal{C})&=\frac{1}{k^d}\sum_{i=1}^{k^d}\mathbb{P}\Big(\psi_\mathcal{C}(c_i+Z)\neq c_i\Big|S_W=c_i\Big)\\
    &=\frac{1}{k^d}\sum_{i=1}^{k^d}\mathbb{P}\big(c_i+Z\notin\mathcal{D}_i\big)\\
    &\stackrel{(a)}\leq \mathbb{P}\left(\|Z\|_\infty > \frac{2/k}{2}\right)\\
    &\stackrel{(b)} = 1 - \left(1 - 2 Q\left(\frac{1}{k\beta}\right)\right)^d,
    %\leq 2d Q\left(\frac{1}{k\beta}\right),
    \numberthis\label{EQ:Error_Prob_UB_bias}
\end{align*}
where (a) holds since the $\mathsf{C}_i$ have sides of length $2/k$ and the error probability is largest for $i\in[k^d]$ such that $\mathsf{C}_i$ is in the interior of $[-1,1]^d$. Step (b) follows from independence and the definition of the Q-function.

%    &\stackrel{(a)}\leq \mathbb{P}\left(\|Z\|_\infty > \frac{2/k}{2}\right)\\
%    &\stackrel{(b)}\leq 2d Q\left(\frac{1}{k\beta}\right),\numberthis\label{EQ:Error_Prob_UB}
%\end{align*}
%where (a) holds since the $\mathsf{C}_i$ have sides of length $2/k$ and the error probability is largest for $i$ such that $\mathsf{C}_i$ is in the interior of $[-1,1]^d$. Step (b) follows from the union bound and the definition of the Q-function.

Taking $k=k_\star$ in \eqref{EQ:Error_Prob_UB_bias} as given in the statement of the theorem
gives the desired bound $\mathsf{P}_{\mathsf{e}}(\mathcal{C})\leq \epsilon$. Collecting the pieces and inserting back to \eqref{EQ:MI_LB_SP_bias}, we obtain
\begin{equation}
    I(S^n;Y)\geq \log\left(\frac{k_\star^{d(1-\epsilon)}}{n}\right)-H_b(\epsilon).
\end{equation}
Together with \eqref{EQ:Bias_MI_relation} this concludes the proof.

%%%%%%%%%%%%%%%%%%%%%%%%%%%%%%%%%%%%%%%%%%%%%%%%%%%%%%%%%%%%%%%%%%%%%%%%%%%%%%%%%%%%%%%%%%%%%%%%%%%%%%%
%%%%%%%%%%%%%%%%%%%%%%%%%%%%%%%%%%%%%%%%%%%%%%%%%%%%%%%%%%%%%%%%%%%%%%%%%%%%%%%%%%%%%%%%%%%%%%%%%%%%%%%
%%%%%%%%%%%%%%%%%%%%%%%%%%%%%           Theorem 5 Proof            %%%%%%%%%%%%%%%%%%%%%%%%%%%%%%%%%%%%
%%%%%%%%%%%%%%%%%%%%%%%%%%%%%%%%%%%%%%%%%%%%%%%%%%%%%%%%%%%%%%%%%%%%%%%%%%%%%%%%%%%%%%%%%%%%%%%%%%%%%%%
%%%%%%%%%%%%%%%%%%%%%%%%%%%%%%%%%%%%%%%%%%%%%%%%%%%%%%%%%%%%%%%%%%%%%%%%%%%%%%%%%%%%%%%%%%%%%%%%%%%%%%%
\subsection{Proof of Theorem \ref{TM:MC_MSE}}\label{SUPPSUBSEC:MC_MSE_proof}

Denote the joint distribution of $(C,Z,V)$ by $P_{C,Z,V}$. Marginal or conditional distributions are denoted as usual by keeping only the relevant subscripts. Lowercase $p$ is used to denote a PMF or a PDF depending on whether the random variable in the subscript is discrete or continuous. In particular, $p_C$ is the PMF of $C$, $p_{C|V}$ is the conditional PMF of $C$ given $V$, while $p_Z=\varphi_\beta$ and $p_V=g$ are the PDFs of $Z$ and $V$, respectively.

First observe that the estimator is unbiased:
\begin{equation}
    \mathbb{E}\hat{h}_\mathsf{MC}=-\frac{1}{n\cdot \nmc}\sum_{i=1}^n\sum_{j=1}^{\nmc}\mathbb{E}\log g\left(\mu_i+Z_j^{(i)}\right)=h(g). 
\end{equation}
Therefore, the MSE expands as
\begin{equation}
    \mathsf{MSE}\left(\hat{h}_\mathsf{MC}\right)=\frac{1}{n^2\cdot \nmc}\sum_{i=1}^n\var\Big(\log g(\mu_i+Z)\Big).\label{EQ:MC_MSE_expansion}
\end{equation}

We next bound the variance of $\log g(\mu_i+Z)$ via Poincar{\'e} inequality for the Gaussian measure $\mathcal{N}(0,\beta^2\mathrm{I}_d)$ (with Poincar{\'e} constant $\beta^2)$. For each $i\in[n]$, we have
\begin{equation}
    \var\Big(\log g(\mu_i+Z)\Big)\leq \beta^2 \mathbb{E}\Big[\big\|\nabla\log g(\mu_i+Z)\big\|_2^2\Big].\label{EQ:Poincare}
\end{equation}
We proceed with separate derivations of \eqref{EQ:MC_MSE_Tanh} and \eqref{EQ:MC_MSE_ReLU}. 

%%%%%%%%%%%%%%%%%%%%%%%%%%%%%%%%%%%%%%%%%%%%%%%%%%%%%%%%%%%%%%%%%%%%%%%%%%%%%%%%%%%%%%%%%%%%%%%%%%%%%%%
%%%%%%%%%%%%%%%%%%%%%%%%%%%%%        Monte Carlo MSE - tanh           %%%%%%%%%%%%%%%%%%%%%%%%%%%%%%%%%
%%%%%%%%%%%%%%%%%%%%%%%%%%%%%%%%%%%%%%%%%%%%%%%%%%%%%%%%%%%%%%%%%%%%%%%%%%%%%%%%%%%%%%%%%%%%%%%%%%%%%%%

\subsubsection{MSE Bound for Bounded Support} Since $\|C\|_2\leq \sqrt{d}$ almost surely, Proposition 3 from \cite{PolyWu} implies
\begin{equation}
    \big\|\nabla\log g(v)\big\|_2\leq \frac{\|v\|_2+\sqrt{d}}{\beta^2}.
\end{equation}
Inserting this into the Poincar{\'e} inequality and using $(a+b)^2\leq 2a^2+2b^2$ we have,
\begin{equation}
    \var\Big(\log g(\mu_i+Z)\Big)\leq\frac{2d(4+\beta^2)}{\beta^2},
\end{equation}
for each $i\in[n]$. Together with \eqref{EQ:MC_MSE_expansion}, this concludes the proof of \eqref{EQ:MC_MSE_Tanh}.

%%%%%%%%%%%%%%%%%%%%%%%%%%%%%%%%%%%%%%%%%%%%%%%%%%%%%%%%%%%%%%%%%%%%%%%%%%%%%%%%%%%%%%%%%%%%%%%%%%%%%%%
%%%%%%%%%%%%%%%%%%%%%%%%%%%%%        Monte Carlo MSE - ReLU           %%%%%%%%%%%%%%%%%%%%%%%%%%%%%%%%%
%%%%%%%%%%%%%%%%%%%%%%%%%%%%%%%%%%%%%%%%%%%%%%%%%%%%%%%%%%%%%%%%%%%%%%%%%%%%%%%%%%%%%%%%%%%%%%%%%%%%%%%

\subsubsection{MSE Bound for Bounded Second Moment}

To prove \eqref{EQ:MC_MSE_ReLU}, we use Proposition 2 from \cite{PolyWu} to obtain
\begin{equation}
    \big\|\nabla\log g(v)\big\|_2\leq \frac{1}{\beta^2}\big(3\|v\|_2+4\mathbb{E}\|C\|_2\big).
\end{equation}
Via the Poincar{\'e} inequality from \eqref{EQ:Poincare}, the variance is bounded as
\begin{align*}
    &\var\Big(\log g(\mu_i+Z)\Big)\\&\leq \frac{1}{\beta^2}\mathbb{E}\Big[(3\|\mu_i+Z\|_2+4\mathbb{E}\|C\|)^2\Big]\\
    &\leq \frac{1}{\beta^2}\Big(9d\beta^2+16M_C+24\beta\sqrt{dM_C}\\&\quad\quad\quad+3\|\mu_i\|_2\left(3+9\beta\sqrt{d}+8\beta\sqrt{dM_C}\right)\Big),\numberthis\label{EQ:MC_MSE_ReLU_var_bound}
\end{align*}
where the last step uses H{\"o}lder's inequality (namely, $\mathbb{E}\|C\|_2\leq \sqrt{\mathbb{E}\|C\|_2^2}$). The proof of \eqref{EQ:MC_MSE_ReLU} is concluded by plugging \eqref{EQ:MC_MSE_ReLU_var_bound} into the MSE expression from \eqref{EQ:MC_MSE_expansion} and noting that $\frac{1}{n}\sum_{i=1}^n\|\mu_i\|_2\leq \sqrt{M_C}$.

%\newpage 
%\bibliographystyle{icml2019}
%\bibliography{ref}

\bibliographystyle{icml2019}
\bibliography{ref}

\begin{thebibliography}{36}
\providecommand{\natexlab}[1]{#1}
\providecommand{\url}[1]{\texttt{#1}}
\expandafter\ifx\csname urlstyle\endcsname\relax
  \providecommand{\doi}[1]{doi: #1}\else
  \providecommand{\doi}{doi: \begingroup \urlstyle{rm}\Url}\fi

\bibitem[Achille \& Soatto(2018)Achille and Soatto]{Achille2018}
Achille, A. and Soatto, S.
\newblock On the emergence of invariance and disentangling in deep
  representations.
\newblock \emph{Journal of Machine Learning Research}, 19:\penalty0 1--34,
  2018.

\bibitem[Amjad \& Geiger(2018)Amjad and Geiger]{geiger2018Bottleneck}
Amjad, R.~A. and Geiger, B.~C.
\newblock Learning representations for neural network-based classification
  using the information bottleneck principle.
\newblock arXiv:1802.09766 [cs.LG], 2018.

\bibitem[Belghazi et~al.(2018)Belghazi, Baratin, Rajeswar, Ozair, Bengio,
  Courville, and Hjelm]{Belghazi2018}
Belghazi, M.~I., Baratin, A., Rajeswar, S., Ozair, S., Bengio, Y., Courville,
  A., and Hjelm, R.~D.
\newblock Mutual information neural estimation.
\newblock In \emph{Proceedings of the International Conference on Machine
  Learning (ICML)}, 2018.

\bibitem[Berrett et~al.(2019)Berrett, Samworth, and Yuan]{berrett2019efficient}
Berrett, T.~B., Samworth, R.~J., and Yuan, M.
\newblock Efficient multivariate entropy estimation via $k$-nearest neighbour
  distances.
\newblock \emph{Annals Stats.}, 47\penalty0 (1):\penalty0 288--318, 2019.

\bibitem[Chen(1997)]{chen1997general}
Chen, J.
\newblock A general lower bound of minimax risk for absolute-error loss.
\newblock \emph{Canadian Journal of Statistics}, 25\penalty0 (4):\penalty0
  545--558, Dec. 1997.

\bibitem[Cisse et~al.(2017)Cisse, Bojanowski, Grave, Dauphin, and
  Usunier]{Parseval2017}
Cisse, M., Bojanowski, P., Grave, E., Dauphin, Y., and Usunier, N.
\newblock Parseval networks: Improving robustness to adversarial examples.
\newblock In \emph{Proceedings of the International Conference on Machine
  Learning (ICML)}, 2017.

\bibitem[Cover \& Thomas(2006)Cover and Thomas]{cover2006elements}
Cover, T.~M. and Thomas, J.~A.
\newblock \emph{Elements of Information Theory}.
\newblock John Wiley \& Sons, 2nd edition, 2006.

\bibitem[Goldfeld et~al.(2019)Goldfeld, Greenewald, Weed, and
  Polyanskiy]{anonymized_ISIT_estimation2019}
Goldfeld, Z., Greenewald, K., Weed, J., and Polyanskiy, Y.
\newblock Optimality of the plug-in estimator for differential entropy
  estimation under {Gaussian} convolutions.
\newblock \emph{Accepted to {IEEE International Symposium on Information Theory
  (ISIT-2019)}}, July 2019.

\bibitem[Haje \& Golubev(2009)Haje and Golubev]{el2009entropy}
Haje, H. F.~E. and Golubev, Y.
\newblock On entropy estimation by m-spacing method.
\newblock \emph{Journal of Mathematical Sciences}, 163\penalty0 (3):\penalty0
  290--309, Dec. 2009.

\bibitem[Hall(1984)]{hall1984limit}
Hall, P.
\newblock Limit theorems for sums of general functions of m-spacings.
\newblock \emph{Mathematical Proceedings of the Cambridge Philosophical
  Society}, 96\penalty0 (3):\penalty0 517--532, Nov. 1984.

\bibitem[Hall \& Morton(1993)Hall and Morton]{hall1993estimation}
Hall, P. and Morton, S.~C.
\newblock On the estimation of entropy.
\newblock \emph{Annals of the Institute of Statistical Mathematics},
  45\penalty0 (1):\penalty0 69--88, Mar. 1993.

\bibitem[Han et~al.(2017)Han, Jiao, Weissman, and Wu]{han2017optimal}
Han, Y., Jiao, J., Weissman, T., and Wu, Y.
\newblock Optimal rates of entropy estimation over {Lipschitz} balls.
\newblock arXiv:1711.02141 [math.ST], 2017.

\bibitem[Hjelm et~al.(2019)Hjelm, Fedorov, Lavoie-Marchildon, Grewal, Bachman,
  Trischler, and Bengio]{Hjelm2019}
Hjelm, R.~D., Fedorov, A., Lavoie-Marchildon, S., Grewal, K., Bachman, P.,
  Trischler, A., and Bengio, Y.
\newblock Learning deep representations by mutual information estimation and
  maximization.
\newblock In \emph{Proceedings of the International Conference on Learning
  Representations (ICLR)}, 2019.

\bibitem[Hsu et~al.(2012)Hsu, Kakade, and Zhang]{hsu2012tail}
Hsu, D., Kakade, S., and Zhang, T.
\newblock A tail inequality for quadratic forms of subgaussian random vectors.
\newblock \emph{Electronic Communications in Probability}, 17, 2012.

\bibitem[Joe(1989)]{joe1989estimation}
Joe, H.
\newblock Estimation of entropy and other functionals of a multivariate
  density.
\newblock \emph{Annals of the Institute of Statistical Mathematics},
  41\penalty0 (4):\penalty0 683--697, Dec. 1989.

\bibitem[Kandasamy et~al.(2015)Kandasamy, Krishnamurthy, Poczos, Wasserman, and
  Robins]{kandasamy2015nonparametric}
Kandasamy, K., Krishnamurthy, A., Poczos, B., Wasserman, L., and Robins, J.~M.
\newblock Nonparametric {von Mises} estimators for entropies, divergences and
  mutual informations.
\newblock In \emph{Advances in Neural Information Processing Systems (NIPS)},
  pp.\  397--405, 2015.

\bibitem[Kolchinsky \& Tracey(2017)Kolchinsky and
  Tracey]{kolchinsky2017estimating}
Kolchinsky, A. and Tracey, B.~D.
\newblock Estimating mixture entropy with pairwise distances.
\newblock \emph{Entropy}, 19\penalty0 (7):\penalty0 361, July 2017.

\bibitem[Kolchinsky et~al.(2019)Kolchinsky, Tracey, and {Van
  Kuyk}]{kolchinsky2019caveats}
Kolchinsky, A., Tracey, B.~D., and {Van Kuyk}, S.
\newblock Caveats for information bottleneck in deterministic scenarios.
\newblock In \emph{Proceedings of the International Conference on Learning
  Representations (ICLR)}, 2019.

\bibitem[Kozachenko \& Leonenko(1987)Kozachenko and
  Leonenko]{kozachenko1987sample}
Kozachenko, L.~F. and Leonenko, N.~N.
\newblock Sample estimate of the entropy of a random vector.
\newblock \emph{Problemy Peredachi Informatsii}, 23\penalty0 (2):\penalty0
  9--16, 1987.

\bibitem[Kraskov et~al.(2004)Kraskov, St{\"o}gbauer, and
  Grassberger]{kraskov2004estimating}
Kraskov, A., St{\"o}gbauer, H., and Grassberger, P.
\newblock Estimating mutual information.
\newblock \emph{Physical Review E}, 69\penalty0 (6):\penalty0 066138, June
  2004.

\bibitem[LeCun et~al.(1999)LeCun, Bottou, Bengio, and Haffner]{MNIST}
LeCun, Y., Bottou, L., Bengio, Y., and Haffner, P.
\newblock Gradient-based learning applied to document recognition.
\newblock \emph{Proceedings of the IEEE}, 86\penalty0 (11):\penalty0
  2278--2324, November 1999.

\bibitem[Levit(1978)]{levit1978asymptotically}
Levit, B.~Y.
\newblock Asymptotically efficient estimation of nonlinear functionals.
\newblock \emph{Problemy Peredachi Informatsii}, 14\penalty0 (3):\penalty0
  65--72, 1978.

\bibitem[Linsker(1988)]{Linsker1988}
Linsker, R.
\newblock Self-organization in a perceptual network.
\newblock \emph{Computer}, 21\penalty0 (3):\penalty0 105--117, March 1988.

\bibitem[Paninski(2003)]{Paninski2003}
Paninski, L.
\newblock Estimation of entropy and mutual information.
\newblock \emph{Neural Computation}, 15:\penalty0 1191--1253, June 2003.

\bibitem[Paszke et~al.(2017)Paszke, Gross, Chintala, Chanan, Yang, DeVito, Lin,
  Desmaison, Antiga, and Lerer]{PyTorch2017}
Paszke, A., Gross, S., Chintala, S., Chanan, G., Yang, E., DeVito, Z., Lin, Z.,
  Desmaison, A., Antiga, L., and Lerer, A.
\newblock Automatic differentiation in {PyTorch}.
\newblock In \emph{NIPS Autodiff Workshop}, 2017.

\bibitem[Polyanskiy \& Wu(2012--2017)Polyanskiy and Wu]{PolyWu-LecNotes}
Polyanskiy, Y. and Wu, Y.
\newblock Lecture notes on information theory.
\newblock 2012--2017.
\newblock URL
  \url{http://people.lids.mit.edu/yp/homepage/data/itlectures_v5.pdf}.

\bibitem[Polyanskiy \& Wu(2016)Polyanskiy and Wu]{PolyWu}
Polyanskiy, Y. and Wu, Y.
\newblock Wasserstein continuity of entropy and outer bounds for interference
  channels.
\newblock \emph{IEEE Transactions on Information Theory}, 62\penalty0
  (7):\penalty0 3992--4002, Jul. 2016.

\bibitem[Robert(2004)]{robert2004montecarlo}
Robert, C.~P.
\newblock \emph{Monte Carlo Methods}.
\newblock Wiley Online Library, 2004.

\bibitem[Saxe et~al.(2018)Saxe, Bansal, Dapello, Advani, Kolchinsky, Tracey,
  and Cox]{DNNs_ICLR2018}
Saxe, A.~M., Bansal, Y., Dapello, J., Advani, M., Kolchinsky, A., Tracey,
  B.~D., and Cox, D.~D.
\newblock On the information bottleneck theory of deep learning.
\newblock In \emph{Proceedings of the International Conference on Learning
  Representations (ICLR)}, 2018.

\bibitem[Shwartz-Ziv \& Tishby(2017)Shwartz-Ziv and Tishby]{DNNs_Tishby2017}
Shwartz-Ziv, R. and Tishby, N.
\newblock Opening the black box of deep neural networks via information.
\newblock arXiv:1703.00810 [cs.LG], 2017.

\bibitem[Singh \& P{\'o}czos(2016)Singh and P{\'o}czos]{singh2016finite}
Singh, S. and P{\'o}czos, B.
\newblock Finite-sample analysis of fixed-k nearest neighbor density functional
  estimators.
\newblock In \emph{Advances in Neural Information Processing Systems}, pp.\
  1217--1225, 2016.

\bibitem[Sricharan et~al.(2012)Sricharan, Raich, and
  Hero]{sricharan2012estimation}
Sricharan, K., Raich, R., and Hero, A.~O.
\newblock Estimation of nonlinear functionals of densities with confidence.
\newblock \emph{IEEE Trans. Inf. Theory}, 58\penalty0 (7):\penalty0 4135--4159,
  Jul. 2012.

\bibitem[Tsybakov \& {Van der Meulen}(1996)Tsybakov and {Van der
  Meulen}]{tsybakov1996root}
Tsybakov, A.~B. and {Van der Meulen}, E.~C.
\newblock Root-$n$ consistent estimators of entropy for densities with
  unbounded support.
\newblock \emph{Scandinavian Journal of Statistics}, pp.\  75--83, Mar. 1996.

\bibitem[van~den Oord et~al.(2018)van~den Oord, Li, and
  Vinyals]{vandenOord2018}
van~den Oord, A., Li, Y., and Vinyals, O.
\newblock Representation learning with contrastive predictive coding.
\newblock arXiv:1807.03748 [cs.LG], 2018.

\bibitem[Villani(2006)]{villani2006optimal}
Villani, C.
\newblock \emph{Optimal transport: old and new}, volume 338.
\newblock Springer Science \& Business Media, 2006.

\bibitem[Wu \& Yang(2016)Wu and Yang]{discrete_entropy_est_Wu2016}
Wu, Y. and Yang, P.
\newblock Minimax rates of entropy estimation on large alphabets via best
  polynomial approximation.
\newblock \emph{IEEE Transactions on Information Theory}, 62\penalty0
  (6):\penalty0 3702--3720, June 2016.

\end{thebibliography}
\flushcolsend
\clearpage

\end{document}